\documentclass[review]{elsarticle}

\usepackage{lineno,hyperref}
\modulolinenumbers[5]
\usepackage{gensymb}
\usepackage{amsmath}
\usepackage{booktabs}
\usepackage{bigstrut}
\usepackage{amssymb}
\usepackage{longtable}
\usepackage{float}
\usepackage[format=plain,font=normalsize]{caption}
\usepackage{multirow}
\journal{Journal of Elsevier}
\newcommand{\tabincell}[2]{\begin{tabular}{@{}#1@{}}#2\end{tabular}}  

\begin{document}

\begin{frontmatter}

\title{A Critical Analysis of Image-based Camera Pose Estimation Techniques}


\author[mymainaddress]{Meng Xu}
\author[mysecondaryaddress]{Youchen Wang}
\author[mythirdaddress]{Bin Xu}
\author[mythirdaddress]{Jun Zhang}
\author[mysecondaryaddress]{Jian Ren}
\author[mymainaddress]{Stefan Poslad}
\author[mythirdaddress]{Pengfei Xu\corref{mycorrespondingauthor}}

\cortext[mycorrespondingauthor]{Corresponding author}
\ead{xupengfeipf@didiglobal.com}

\address[mymainaddress]{Queen Mary University of London, London}
\address[mysecondaryaddress]{Beihang University, Beijing}
\address[mythirdaddress]{Didi Chuxing, Beijing}

\begin{abstract}
Camera, and associated with its objects within the field of view, localization could benefit many computer vision fields, such as autonomous driving, robot navigation, and augmented reality (AR). After decades of progress, camera localization, also called camera pose estimation could compute the 6DoF pose of objects for a camera in a given image, with respect to different images in a sequence or formats. Structure-based localization methods have achieved great success when integrated with image matching or with a coordinate regression stage. Absolute and relative pose regression methods using transfer learning can support end-to-end localisation to directly regress a camera pose but achieve a less accurate performance. Despite the rapid development of multiple branches in this area, a comprehensive, in-depth and comparative analysis is lacking to summarise, classify and compare, structure-based and regression-based camera localization methods. Existing surveys either focus on larger SLAM (Simultaneous Localization and Mapping) systems or on only part of the camera localization method, lack detailed comparisons and descriptions of the methods or datasets used, neural network designs such as loss designs, and input formats, etc. In this survey, we first introduce specific application areas and the evaluation metrics for camera localization pose according to different sub-tasks (learning-based 2D-2D task, feature-based 2D-3D task, and 3D-3D task). Then, we review common methods for structure-based camera pose estimation approaches, absolute pose regression and relative pose regression approaches by critically modelling the methods to inspire further improvements in their algorithms such as loss functions, neural network structures. Furthermore, we summarise what are the popular datasets used for camera localization and compare the quantitative and qualitative results of these methods with detailed performance metrics. Finally, we discuss future research possibilities and applications.
\end{abstract}


\end{frontmatter}


\section{Introduction}
Camera pose is used to describe the position and orientation of a camera in a world coordinate system, with respect to six degrees of freedom (6DoF), using different representations, e.g., a transformation matrix. The 6DoF can be grouped into two categories, translations and rotations: translations are linear, horizontal straightness and vertical straightness; rotations are pitch, yaw and roll. Camera pose also includes the estimation of objects' poses in scenes or scenarios for the camera. Camera pose estimation is useful for a range of applications areas, such as augmented reality, robot navigations, autonomous vehicles. These use the camera pose for further calculations, such as object positions and scene perception. Compared with alternative location sensing devices, such as Light detection and ranging (LiDAR), Global navigation satellite system (GNSS). Camera pose estimation is easier to deploy and can be extended to some downstream high-level tasks (e.g., robot grabbing, robot navigation), where the camera determining the 6DoF can be fused with other sensors (e.g., Inertial Measurement Unit (IMU) sensors, Wi-Fi). The camera localisation task estimates the 6DoF pose of the camera under a world coordinate system in relation to objects from images or videos captured by cameras. With the recent rapid progress of deep learning techniques applied to computer vision, camera pose estimation methods developed from structure-based methods (recovering the camera pose by establishing the correspondence between features in a query image and a 3D structure feature in a scene model) to regression-based methods (regressing the camera pose of a reference through a regressor by optimizing the weights of a neural network). For both indoor and outdoor environments, images or videos captured by cameras could estimate the camera pose, specifically, indoor environments require more accurate pose estimation as indoor spaces are more cluttered. 

This paper focuses on image-based camera pose estimation methods. The system inputs are from camera images, which may include RGB and/or depth images, a single image or image sequences, or videos, from moving or stationary cameras. The final output of the system is the 6DoF pose, but there may also be some intermediate stage results (e.g., retrieval stage outputs, query image related retrieval images, image matching stage outputs corresponding from image pairs) from structure-based localization methods. We divide camera pose methods into two main branches: structure-based localisation methods and direct regression-based localisation methods. In addition to introducing these methods, we also list a detailed artificial neural network analysis and internal structure comparison. In addition, this survey reports the comparison of relative datasets, quantitative and qualitative results, and gives some potential research directions.

\subsection{Application domains}
To the best of our knowledge, there is no current in-depth survey of camera pose applications. Augmented reality (AR) technology is currently a very popular image/video synthesis technology. It can superimpose three-dimensional (3D) virtual objects outside the real environment onto images of the real environment through projection to enhance the real-time images and support seamless integration of virtual and real images. It has a wide range of application prospects, such as military training, education, games and entertainment. The precise positioning of the camera is one of the core steps of the technology, which is to obtain the 6 degrees of freedom (6DoF) of the camera. For AR-related technologies which are usually used indoors, a mobile machine-mounted camera is normally used as it moves more smoothly than a human-moving mounted camera for AR games because it avoids human body micromovements and limb movement. Combining human movements and pose estimation of the camera could improve the AR game experience and have more practical applications, such as 3D reconstruction, etc.

Autonomous driving systems need a positioning module to sense the state of the vehicle, usually input from sensors such as LiDAR and camera sensors. The perception, navigation, planning and control systems of robots or cars need copious and robust, knowledge of their location to decide the actions to do in the next step. Devices on cars or robots use sensor data to measure the pose relative to the initial pose (the camera pose when the camera starts to move) to determine the current pose and then use a matching algorithm, navigation beacons, etc. for positioning and navigation. Outdoor localization accuracy could be lower for navigation compared to those indoor environments because objects may be larger, e.g., cars and objects are further apart, although, parking and collision avoidance may need a higher accuracy. For autonomous driving, real-time localization may be important.

Robot-related applications can use camera pose estimation for visual feedback, such as determining the position and direction of objects and helping manual visual inspections for quality control and safety inspections. It can also adaptively control the trajectory of walking robots. In addition, visual information can be used to track paths, detect obstacles, and identify signs or the environment to determine the location of the robot and avoid obstacles.
\subsection{Survey organization}

The remainder of this survey is organized as follows, section \ref{sec2} introduces the related survey papers and compares them with our survey’s scope and focus (section \ref{subsec2.1}), and critically analyzes the limitation of existing surveys (section \ref{subsec2.2}). This survey mainly focuses on camera pose using detailed problem modelling and gives a technical comparison. Section \ref{sec3} models the camera pose estimation problem with fundamental concepts of images, cameras, and pose (sections \ref{subsec3.1} and \ref{subsec3.2}) and some evaluation metrics (section \ref{subsec3.3}) according to different computation models; 

\begin{figure}[htbp]
\centering 
\includegraphics[width=\textwidth]{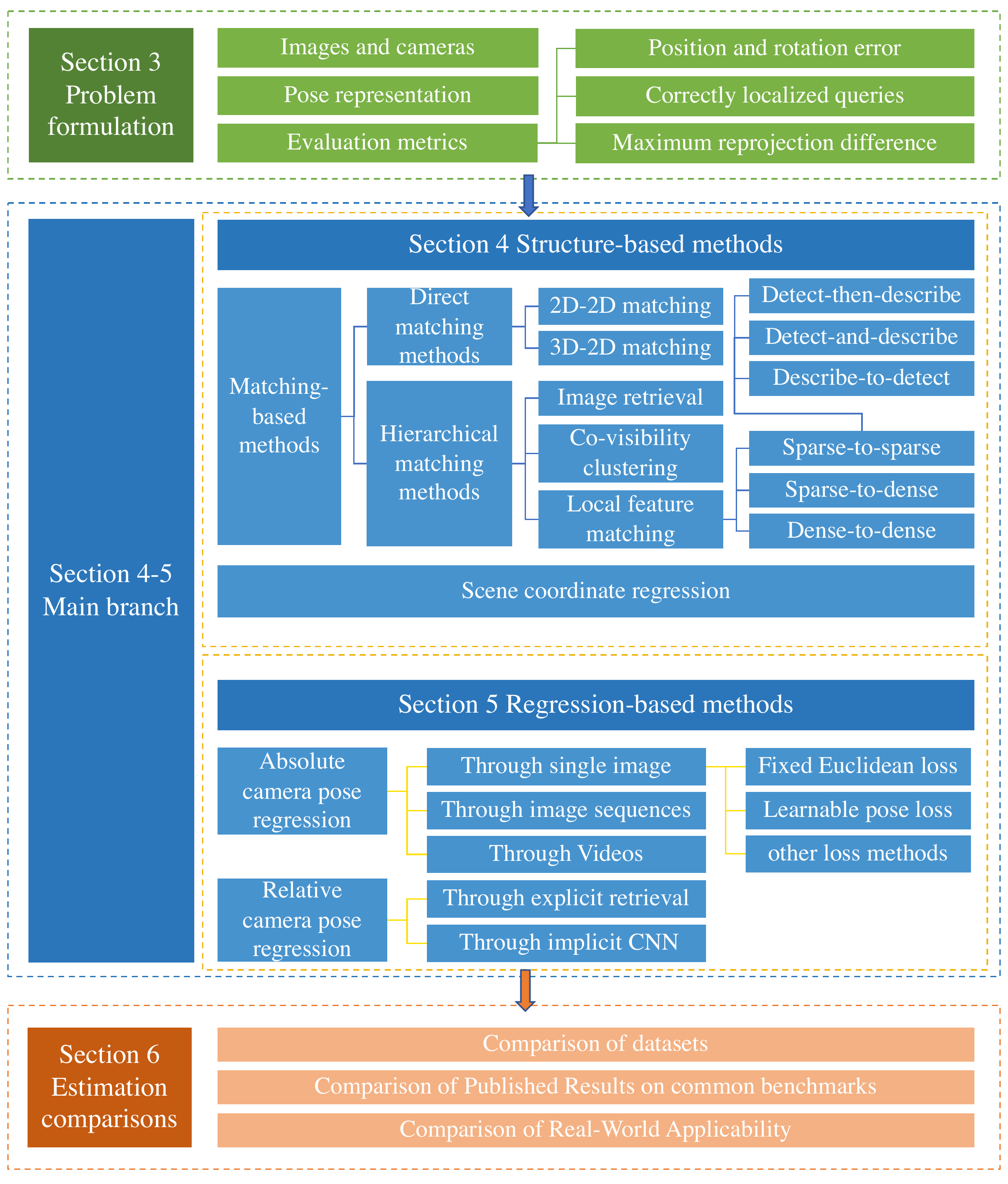} 
\caption{The overview of survey sections 3-6} 
\label{fig1} 
\end{figure}

Section \ref{sec4} analyses in more detail feature-based localization methods that recover the camera pose by establishing the correspondence between features in a query image and a 3D structure feature in a scene model. Matching based localization methods explicitly establish 2D-3D correspondences via matching descriptors (section \ref{subsec4.1}) while scene coordinate regression-based localization methods directly regress 3D scene coordinates from the query image (section \ref{subsec4.2});

Section \ref{sec5} compares the network and loss functions for regression-based localization methods, which regresses the camera pose through the query images. These are classified into absolute camera pose (camera pose that uses global coordinate system) regression method (APR, section \ref{subsec5.1}) and relative pose (camera relative pose between frames) regression method (RPR, section \ref{subsec5.2}). Compared to APR methods, RPR methods first regress the relative pose using retrieval methods from a geo-referenced dataset or other methods and then compute the absolute pose, instead of directly regressing the absolute pose. 

Section \ref{sec6} summarizes the datasets used in the previously proposed methods (section \ref{subsec6.1}) with the dataset size, environment, capture devices, application area, etc., the benchmark results of published papers (section \ref{subsec6.2}) and a real-world application comparison (section \ref{subsec6.3}). Section \ref{sec7} finally summarizes the previously published methods in the camera pose estimation area and proposes future possible directions. Figure \ref{fig1} illustrates the overview of sections \ref{sec3}-\ref{sec6}. 

\section{Related work}
\label{sec2}
\subsection{Related surveys}
\label{subsec2.1}
Piasco et al. \cite{ref2.1} focus on the input data, which classifies the heterogeneous input data into optical, geometric, semantic and cross-data (a fusion of several types of data) information and emphasizes the application of different data features in direct and indirect vision-based localization methods, especially the influence of features on positioning under appearance changes, e.g., light change. Sattler et al. \cite{ref2.2} focus on theoretically modelling the absolute pose regression system and compare structure-based and retrieval-based methods’ pipelines with experiments. This survey finally concludes that APR methods could be improved to compete with other methods, e.g., RPR methods, although a performance gap still exists between them. Wu et al. \cite{ref2.3} model the camera localization problem as a SLAM system and presents the survey’s structure according to whether the environment is known or unknown, whether the mapping process is real-time or offline, etc. 

However, the above surveys only pay attention to a narrow area (e.g., input data of the system, APR system, SLAM system classification), without offering any comprehensive guidance to camera pose localization methods' overview. In contrast to the above surveys, we propose a far wider analysis of camera localization into structure-feature based methods (including matching-based localization and scene coordinate-based localization) and regression-based methods (including absolute camera pose regression-based localization and relative pose regression-based localization).

Some additional surveys focus camera pose more on robotics. Debeunne et al. \cite{ref2.4} introduce vision-based localization methods, LiDAR-based localization methods and their fusion. Chen et al. \cite{ref2.5} discuss visual odometry, mapping, localization and SLAM methods. Shavit et al. \cite{ref2.6} focus on deep learning-based absolute localization methods. Instead of introducing all the SLAM system research work or just explaining their absolute estimation methods, our survey explains camera pose estimation methods using only image input, focusing on analyzing and summarizing the problem definition, the algorithm pipeline, and general approaches to improve the performance of such methods.

There is a lack of analysis and comparison of the two main branches of methods for camera pose estimation, structure-based methods and direct regression-based methods, with respect to architecture, limitations and the benefits of each algorithm.

\subsection{Analysis of limitations of existing surveys}
\label{subsec2.2}
\begin{table}[htbp]
  \centering
  \caption{Comparison of existing camera pose surveys}
   \resizebox{\textwidth}{!}{
    \begin{tabular}{|p{1.75em}|p{13.335em}|p{6.415em}|p{1.75em}|p{1.75em}|p{1.75em}|p{1.75em}|}
    \hline
    Surv ey & Main classification structure & Focus & Data sets & Struc ture & APR & RPR \bigstrut\\
    \hline
    \cite{ref2.1} & Optical, geometric, semantic and cross-data  information & Data structure & $\checkmark$ & \multicolumn{1}{c|}{} & \multicolumn{1}{c|}{} & \multicolumn{1}{c|}{} \bigstrut\\
    \hline
    \cite{ref2.2} & Absolute regression methods, relative regression methods & APR modelling & \multicolumn{1}{c|}{} & \multicolumn{1}{c|}{} & $\checkmark$ & $\checkmark$ \bigstrut\\
    \hline
    \cite{ref2.3} & Known environment, unknown environment & SLAM classification & \multicolumn{1}{c|}{} & \multicolumn{1}{c|}{} & \multicolumn{1}{c|}{} & \multicolumn{1}{c|}{} \bigstrut\\
    \hline
    \cite{ref2.4} & Vision localization, LiDAR localization, fusion methods & Sensor fusion & \multicolumn{1}{c|}{} & \multicolumn{1}{c|}{} & \multicolumn{1}{c|}{} & \multicolumn{1}{c|}{} \bigstrut\\
    \hline
    \cite{ref2.5} & SLAM, Visual Odometry (VO), Mapping, Localization & SLAM system & \multicolumn{1}{c|}{} & \multicolumn{1}{c|}{} & \multicolumn{1}{c|}{} & \multicolumn{1}{c|}{} \bigstrut\\
    \hline
    \cite{ref2.6} & End-to-end localization, hybrid localization & End-to-end networks & \multicolumn{1}{c|}{} & $\checkmark$ & $\checkmark$ & $\checkmark$ \bigstrut\\
    \hline
    Ours & Structure-based localization, regression-based localization & Camera pose estimation & $\checkmark$ & $\checkmark$ & $\checkmark$ & $\checkmark$ \bigstrut\\
    \hline
    \end{tabular}}%
  \label{tab1}%
\end{table}%

Table \ref{tab1} summarises the main structure and focus of each survey. A ‘tick’ indicates whether they contain a dataset comparison, define a structure-based localization method versus an APR method or RPR method. Existing surveys tend to focus on either larger SLAM systems or only on specific camera localization methods. Existing camera pose surveys don’t tend to compare their methods with respect to datasets, and to artificial neural networks (ANNs) with respect to loss functions and input formats such as single image, image sequences and videos, in detail. Moreover, because some pose estimation methods are built upon image retrieval or image matching, our review establishes a model of a two-stage method. The first stage is to retrieve the most similar image of the reference image or to obtain the matching correspondences from the input image pair. The second stage is to regress to the camera pose based on the retrieval or matching results. This survey summarises and classifies image matching methods for camera pose estimation in the structure-based pose estimation stage and seeks to address the lack of a description of such matching or retrieval-based camera pose estimation problems in other surveys. Our survey also reviews structure-based methods and regression-based methods (including APR methods and RPR methods) concerning the analysis of ANNs including loss functions. In addition, this survey also establishes the localization focus formulation of different models, e.g.,2D-2D localization, 2D-3D localisation and 3D-3D localisation, with different multiple datasets formats (e.g., single image, image sequence, videos) and environments. 

\section{Problem formulation}
\label{sec3}
To understand camera pose, we first introduce problem formulation with images, cameras, pose representation and evaluation metrics. Given an image $I_C$, from a monocular or depth camera $C$, a series of methods is applied to the image to get the 6DoF (degree of freedom) coordinates, which represents the position and orientation of the image in 3D space. The visual localization task obtains the pose within a known scene by matching a query object image in a trained model from images of objects in a dataset. According to the types of inquiry, we categorize camera pose methods into three types: (1) structure-based localization to estimate the camera pose by matching 2D pixel features to 3D point scene coordinates or to estimate the camera pose by matching the 3D pixels with a 3D map; (2) absolute pose regression-based localization to directly regress the pose from a 2D map built from an end-to-end neural network; (3) relative pose regression-based localization to regress the relative pose using retrieval methods from a geo-referenced dataset or other methods and then compute the absolute pose.

\subsection{Images and cameras}
\label{subsec3.1}
Generally, for a camera pose estimation task, cameras include monocular, stereo, and RGB-D cameras that take the combination image of an RGB image and its corresponding depth image. The monocular camera applies a pinhole camera model which projects points in the real world, using an external parameter matrix from the world coordinate system to the camera coordinate system, then using an internal parameter matrix from the camera coordinate system to the pixel coordinate system. Through this reprojection process, the points in the world would be represented in an image in the form of pixels while monocular and stereo cameras generate RGB images, a RGB-D camera could simultaneously return RGB images and depth images.
\subsection{Pose representation}
\label{subsec3.2}
Each pose $p$ includes a 3D camera position and orientation. There are many formats to represent the change of position and orientation, i.e., translation and rotation. For orientation, a 3×3 rotation matrix $R$, a 4-digit quaternion, and Euler angles (yaw, pitch, roll) are each interconvertible to represent the same rotation. Usually, the 3-digit coordinates $x$ in 3D space and a 3-digit normalized quaternion $q$ are chosen separately as the position and orientation representations. Thus, the ground truth pose vector $p$ and estimation pose vector $\hat{p}$ could be defined by the combination of translation and rotation:
\begin{equation}
\begin{split}
     p = (x, q) \\
     \hat{p} = (\hat{x}, \hat{q})    
\end{split}
\end{equation}
\subsection{Evaluation metrics}
\label{subsec3.3}
The evaluation approaches for localization tasks change according to the metric focus and localization methods. While evaluating the performance of camera pose estimation methods, we need to compare the computed pose from the estimation method with the ground truth pose to measure how close the estimated result is to the ground truth. Since the camera pose is associated with 3D model coordinates, the standard approach for obtaining ground truth 6DoF poses is to use structure-from-motion (SfM) tools, such as Bundler \cite{ref3.1}, COLMAP \cite{ref3.2}, and VisualSFM \cite{ref3.3} or to use such coordinated provided directly by scanning devices such as Microsoft Kinect.
\subsubsection{Position and rotation error (learning-based 2D-2D)}
For the datasets that directly provide ground truth poses, the pose accuracy of a method is measured by the deviation between the estimated and the ground truth pose, two prominent error metrics for direct localization methods are the absolute pose error (APE) and the relative pose error (RPE). The APE is well-suited for measuring the performance of visual SLAM systems. In contrast, the RPE is well-suited for measuring the drift of a visual odometry system, for example, the drift per second \cite{ref3.4}.
\paragraph{Absolute pose error (APE)} ~{}

When the algorithm input is a single image, the absolute pose error is measured by the combination of absolute position error and orientation error, in which the position error is measured as the Euclidean distance in m between the estimated position $\hat{x}$ and the ground truth position $x$.
\begin{equation}
    t_{ape} = ||x-\hat{x}||_2
\end{equation}

The absolute orientation error $|\alpha|$, measured as an angle in degrees, represents the minimum rotation angle $\alpha$ required to align the ground truth and estimated orientations. The rotation error $\alpha$ could be calculated backwards from the trace of the real and estimated rotation matrix, $R$ and $\hat{R}$, which could also be represented with the real and estimated quaternion, $q$ and $\hat{q}$. 
\begin{equation}
\begin{split}
    2cos(\alpha) = trace(R^{-1}\hat{R})-1 \\
    rot_{ape_{err}} = \alpha = \frac{1}{2}arccos(trace(R^{-1}\hat{R})-1) \\
    rot_{ape_{err}} = \alpha = 2arccos|q\hat{q}|\frac{180\degree}{\pi}    
\end{split}
\end{equation}
\paragraph{Relative pose error (RPE)}~{}

The algorithm is input as image pairs as a time-series from sequential images. Similarly, from the absolute pose error, relative pose error is measured by the combination of relative position error and orientation error, in which the position error is measured as the Euclidean distance shift speed in m/s between the estimated relative position $\hat{x}_{rel}$ and the ground truth relative position $x_{rel}$. The orientation error is measured as the minimum angle deviation rate in degree/s between the estimated relative quaternion $\hat{q}_{rel}$ and the ground truth relative orientation $q_{rel}$ using a quaternion representation. 
\begin{equation}
\begin{split}
     t_{rpe_{err}} = ||x_{rel}-\hat{x}_{rel}||_2 \\
     rot_{rpe_{err}} = \alpha_{rel} = 2arccos|q_{rel}
    \hat{q}_{rel}|\frac{180\degree}{\pi}
\end{split}
\end{equation}

The position and rotation errors are commonly reported using statistical data metrics, e.g., median, mean and the standard deviation of a sequence of images that may each have a position and rotation error.

\subsubsection{Correctly localized queries (feature-based 2D-3D)}
In contrast to directly comparing localization errors, some indirect methods measure the correctly localized queries to represent the performance of the localization algorithms. According to whether the query error thresholds are preset fixed or sampled, we introduce two different error metrics, a fixed thresholds error and a sampled thresholds error.
\paragraph{Fixed thresholds error}~{}

Another metric for indirect camera pose methods measures the percentage of images registered within given error thresholds as the localization performance, i.e., within $X$ meters and $Y$ degrees of their ground truth pose. 

Sattler et al. \cite{ref3.5} define a set of significant margins under strict pose thresholds. Their metrics report the percentage of dataset queries localized within a given error bound on the estimated camera position and orientation, which includes high-precision (0.25m, 2$^{\circ}$), medium-precision (0.5m, 5$^{\circ}$), and coarse-precision (5m, 10$^{\circ}$). These thresholds highlight the overall accuracy as the percentage of images whose localization poses are below the thresholds. The higher the percentage, the better the performance the algorithm shows.
\paragraph{Sampled thresholds error}~{}

Using the same error thresholds for all the images in a dataset will lead to some limitations. For example, images closer to the camera will intuitively get a smaller error range compared to further away images. Thus, sampled error thresholds could set the error thresholds per image \cite{ref3.6} using a set of sampling k ratios, e.g., 50\%, 30\% and 10\% respectively.

Using the ratio @k, or ratio @k\%, one can measure the percentage of queries that have a good match with the k or k\% top-ranked images. Usually, k is set to 10 or 1\%. For example, if the ratio is set to @95\% precision, this means the algorithm is allowed to make a mistake in 5\% of all cases on the percentage of correctly localized images, and only the top 95\% of database candidates need to be considered.
\subsubsection{Maximum reprojection difference (3D-3D task)}

To measure pose accuracy based on reprojections means to measure the difference between the reprojection of a set of 3D points for the ground truth and estimated poses. Defining certain thresholds around the reprojection of the 3D points could avoid the impact of perturbations on the camera that leads to a change in the reprojected 2D locations of 3D points. The maximum reprojection error between the ground truth pose $T$ and estimated pose $\hat{T}$ for the images could be described as $r_i^{\infty}=\mathop{max}\limits_{l\in[1,N_f^i]}||\pi(p_l,T_i)-\pi(p_l,\hat{T}_i||_2 $, and the performance of this algorithm is measured by the percentages of images with $r_i^{\infty}$ that is lower than the preset thresholds of the whole dataset.
\section{Structure feature-based localization methods}
\label{sec4}
The main branch of this camera pose estimation survey consists of structure feature-based and regression-based pose estimation methods. In this section we first model structure feature-based methods. A structure-feature-based localization pipeline refers to methods to recover the camera pose by establishing a correspondence between features in a query image and a 3D structure feature in a scene model, where the 3D point-cloud model is built using structure from motion (SFM) or simultaneous localization and mapping (SLAM) that records the structure of the whole scene. Compared with methods that regress camera pose solely based on object features in images, a structure-feature-based pipeline is more dependent on a priori information of the 3D scene model. 

After establishing the correspondence between the 3D point cloud in the scene model and query images, we can recover the camera pose through geometric constraints which is a classic pipeline that applies a Perspective-n-Point (PnP) to solve to compute the camera pose and uses a Random sample consensus (RANSAC) \cite{ref4.1} method to get rid of outliers.

According to methods that establish a correspondence between a queried image of an object and a 3D sensed model of it, we can divide structure-feature-based localization approaches into two categories: Matching Based Localization and Scene Coordinate Regression-Based Localization. The first method is based on descriptor matching, and the other is based on a trainable localization pipeline. These approaches are described in more detail as follows.

\subsection{Matching based localization}
\label{subsec4.1}
Matching based localization methods use feature descriptors to establish a correspondence between a query image of objects and existing images of object scenes. Usually, the 3D scene model assigns one or several visible local descriptors to each 3D point. Given a query image, we need to extract stable and distinct features to build a correspondence between images and scene models. 

Thus, a descriptor matching based localization task is converted to a feature descriptor matching task. To match features from the scene model, usually, we can compare the distance between descriptors. We can divide features into two categories by the matching method used: direct matching and hierarchical matching. In this paper, we define direct matching methods as direct matching 2D query image feature sets with a 3D scene feature point and hierarchical matching methods as matching 2D query image feature sets with 2D scene database image features to indirectly establish 2D-3D correspondence.

\subsubsection{Direct matching methods}
\label{subsubsec4.1.1}
Previously, large-scale visual localization has been treated as a place recognition problem \cite{ref4.2, ref4.3}. The location for the query image is determined by the most similar image retrieved from the database. But the accuracy of retrieval localization methods does not satisfy challenging applications which need to use an accurate 6DoF pose. To achieve higher accuracy, the use of 3D scene models to estimate pose has been increasingly proposed and used by researchers.

\paragraph{2D-3D matching}~{}

\begin{figure}[htbp]
\centering 
\includegraphics[width=\textwidth]{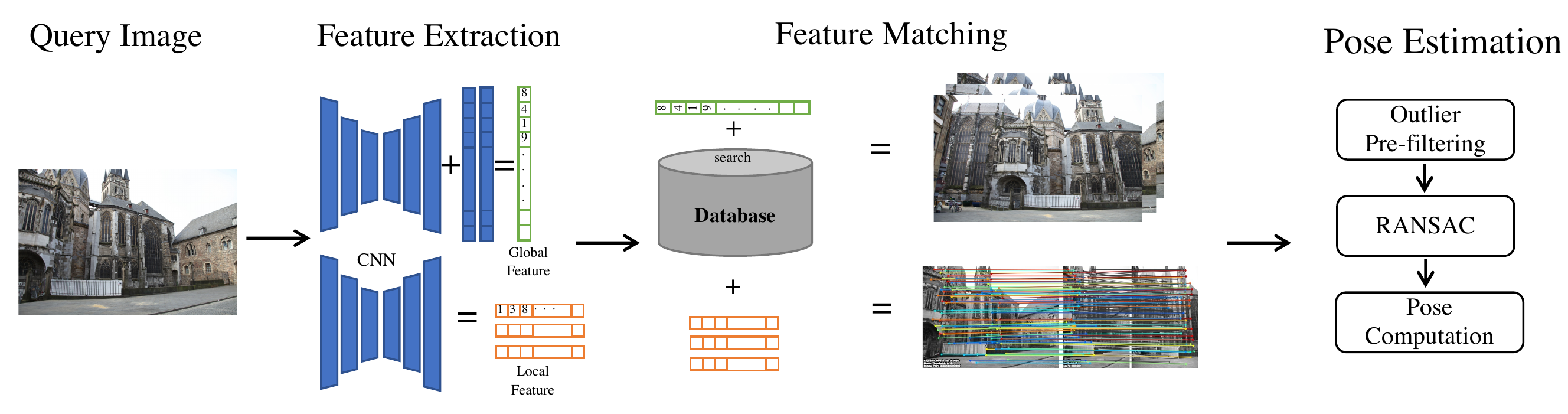} 
\caption{The processing pipeline for direct matching-based localization methods} 
\label{fig2} 
\end{figure}

 It can be intuitive to directly match 2D feature points of query images with 3D feature points to build up a correspondence set, as shown in Figure \ref{fig2}. The main challenge in direct matching methods is to efficiently and effectively find a large enough number of high-quality correspondences to facilitate pose estimation.

In the beginning, much work focused on improving matching methods. For 2D image features and 3D point matching, it’s unclear if any ordering of image features is better than any other. It’s time-consuming to consider all features in the query image. To accelerate 2D-3D matching. Li et al. \cite{ref4.4} propose a “Point-to-feature” matching method, a prioritized point matching algorithm, which matches a subset of scene 3D points to features in the query image and is ordered by a visibility graph. Furthermore, they accelerate search steps by compressing the 3D scene model. Rather than delete images in the database, they found the smallest point cloud subset covers each image at least K times. But Sattler et al. \cite{ref4.5} believe that “Feature-to-point” matching methods also reduce the long-term matching process time. They propose a priority matching method termed Vocabulary-based Prioritized Search (VPS) to speed up descriptor matching. They processed the features in ascending order of their matching costs (starting with features whose activated visual words contain only a few descriptors). Following this, they discussed the advantages and disadvantages of 2D-3D, 3D-2D matching \cite{ref4.6}. Feature points in query images are several orders of magnitude less than those in the model. 3D-2D matching is more effective, but the location accuracy can be lost. 2D-3D matching can filter out wrong matches using a ratio test \cite{ref4.24}. But for a crowded feature space in a large scene model, it is hard to filter out fuzzy matches using a higher threshold of the ratio test. Sattler et al. \cite{ref4.7} proposed an Active Search method that combines 2D-3D and 3D-2D matching. After finding a 2D-to-3D pair, they actively search for 3D-to-2D correspondences for the 3D points closest to the matched point. They used coarse-level features in the vocabulary tree \cite{ref4.8} to recover matches. Sattler et al. \cite{ref4.7} compressed the model by quantizing the point descriptors to achieve run-time localization. By assigning different labels to 3D points, they could use a loose matching strategy to create a locally unique matching set. Similarly, Feng et al. \cite{ref4.9} use a binary descriptor combined with features from accelerated segment test (FAST) \cite{ref4.25} feature point detection to complete feature extraction and proposed an improved binary descriptor retrieval method. They assigned the label information of multiple features to the same 3D point and constructed a supervised trained Random-Forest to complete the matching step. Based on \cite{ref4.5, ref4.6}, Sattler et al. proposed an efficient and effective pipeline. This pipeline uses quantitative feature descriptors to accelerate 2D-3D matching, and 3D-2D matching methods to recall the matching loss due to quantization. In a large-scale environment, similar or repeated feature points always lead to location determination failures. To build a more efficient localization system, Liu et al. proposed a method \cite{ref4.10} that uses global context information to solve this problem. This not only focuses on point-to-Feature 2D-3D matching but jointly processes the feature set of the query graph and the corresponding candidate matching set using a Random Walk with Restart (RWR) algorithm \cite{ref4.11}. 
 
Except for improving matching methods, some work used additional information or outlier filter methods to improve localization accuracy. Svarm et al. \cite{ref4.13} proposed a reliable and tractable outlier rejection scheme that can handle massive amounts of outliers in data. Linus also proposed using vertical coordinates to improve 3D localization \cite{ref4.15}. Similar to \cite{ref4.14}, rather than to just improve the matching precision, this takes a different approach. This allows a matching scheme to generate a large number of matches, whether correct or incorrect, to ensure that no matching matches are missed. Then, they proposed a voting-based geometric verification process, which uses a priori information concerning the direction of gravity and the height of the camera, to filter outliers.
\paragraph{3D-3D matching}~{}

For RGB-D images or stereo systems, we need to match 3D points with the scene point cloud model. Without any already defined descriptors, we usually use the iterative closest point (ICP) method to match 3D points. Except directly using the spatial information, we can still extract features such as shape, density, etc. By comparing descriptors, we can establish a 3D-3D correspondence. Approaches such as \cite{ref4.16, ref4.17} are classic low-level hand-crafted geometric 3D feature descriptors. Choi et al. \cite{ref4.18} use such descriptors for 3D reconstruction to achieve considerable results. But these descriptors can be unstable or inconsistent when used in real-world partial surfaces from 3D scanning data and are difficult to adapt to new datasets. 3D-ShapeNet used 3D deep learning to model point cloud shapes \cite{ref4.19}. Similarly, \cite{ref4.20, ref4.21} focused on extracting features from complete 3D object models at a global level. To provide a more robust descriptor when dealing with partial data suffering from various occlusion patterns and viewpoint differences, \cite{ref4.22, ref4.23} proposed approaches that compose only local level features at a small range. 

Compared with 2D-3D matching based, 3D-3D localization is relatively underexplored.

\subsubsection{Hierarchical matching methods}

\begin{figure}[htbp]
\centering 
\includegraphics[width=\textwidth]{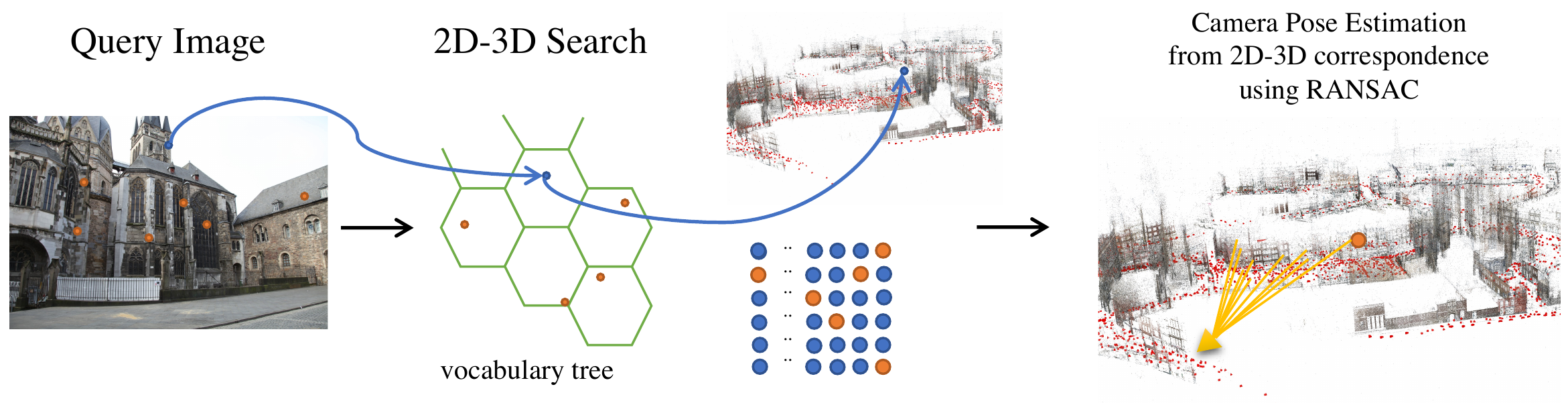} 
\caption{The pipeline of hierarchical matching based localization methods} 
\label{fig3} 
\end{figure}

The direct matching approaches introduced in section \ref{subsubsec4.1.1} mostly rely on estimating correspondences between a 2D feature in the query and 3D points in a sparse model using local descriptors. For direct matching methods, we need to search each 3D point for a query feature, which is not efficient. Although many papers try different ways to improve efficiency and accuracy, they still show a fragile robustness for repetitive local features in the matching process. We aim to substantially increase the robustness of the localization while retaining tractable computational requirements, e.g., the calculation amount cannot exceed the model load and can deal with a certain amount of feature repeatability. Therefore, a coarse-to-fine hierarchical localization paradigm \cite{ref5.1} is proposed to solve this problem. as shown in Figure \ref{fig3}. 

In 2009, Irschara et al. \cite{ref5.2} use a retrieval-based method \cite{ref5.19} to search for the smallest scene model subset. By only estimating correspondences in this subset, they can achieve real-time localization in large-scale datasets. But the robustness of retrieval-based methods is limited by the poor invariance of hand-crafted local features. Recent features emerging from convolutional neural networks (CNN) exhibit far better robustness at a lower computation cost. The pipeline of a hierarchical approach is simple and effective. It follows image retrieval, co-visibility (meaning that two images see common areas), clustering and local feature matching.
\paragraph{Image retrieval}~{}

We address the image retrieval problem as follows, given a query image, a system should efficiently retrieve similar images from the database. Retrieval systems have matured to incorporate spatial verification \cite{ref5.3, ref5.4,ref5.5} and query expansion \cite{ref5.6, ref5.7}. Over more recent years, several image clustering methods based on local features have been proposed, such as Bag-of-Words (BoW) \cite{ref5.22}, Vector of Local Aggregated Descriptors (VLAD), etc. For local aggregation retrieval methods that rely on local features but ignore global contextual information, these tend to show a poor performance when repeated local features appear in a large dataset. Since Krizhevsky et al. \cite{ref5.8} show the advantage of learning-based features, further research followed that used CNN layer activations as off-the-shelf image descriptors that appear as objective results in retrieval tasks \cite{ref5.9, ref5.10}. Following classic retrieval approaches, such work uses CNN to aggregate local features \cite{ref5.11, ref5.12}. Chum et al. \cite{ref5.11} use a classification network followed by the use of a Maximum-Activations-of-Convolutions (MAC) layer to extract feature vectors. This used a 3D model reconstructed by unlabeled images to produce a training dataset. In addition, this proposed a search strategy to find hard positive and hard negative image pairs for training which can impact the results. Arandjelovic et al. \cite{ref5.12} proposed NetVLAD on the basis of VLAD \cite{ref5.20}, as a trainable VLAD layer. They used Google street view\footnote{See https://www.google.com/streetview/} to produce a weakly supervised dataset. For a less accurate location ground truth obtained by global navigation satellite systems (GNSS), triplet-loss is used to ensure that the feature distance of all positive images should be smaller than the feature distance of all negative images. Similarly, with NetVLAD, Radenović et al. \cite{ref5.14} also train their model using hard positive and hard negative through triplet-loss. After extracting dense features through FCN, they used a Generalized-mean pooling layer (GEM) to generate global features. Not only is triplet loss used to learn spatial information, but this approach also exploits the use of second-order spatial attention \cite{ref5.21} in descriptor learning and combines it with second-order descriptor loss to improve the learning global image representation. All of the approaches mentioned above used triplet loss to train the network by accurate location information obtained by the SfM model. Revaud et al. \cite{ref5.16} proposed a method to directly optimize the global retrieval ranking. The author used listwise ranking loss to directly optimize mAP (mean average precision). For the computational memory requirement, a multi-stage backpropagation method was proposed to train the network. In contrast to the above method, DELF \cite{ref5.13} used image classification labels to supervise metric learning. The main innovation is, after extracting dense local features through CNN, they used a landmark classifier to focus on extracting key points and descriptors. Finally, global features are generated through an FCN layer. Teichmann et al. \cite{ref5.17} proposed regionally aggregated match kernels to leverage selected image regions and produce a discriminative image representation. Husain et al. \cite{ref5.18} proposed an approach REMAP, to ensemble the multi-resolution region-based features, which explicitly employs regions discriminative power, measured using Kullback-Leibler (KL) divergence values, to control the aggregation process. All these Global-Single-Pass methods, use global descriptors generated by a single forward-pass through a CNN, have been designed to focus on global contextual information for robust and discriminative global feature descriptors. 

\paragraph{Local feature matching}~{}

We address the local feature matching problem as given a query image and a set of similar scene images, in contrast, to direct matching methods, we should efficiently extract local features and match them based upon query-based similar scene images, instead of 3D feature points in a scene model, according to matching methods such as sparse-to-sparse matching, sparse-to-dense matching and dense-to-dense matching.

\begin{enumerate}[(1)]
\item Sparse-to-sparse matching
\end{enumerate}

Sparse-to-sparse matching is based on sparse feature extraction from images. This method can be further divided into three types: detect-then-describe, detect-and-describe, and describe-to-detect, according to the role of the detector and descriptor in the learning process.

\begin{itemize}
\item The \emph{Detect-then-describe} approach is usually two-stage. First, a keypoint detection is performed then the feature descriptor are extracted around proposed key points. A good local feature typically should be robust and invariant against scale transformation, rotation, and viewpoint, changes. At the start, handcrafted keypoint detectors (such as SIFT, Harris or SUSAN) used gradient and other information to detect key points. This is feasible but can be considered too computationally intensive for use in real-time applications. To accelerate the detection step, Rosten et al. \cite{ref6.1} proposed FAST which was one of the first attempts to use machine learning to derive a corner keypoint detector. Further work extended FAST by adding a descriptor \cite{ref6.2}, or orientation estimation \cite{ref6.3}. Verdie et al. \cite{ref6.4} proposed a new regression-based approach to extract feature points that are repeatable under drastic illumination changes. Lenc et al. \cite{ref6.5} introduce a novel learning formulation for covariant detectors. They proposed to cast detection as a regression problem, then derived a covariance constraint that can be used to automatically learn. Zhang et al. \cite{ref6.6} extend the covariant constraint proposed by \cite{ref6.5} by defining the concepts of standard patch and canonical features, which makes the learning process more robust and less sensitive to the initialization setting. Since it is often unclear what points are "interesting", human labelling cannot be used to find a truly unbiased solution. Savinov et al. \cite{ref6.7} cast detection as an unsupervised formulation. They trained a neural network that maps an object point to a single real-valued response and then ranked points according to this response. DeTone et al. \cite{ref6.8} presented a point tracking system powered by two deep convolutional neural networks MagicPoint and MagicWarp. After that, MagicPoint was extended in \cite{ref6.9} to Superpoint. After detecting the key point, the next step is to extract the descriptor on a sparse set of key points. In the beginning, most work focused on learning descriptors from image patches. Zagoruyko et al. \cite{ref6.10} use Siamese CNN networks to learn discriminant patch representations from a large set of known pairs of corresponding and non-corresponding patches. Han et al. \cite{ref6.11} proposed MatchNet as a deep convolutional network that extracted features from patches and a network of three fully connected layers that computed the similarity between the extracted features. They converted the descriptor regression task into a classification problem under a cross-entropy loss. Simo-Serra et al. \cite{ref6.12} proposed a strategy of aggressive mining of hard positives and negatives on multi-view stereo (MVS) datasets. They also used a Siamese network architecture that employed two CNNs with identical parameters to compare pairs of patches and to treat the CNN outputs as patch descriptors. Besides patch correspondence-based learning, the descriptor is usually trained by a metric loss, such as the triplet loss or a contrastive loss. Balntas et al. \cite{ref6.13} proposed to utilize triplets of training samples, together with in-triplet mining of hard negatives. In addition, they discussed the loss functions when learning with triplets or pairs and investigated their characteristics. L2-Net \cite{ref6.14} focused only on the relative distance which makes positive pairs become the nearest to each other for L2 distance. Mishchuk et al. \cite{ref6.15} proposed HardNet to minimize the distance between the matching descriptor and the closest non-matching descriptor. They proved that their proposed loss is better than complex regularization methods. Tian et al. \cite{ref6.16} proposed SOSNet with a novel regularization term, named Second Order Similarity Regularization (SOSR). This not only forces the distances between matching descriptors to decrease or distances between nonmatching ones to increase, but it also forces the distances between nonmatching descriptors' distances respectively to be equal. Wang et al. \cite{ref6.17} proposed a weakly-supervised framework CAPS that can learn feature descriptors solely from relative camera poses between images. They translate relative camera poses into epipolar constraints between image pairs and enforce the predicted matches to obey this constraint.
\item \emph{Detect-and-describe.} Recently, some work implemented an end-to-end feature detection and descriptor pipeline. As opposed to patch-based neural networks, Detect-and-describe approaches operate on full-sized images and jointly compute interest point locations and associated descriptors in one forward pass. Yi et al. \cite{ref6.18} proposed LIFT, a full-featured point handling pipeline, including detection, orientation estimation, and feature description. LF-Net \cite{ref6.19} is an entire feature extraction pipeline. To train the network end-to-end, they design a two-branch network and optimize by confining it to one branch, while preserving differentiability in the other. SuperPoint \cite{ref6.9}, created a large dataset of pseudo-ground truth interest point locations in real images, supervised by the interest point detector itself. They jointly trained a network called SuperPoint for interest point detection and description. Other than only learning key points, a descriptor R2D2 \cite{ref6.21} is used to train a predictor of the local descriptor discriminator. They argued that salient but discriminative regions can harm performance. ASLFeat \cite{ref6.22} is based on D2-Net \cite{ref6.23}. They improved the ability to model the local shape for stronger geometric invariance, and the ability to localize key points, more accurately.
\item \emph{Describe-to-detect} methods refer to extracting descriptors before detecting key points. Dusmanu et al. \cite{ref6.23} proposed a method D2-Net that detects key points on a dense feature map generated by CNN. By postponing the detection to a later stage, the obtained key points are more stable. Similar to D2D \cite{ref6.24}, they proposed a relative and an absolute saliency measure of local deep feature maps along the spatial and depth dimensions to define key points. Benbihi et al. \cite{ref6.25}, proposed a detection method DELF, valid for any trained CNN where key points are regarded as the local maxima of a saliency map computed as the feature gradient for the input image. 
\end{itemize}

\begin{enumerate}[(2)]
\item Sparse-to-dense matching
\end{enumerate}

In this paper, sparse-to-dense is defined as matching a sparse set of local features with a dense feature map extracted from the image. Germain et al. \cite{ref6.26} proposed an approach for robust and accurate outdoor visual localization. After getting the sparse feature points in the retrieved reference image, they search for the corresponding 2D locations in the query image exhaustively. Inspired by this paper, they proposed S2DNet \cite{ref6.27}, a sparse-to-dense matching pipeline, where they designed and trained a network to predict correspondence for query points. 
\begin{enumerate}[(3)]
\item Dense-to-dense matching
\end{enumerate}

Dense-to-dense matching approaches get rid of the detection stage altogether by finding mutual nearest neighbors in dense feature maps. Most deep feature dense-to-dense matching methods have focused on learning dense descriptors over the image. NCNet \cite{ref6.30} trains a CNN to search in the 4D space of all possible correspondences, with the use of 4D convolutions. Melekhov et al. \cite{ref6.28} proposed a novel approach called DGC-Net. They leverage the advantages of optical flow approaches which have recently achieved significant progress. By extending optical flow to the case of large transformations, they can provide dense and subpixel accurate estimates in a complex environment. Wiles et al. \cite{ref6.29} proposed a new approach to determining correspondences between image pairs under large changes. They designed a model to learn a conditioned feature and distinctiveness score which is then used to choose the best matches by an attention mechanism. In InLoc \cite{ref6.31}, they collected a new dataset with reference 6DoF poses for large-scale indoor localization, and dense feature extraction. Their approach is based on matching on a sequence of progressively stricter verification steps.
\subsubsection{Summary}
These matching methods here attempted to efficiently generate an accurate correspondence between query and scene. This is then used to calculate camera pose by applying a Perspective-n-Point (PnP) solver inside a RANSAC loop. Therefore, the precision of the matching module largely determines the accuracy of positioning.

\subsection{Scene coordinate regression-based localization}
\label{subsec4.2}
In contrast to matching based methods that explicitly establish 2D-3D correspondences via matching descriptors, scene coordinate regression-based localization methods directly regress 3D scene coordinates from the query image. Namely, either a random forest or a neural network is trained to directly predict 3D scene coordinates for the pixels. In this way, correspondences between 2D points in the image and 3D points in the scene can be obtained densely without feature detection and description, and explicit matching. 

In 2017, Rosten et al. \cite{ref6.32} proposed a differentiable RANSAC by soft argmax and probabilistic selection, called DSAC. For end-to-end learning, they put DSAC into the camera localization pipeline. Their trainable localization pipeline exceeds the state-of-the-art results. Such a common pipeline was then improved via reprojection loss \cite{ref6.33, ref6.34, ref6.35}. Cai et al. \cite{ref6.36} use multi-view geometric constraints to enable unsupervised learning. Brachmann et al. \cite{ref6.37} proposed a joint classification regression forest which is trained to predict scene identifiers and scene coordinates. In \cite{ref6.38}, scene coordinate regression is formulated as two separate tasks of object instance recognition and local coordinate regression. Applied to different scenes without any retraining or adaptation, SANet \cite{ref6.39} proposed a method to extract a scene representation from some reference scene images and 3D points, instead of encoding specific scene information in network parameters.

However, most existing scene coordinate regression methods can only be adopted on small-scale scenes. They have not yet proven their capacity to be as effective in large-scale scenes.

\section{Regression-based pose estimation methods}
\label{sec5}
Apart from structure feature-based pose estimation methods, this section focuses on regression-based pose estimation methods, in which we divide this area into absolute camera pose regression and relative camera pose regression according to whether the process is in an end-to-end direct or is a two-stage that integrates image retrieval or CNN process to get a reference image’s pose and then gets the camera pose. 

\subsection{Absolute camera pose regression}
\label{subsec5.1}
Absolute camera pose regression aims to predict a reference image’s 6DoF pose through a CNN by optimizing the weights of the network, which directly output the position and orientation information from an image to the regressor. According to the principle of whether the input of the network is a single image, image sequence or video, the absolute camera pose regression is introduced in three parts. i.e., absolute pose regression through single monocular image, absolute pose regression with image sequences auxiliary, and absolute pose regression through video.

Following on from single monocular image or auxiliary learning, existing research work on APR problems show improvements mainly through 1) replacing the encoder network or adding some modules; 2) modifying networks loss function; 3) enhancing image data by using more images or adding constraints on time, space, etc. 

\subsubsection{Absolute pose regression through single monocular image}

\begin{figure}[htbp]
\centering 
\includegraphics[width=0.6\textwidth]{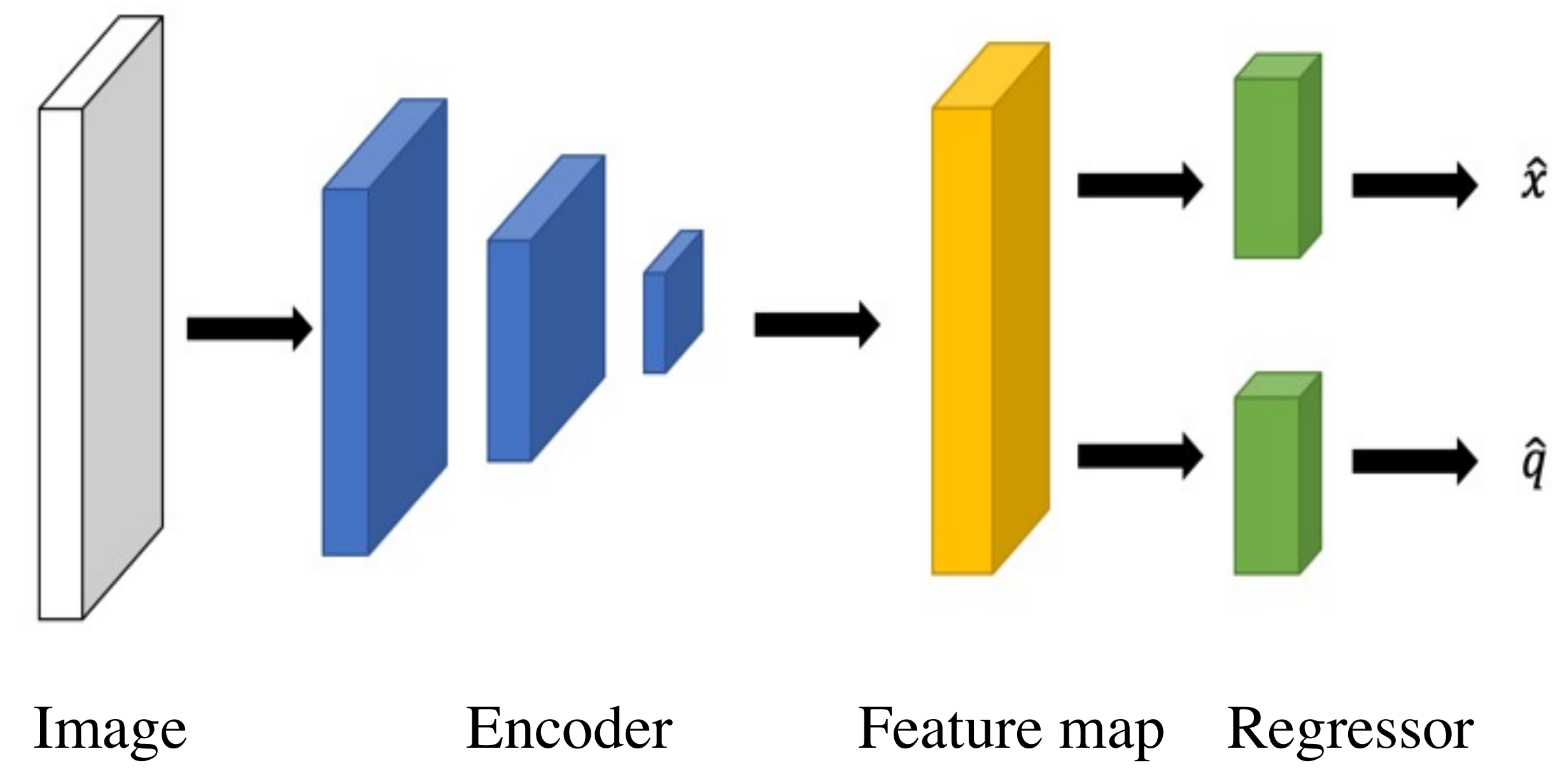} 
\caption{An overview of absolute pose regression methods architecture} 
\label{fig4} 
\end{figure}

PoseNet is the first work that could directly regress the 6DoF pose from single images. Methods based on PoseNet have a similar fashion, which can be expressed as "encoder, localizer and regressor", which is shown in Figure \ref{fig4}.

\paragraph{Problem modelling} ~{}

In this section, we introduce the pose regression method through a single monocular image. The whole pipeline is input – network – output, which could directly estimate the pose relating to the camera of the capturing image. An RGB image $I_c$ captured by camera $C$ is the input of the CNN, after extracting features and regressors, the displacement between C and origin point can be expressed by a position vector $x\in R^3$, and the orientation vector $q\in R^4$ in quaternion form, after the orientation normalization into 3 dimensions, the pose emerges into a pose vector $p\in R^6$, which is shown as $p=[x,q]$.
\paragraph{Methods} ~{}

\begin{enumerate}[(1)]
\item Fixed Euclidean loss parameters
\end{enumerate}

Fig 1 shows the typical architecture of deep absolute pose regression, which uses a single image as the input. The output is a global pose result including position and orientation. This kind of method extracts high dimensional features through single images, then outputs these features with pose is expressed in a linear fashion as a 6-dimensional vector.

As mentioned, PoseNet \cite{ref7.2} is the first work to regress camera pose from single RGB images by training convolutional neural networks (CNNs), which does not rely on separate mechanisms or cross-frames/key frames to estimate pose. PoseNet shows robustness against the SIFT-based SfM (Structure from motion) method, the latter fails sharply after decreasing the training samples to a certain threshold.

To advance PoseNet, methods that use a single image as the input to improve the localization performance and modify the network, or update the loss function, have been proposed. Methods that use a fixed loss all share the same strategy. They learn location and orientation simultaneously using a stochastic gradient descent using the following objective loss function:
\begin{equation}
    l = ||\hat{x}-x||_2 + \beta ||\hat{q}-\frac{q}{||q||}||_2
\end{equation}

$\beta$ is a scaling factor to balance the value from any position error and orientation error. Euclidean loss attempts to learn the position and normalized quaternion difference.  

To improve the localization performance and to understand the model uncertainty, a Bayesian CNN \cite{ref7.3} with Bernoulli distributions has been proposed. The main contribution of the Bayesian CNN is in extending PoseNet to a Bayesian model which can determine the uncertainty of localization. To implement this, dropout layers were added after the sub-net and final output layer to get the stochastic pose samples The evaluation showed that there is a strong correlation between the uncertainty estimation and location error, so uncertainty can be used to predict the location error. This improves PoseNet’s relocalization accuracy for indoor and outdoor scenes. 

As mentioned, PoseNet has a 2048-dimensional fully connected (FC) layer, thus, this enables a Long-Short-Term-Memory (LSTM) layer model to reduce the feature dimensionality and to improve location accuracy \cite{ref7.4, ref7.8}. Walch et al. \cite{ref7.4} suggested making use of Long-Short Term Memory (LSTM) units on the PoseNet FC output, which performs a structured dimensionality reduction and chooses the most useful feature correlations for the task of pose estimation. Four LSTM units are used in the up, down, left and right directions respectively. This method outperforms PoseNet by almost 30\% for the positional error and 55\% for the orientation error. LSTM is also applied to temporally improve localization accuracy using image sequences, which will be introduced in the next part.

To further improve the accuracy of localization, an hourglass network was proposed to add another part to encode the rich and comprehensive information from coarse object structures and a second part to recover the fine-grained object details. 

Sharing the same thoughts of leveraging machine learning for camera localization, SVS PoseNet \cite{ref7.1} proposed a new network-based upon a classification network while using the same parameters rather than using hyperparameters optimization for each training dataset, which achieves a better performance for an outdoor dataset, e.g., Cambridge dataset \cite{ref7.2}).

The orientation expression in PoseNet is not unique, while the training strategy for orientation and translation is separately optimised. Furthermore, it’s a higher time cost to compute sparse frames. To tackle these problems, BranchNet created a new two-branch network that simultaneously learns the orientation and translation representations to effectively reduce the sparsity of sampled poses.

The methods above, enhance the origin architecture using the following extensions. The Hourglass PoseNet \cite{ref7.5} overall network consists of three components named encoder, decoder and regressor. It uses a modified ResNet34 as encoder-decoder, which can be considered as the encoder part in the whole pipeline (Fig 1). SVS PoseNet uses VGG16 with two additional FC layers for independent orientation and position prediction. SVS PoseNet also proposed data augmentation in 3D space through synthetic viewpoint generation. While BranchNet \cite{ref7.6} uses a different fashion from PoseNet. That is, orientation and translation vectors are predicted by two different branches after the $5^{th}$ Inception module. All of this work uses the same loss function as PoseNet.

However, problems arise when setting the balancing factor $\beta$. Because the loss function uses a joint loss, which needs careful tuning especially in a distinct scene. Otherwise, the uncertainty of network output will increase greatly. To address this problem, work that uses learnable pose loss function parameters has emerged.
\begin{enumerate}[(2)]
\item Learnable pose loss parameters
\end{enumerate}

To enhance the localization performance, Geometric PoseNet \cite{ref7.7} proposed learnable weights pose loss to balance the performance and improve the stability. Thus, compared to PoseNet, this method can keep the scalability and robustness while it doesn’t need to adjust the fixed balance factor hyperparameters in the loss function.

To learn an object’s position and orientation information from an image, the fixed Euclidean loss applies balanced hyperparameters, which independently learns these two components, but it is costly to learn the weights of these. By learning the estimate of the homoscedastic task uncertainty \cite{ref7.3} during training to represent uncertainty regularization terms and the residual regressions to represent the regression performance, the loss could be mutually constrained.
\begin{equation}
    l_\sigma(I) = l_x(I)\hat{\sigma}_x^{-2}+log\hat{\sigma} ^2 +l_q(I)\hat{\sigma}^{-2}_q+log\hat{\sigma}_q^2
\end{equation}

Replacing $\hat{S}:=log\hat{\sigma}^2$, the final form of the learnable loss function is:
\begin{equation}
    l_\sigma(I)=l_x(I)exp(-\hat{s}_x)+\hat{s}_x+l_q(I)exp(-\hat{s}_q)+\hat{s}_q
\end{equation}

Furthermore, different methods can be used to apply a learnable geometric loss \cite{ref7.7} function to obtain geometry constraints while adding other modules or functionality as follows. AtLoc \cite{ref7.8} adds an attention module before determining the regression coordinates to force the network to concentrate on the main part of the input images, which is a unique, static and stable area. In addition, AtLoc utilities ResNet34 as the encoder network which when pre-trained on the ImageNet dataset, finally regresses the 2048-dimensional full connected (FC) layer of PoseNet, like AtLoc, AdPR adds a discriminator network and adversarial learning. This not only regresses the pose but could also refine the pose. When extracting features, AdPR \cite{ref7.9} applies the ResNet-18 Network, as it can achieve the best performance when compared with VGG16 and AlexNet. APANet \cite{ref7.11} also employs an adversarial network to generate related images to the input image to better estimate the camera pose. PVL \cite{ref7.10} adopted a prior-guided dropout mask to avoid the influence of uncertainty of dynamic objects in dynamic environments. A dropout module is added before the feature extractor encoder to output multiple uncertainty possibilities, which could improve the pose robustness under challenging conditions, e.g., illimitation, viewpoint changes. After extraction, the self-attention module is added to reweight the feature map.

Furthermore, different methods can be used to apply a learnable geometric loss \cite{ref7.7} function to obtain geometry constraints while adding other modules or functionality as follows. AtLoc \cite{ref7.8} adds an attention module before determining the regression coordinates to force the network to concentrate on the main part of the input images, which is a unique, static and stable area. In addition, AtLoc utilities ResNet34 as the encoder network which when pre-trained on the ImageNet dataset, finally regresses the 2048-dimensional full connected (FC) layer of PoseNet, like AtLoc, AdPR adds a discriminator network and adversarial learning. This not only regresses the pose but could also refine the pose. When extracting features, AdPR \cite{ref7.9} applies the ResNet-18 Network, as it can achieve the best performance when compared with VGG16 and AlexNet. APANet \cite{ref7.11} also employs an adversarial network to generate related images to the input image to better estimate the camera pose. PVL \cite{ref7.10} adopted a prior-guided dropout mask to avoid the influence of uncertainty of dynamic objects in dynamic environments. A dropout module is added before the feature extractor encoder to output multiple uncertainty possibilities, which could improve the pose robustness under challenging conditions, e.g., illimitation, viewpoint changes. After extraction, the self-attention module is added to reweight the feature map.

Another method to improve the localization performance is to synthetically generate training data, SPP-Net \cite{ref7.12} shows a novel DNN architecture based on Spatial Pyramid max-pooling units, which also share the same loss function as geo.PoseNet \cite{ref7.7}.

\begin{enumerate}[(3)]
\item Other loss methods to enhance localization
\end{enumerate}

Neither using fixed-parameter loss nor using learnable loss functions, GeoPoseNet \cite{ref7.13} and GPoseNet \cite{ref7.7} consider other modules to enhance localization, GeoPoseNet proposed the reprojection loss to learn the mean of all residuals from points $g_i\in g^{\prime}$, which describes the reprojection error of scene geometry. $l_g(I)=\frac{1}{g^\prime}\sum_{g_i\in g^{\prime}}||\pi(x,q,g_i)-\pi(\hat{x},\hat{q},\hat{g_i})||_\gamma$, where $\gamma$ represents the normalization operation, $\pi$ maps a 3-D point $g^{\prime}$ to 2-D image coordinates $\binom{u}{v}$, i.e., $\pi(x,q,g)\mapsto \binom{u}{v}$. The reprojection loss transfers the jointly learnable loss to an image coordinates difference, which could vary the weighting between position and orientation, according to the different scenes during the model training.

GPoseNet \cite{ref7.13} builds a novel model by adding 2 Stochastic Variational Inference Gaussian Process Regressions (SVI GPs) regressors after the fully connected layer to learn the probability distribution of the output pose and to reduce the hyperparameter usage. The loss function of GPoseNet, which combines the SVI GPs loss using variational lower bound of two log marginal likelihoods $L_svi$ and CNN loss with the hyperparameter $\beta_{n_t}$ and $\beta_{n_q}$ of PoseNet, in which the hyperparameters $\beta_{g_t}$ and $\beta_{g_q}$ are set to be equal in the experiments \cite{ref7.13}, is as follows: 
\begin{equation}
    l=\beta_{g_t}l_{svi}(s_t,S_t,Z_t)+\beta_{g_q}l_{svi}(m_q,S_q,Z_q)+\beta_{n_t}||\hat{t}-t||_2+\beta_{n_q}||\hat{q}-q||_2
\end{equation}
\paragraph{Critical thinking} ~{}

\begin{enumerate}[(1)]
\item How do the methods change the networks?
\end{enumerate}
\begin{table}[htbp]
  \centering
  \caption{Network architecture comparison of APR methods through single images}
  \resizebox{\textwidth}{!}{
    \begin{tabular}{|p{4em}|p{9em}|p{10.085em}|p{11.085em}|}
    \hline
    \multicolumn{1}{|l|}{Loss Type} & Method & Encoder & Localizer \bigstrut\\
    \hline
    \multicolumn{1}{|l|}{\multirow{6}[12]{*}{\tabincell{l}{Fixed\\loss}}} & Posenet \cite{ref7.2} & GoogLeNet (3 softmax layer + fc layers$\rightarrow$3 regressor layer) & 1FC \bigstrut\\
\cline{2-4}      & Bayesian PoseNet \cite{ref7.3} & GoogLeNet (add dropout after 9th Icp) & 1FC+dropout \bigstrut\\
\cline{2-4}      & LSTM \cite{ref7.5} & GoogLeNet & 4LSTM+1FC \bigstrut\\
\cline{2-4}      & Hourglass PoseNet \cite{ref7.5} & ResNet34 Encoder-Decoder & 3FC \bigstrut\\
\cline{2-4}      & SVS PoseNet \cite{ref7.1} & VGG16 (conv layers) & 3FC \bigstrut\\
\cline{2-4}      & BranchNet \cite{ref7.6} & GoogLeNet (truncated after the 5th Icp) & 2 x [GoogLeNet (6th-9th Icp) + 1 FC] \bigstrut\\
    \hline
    \multicolumn{1}{|l|}{\multirow{6}[12]{*}{\tabincell{l}{Learnable\\Loss}}} & Geo.PoseNet \cite{ref7.7} & GoogLeNet & 1FC+1normalisation layer for orientation to unit length \bigstrut\\
\cline{2-4}      & AtLoc \cite{ref7.8} & ResNet-34 & 1FC \bigstrut\\
\cline{2-4}      & AdPR \cite{ref7.9} & ResNet-18 & 1FC \bigstrut\\
\cline{2-4}      & PVL \cite{ref7.10} & Prior Guided Dropout + Resnet34 & Composite Self-Attention + 1FC \bigstrut\\
\cline{2-4}      & APANet \cite{ref7.11} & ResNet-34 & 1FC \bigstrut\\
\cline{2-4}      & SPPNet \cite{ref7.12} & 3× (4 layers of 1×1 convolutions) + Spatial Pyramid max-pooling units & 3FC \bigstrut\\
    \hline
    \multicolumn{1}{|l|}{\multirow{2}[4]{*}{\tabincell{l}{Other\\Loss}}} & Geo.PoseNet (reprojection error loss) \cite{ref7.7} & GoogLeNet & 1FC+1normalisation layer for orientation to unit length \bigstrut\\
\cline{2-4}      & GPoseNet \cite{ref7.13} & GoogLeNet & 1FC + 2 SVI GP regressors \bigstrut\\
    \hline
    \end{tabular}}%
  \label{tab2}%
\end{table}%

The absolute pose regression methods using a single image, improve the algorithm performance by changing the network architecture of the traditional encoder-decoder-localizer module or adding some external parts to filter away information and add temporal information to better localize the camera pose. 

Table \ref{tab2} shows the network architectures of APR methods that use a single image in our survey. From which we can see that the modified GoogLeNet, ResNet34, ResNet-18 and VGG16 pre-trained with classification task (such as on ImageNet or Places datasets) are popular choices as the encoder of the APR network. A dropout layer is added in some methods (Bayesian and PVL) to help compute probability distribution or prior guided mask, while an LSTM module is applied to offer temporal information auxiliary. An attention module is also widely utilized to force the network to focus on geometrically robust objects. The localizer part in all the methods has at least a regressor to output the translation and orientation information, in which the 4-D quaternion is normalized to a unit length.
\begin{enumerate}[(2)]
\item How do methods change the loss function?
\end{enumerate}
\begin{table}[htbp]
  \centering
  \caption{Loss function and publication information comparison of APR methods thorough single image}
  \resizebox{\textwidth}{!}{
    \begin{tabular}{|p{2em}|p{10em}|p{8em}|p{14em}|}
    \hline
    \multicolumn{1}{|l|}{\tabincell{l}{Loss\\Type}} & Method & Year-Pub.-Cited & \multicolumn{1}{|l|}{Loss function} \bigstrut\\
    \hline
    \multicolumn{1}{|l|}{\multirow{6}[12]{*}{\tabincell{l}{Fixed\\Loss}}} & PoseNet \cite{ref7.2} & 2015-ICCV-1146 & \multirow{6}[12]{*}{$l = ||\hat{x}-x||_2 + \beta ||\hat{q}-\frac{q}{||q||}||_2$} \bigstrut\\
\cline{2-3}      & Bayesian PoseNet \cite{ref7.3} & 2016-ICRA-330 &  \bigstrut\\
\cline{2-3}      & LSTM PoseNet \cite{ref7.5} & 2017-ICCV-273 &  \bigstrut\\
\cline{2-3}      & Hourglass PoseNet \cite{ref7.5} & 2017-ICCVW-80 &  \bigstrut\\
\cline{2-3}      & SVS PoseNet \cite{ref7.1} & 2017-IROS-56 &  \bigstrut\\
\cline{2-3}      & BranchNet \cite{ref7.6} & 2017-ICRA-62 &  \bigstrut\\
    \hline
    \multicolumn{1}{|l|}{\multirow{6}[12]{*}{\tabincell{l}{Learnable\\Loss}}} & Geo.PoseNet \cite{ref7.7} & 2017-CVPR-384 & \multirow{6}[12]{*}{\tabincell{l}{$l_\sigma(I)=l_x(I)exp(-\hat{s}_x)+\hat{s}_x$\\$+l_q(I)exp(-\hat{s}_q)+\hat{s}_q$}} \bigstrut\\
\cline{2-3}      & AtLoc \cite{ref7.8} & 2019-AAAI-6 &  \bigstrut\\
\cline{2-3}      & AdPR \cite{ref7.9} & 2019-ICCVW-5 &  \bigstrut\\
\cline{2-3}      & PVL \cite{ref7.10} & 2019-ICCV-9 &  \bigstrut\\
\cline{2-3}      & APANet \cite{ref7.11} & 2020-ECCVW-0 &  \bigstrut\\
\cline{2-3}      & SPPNet \cite{ref7.12} & 2018-BMVC-6 &  \bigstrut\\
    \hline
    \multicolumn{1}{|l|}{\multirow{2}[4]{*}{\tabincell{c}{Other\\Loss}}} & Geo.PoseNet (reprojection error loss) \cite{ref7.7} & 2017-CVPR-384 & $l_g(I)=\frac{1}{g^\prime}\sum_{g_i\in g^{\prime}}||\pi(x,q,g_i)-\pi(\hat{x},\hat{q},\hat{g_i})||_\gamma$ \bigstrut\\
\cline{2-4}      & GPoseNet \cite{ref7.13} & 2018-BMVC-18 & $l=\beta_{g_t}l_{svi}(s_t,S_t,Z_t)+\beta_{g_q}l_{svi}(m_q,S_q,Z_q)+\beta_{n_t}||\hat{t}-t||_2+\beta_{n_q}||\hat{q}-q||_2$ \bigstrut\\
    \hline
    \end{tabular}}%
  \label{tab3}%
\end{table}%

The research of applying the loss function tends to be automatic, hyperparameter-free and more informative to reduce the use of empirical fixed parameters. A fixed loss function computes the translation and orientation sum using a balance factor to balance different weighted items, which requires a long time to optimize the loss of the training data. Later a learnable loss \cite{ref7.7} was proposed by adding homoscedastic uncertainty to automatically balance the translation and orientation loss, which avoids using hyperparameters and surpasses the performance of the fixed loss methods. Apart from fixed loss and learnable loss methods, other methods proposed the use of reprojection error loss and GPoseNet loss to add other information formats. e.g., a probability distribution of the output pose, to improve the loss function. Table \ref{tab3} compares the loss and publication information of APR methods.
\subsubsection{Absolute pose regression through image sequences auxiliary}

Another method to regress absolute pose is for auxiliary learners to use image sequences. Auxiliary learning refers to the combination of using absolute pose regression and auxiliary task constraints (e.g., visual odometry). Loss functions in auxiliary learning methods usually consist of APR loss and auxiliary tasks loss. All the above-mentioned methods can be used to get the absolute camera pose. They may even use relative pose regression loss in the pipeline.

\paragraph{Problem modelling} ~{}

Traditional structure-based methods still have advantages over deep neural networks. Therefore, absolute pose regression with images sequences using auxiliary learning, has been proposed. Unlike methods using single images, auxiliary learning through image pairs, typically learn the absolute pose by firstly estimating the relative pose with which auxiliary constraints. This could involve globally consistent pose predictions to improve the localization performance, i.e., to reduce positioning error and to improve positioning robustness. 

\paragraph{Methods} ~{}

MapNet \cite{ref8.1} proposed to add an additional loss term from image pairs as a geometric constraint, which could significantly enhance the localization ability. Other methods share the same intuition as MapNet using auxiliary learning, which minimizes the combination of the per-image absolute pose loss and the relative pose loss between image pairs with a weight coefficient factor $\alpha$. The loss function is shown as: 
\begin{equation}
    l(I_{total})=l(I_i)+\alpha\sum_{i\ne j}loss(I_{ij})
\end{equation}

Where $loss(I_{ij})$ means the relative camera pose $p_i$ and $p_j$ between image pairs $I_i$ and $I_j$, which is computed by the learnable loss function (equation (7)).

In addition, MapNet also transforms the quaternion value to the logarithm of the quaternion, which presents a 3DoF rotation with 3 dimensions that is not over-parameterized. $log q$ is defined below, where $u$ and $v$ each represent the real and imaginary part of a unit quaternion:
\begin{equation}
    log q= \begin{cases}
    \frac{v}{||v||}cos^{-1}u, & \text{if $||v||$ $\ne$ 0} \\
    0, & \text{otherwise}
    \end{cases}
\end{equation}

Xue et al. \cite{ref8.2} follow a similar notion to regress global camera pose through spatial-temporal constraints, in which local features enhance global localization, named Local Supports Global (LSG). Furthermore, LSG proposed the use of a content-augmented valuation to estimate pose uncertainty and motion-based refinement, to optimize pose prediction via motion constraints. LSG employs a global pose loss $L_g$ from absolute regression, visual odometry loss $L_{vo}$, geometric constraint and joint loss $L_{joint}$ motion constraint that together optimize the pose regression as follows.
\begin{equation}
    l_{total}=l_g+l_{vo}+l_{joint}
\end{equation}

Where the motion-based joint loss is coupled as the global and local pose from the local window: 
\begin{equation}
    l_{joint}=\sum_{i=1}^ND(P_{i+1},P_{i+1}^{vo},P_i)
\end{equation}

VlocNet \cite{ref8.3} also simultaneously learns the visual odometry as an auxiliary task to regress the global pose with two sub-networks. Geometric consistency loss is adapted to minimize the pose error, which is defined as:
\begin{equation}
    l(I_{total})=(I_{i_x}+I_{{ij}_x})exp(-\hat{s}_x)+(I_{i_q}+I_{ij}^q)exp(-\hat{s}_q)+\hat{s}_q
\end{equation}

VlocNet++ \cite{ref8.4} introduces semantic knowledge to pose regression, which fuses the geometric-temporal information with semantic features together. The loss of VlocNet++ combines global pose regression, visual odometry loss and the cross-entropy loss for semantic segmentation loss together, with three factors $\hat{s}_{loc}$, $\hat{s}_{vo}$, $\hat{s}_{seg}$ to balance the three terms.
\begin{equation}
    l(I_{total})=l_{loc}exp(-\hat{s}_{loc})+\hat{s}_{loc}+l_{vo}exp(-\hat{s}_{vo})+\hat{s}_{vo}+l_{seg}exp(-\hat{s}_{seg})+\hat{s}_{seg}
\end{equation}

As an extension of AtLoc, AtLocPlus \cite{ref7.8} also incorporates temporal constraints to simultaneously learn the absolute pose loss and the relative pose loss, which leads to a better performance than AtLoc using a single image input. AtLocPlus shares the same loss function with MapNet.

DGRNet \cite{ref8.5} proposed a novel architecture with relative pose regression sub-network RCNN1 and global pose regression sub-network RCNN2 and fully connected fusion layer FCFL to extract features through images. Cross transformation constraints (CTC) and Mean square error (MSE) are applied to the loss function to improve the regression performance. DGRNet jointly uses the global and relative loss with the CTC functions $\hat{l}_i$ and the ground truth $\hat{P}_i$ in frame $i$ as follows: 
\begin{equation}
    w=\mathop{argmin}\limits_w\frac{1}{N}^N_{i=1}{\sum_{k=0}^6(l_k^i)+\sum_{j=0}^4||P^ij-\hat{P}^ij||_2^2}
\end{equation}

\paragraph{Critical thinking} ~{}

\begin{enumerate}[(1)]
\item How methods add constraints multi-task?
\end{enumerate}
\begin{table}[htbp]
  \centering
  \caption{Constraints comparison with multi-tasks of APR methods thorough image sequences}
  \resizebox{\textwidth}{!}{
    \begin{tabular}{|p{7em}|p{5em}|p{5em}|p{4em}|p{5em}|p{6em}|}
    \hline
    \multirow{2}[4]{*}{Methods} & \multicolumn{3}{c|}{Output} & \multirow{2}[4]{*}{\tabincell{l}{Geometric\\-aware\\temporal\\constraints}} & \multirow{2}[4]{*}{\tabincell{l}{Other\\constraints}} \bigstrut\\
\cline{2-4}    \multicolumn{1}{|c|}{} & Localization & Visual odometry & Semantic segmentation & \multicolumn{1}{l|}{} & \multicolumn{1}{l|}{} \bigstrut\\
    \hline
    MapNet \cite{ref8.1} & $\checkmark$ & $\checkmark$ & \multicolumn{1}{l|}{} & $\checkmark$ & / \bigstrut\\
    \hline
    LSG \cite{ref8.2} & $\checkmark$ & $\checkmark$ & \multicolumn{1}{l|}{} & $\checkmark$ & Motion-based constraints \bigstrut\\
    \hline
    VlocNet \cite{ref8.3} & $\checkmark$ & $\checkmark$ & \multicolumn{1}{l|}{} & $\checkmark$ & / \bigstrut\\
    \hline
    VlocNet++ \cite{ref8.4} & $\checkmark$ & $\checkmark$ & \multicolumn{1}{l|}{} & \multicolumn{1}{l|}{} & Semantic constraints \bigstrut\\
    \hline
    AtLocPlus \cite{ref7.8} & $\checkmark$ & $\checkmark$ & \multicolumn{1}{l|}{} & $\checkmark$ & / \bigstrut\\
    \hline
    DGRNet \cite{ref8.5} & $\checkmark$ & $\checkmark$ & $\checkmark$ & $\checkmark$ & / \bigstrut\\
    \hline
    \end{tabular}}%
  \label{tab4}%
\end{table}%

With image sequences and adding geometric-aware temporal constraints or other constraints, methods can not only obtain the localization result but also obtain visual odometry information. Furthermore, DGRNet could also get the semantic segmentation results through the network. Table \ref{tab4} shows a general comparison of what output and constraints image sequences auxiliary-based APR methods apply.

\begin{enumerate}[(2)]
\item How do the methods improve the network and loss function?
\end{enumerate}
\begin{table}[htbp]
  \centering
  \caption{Loss function and publication information comparison of APR methods thorough image sequences}
  \resizebox{\textwidth}{!}{
    \begin{tabular}{|l|l|l|l|}
    \hline
    Methods & Year-pub-cited & Encoder & Loss Function \bigstrut\\
    \hline
    MapNet \cite{ref8.1} & 2018-CVPR-155 & \tabincell{l}{ResNet34+global\\average pooling} & \tabincell{l}{$l(I_{total})=l(I_i)$\\$+\alpha\sum_{i\ne j}loss(I_{ij})$} \bigstrut\\
    \hline
    LSG \cite{ref8.2} & 2019-ICCV-11 & \tabincell{l}{ResNet34+ResBlock} & $l_{total}=l_g+l_{vo}+l_{joint}$ \bigstrut\\
    \hline
    VlocNet \cite{ref8.3} & 2018-ICRA-113 & \tabincell{l}{ResNet50\\(ReLUs$\rightarrow$ELUs)} & \tabincell{l}{$ l(I_{total})=(I_{i_x}+I_{{ij}_x})exp(-\hat{s}_x)$\\$+(I_{i_q}+I_{ij}^q)exp(-\hat{s}_q)+\hat{s}_q$} \bigstrut\\
    \hline
    VlocNet++ \cite{ref8.4} & 2018- RA-L-105 & \tabincell{l}{ResNet50\\(ReLUs$\rightarrow$ELUs)\\+ global\\average pooling} & \tabincell{c}{$l(I_{total})=l_{loc}exp(-\hat{s}_{loc})+\hat{s}_{loc}$\\$+l_{vo}exp(-\hat{s}_{vo})+\hat{s}_{vo}$\\$+l_{seg}exp(-\hat{s}_{seg})+\hat{s}_{seg}$} \bigstrut\\
    \hline
    AtLocPlus \cite{ref7.8} & 2019-AAAI-6 & ResNet-34 & \tabincell{l}{$l(I_{total})=l(I_i)$\\$+\alpha\sum_{i\ne j}loss(I_{ij})$} \bigstrut\\
    \hline
    DGRNet \cite{ref8.5} & 2019- PRICAI-16 & \tabincell{l}{ResNet-50\\(Res1 to Res4) \\(BN+ ELUs)} & \tabincell{l}{$w=\mathop{argmin}\limits_w\frac{1}{N}^N_{i=1}\sum_{k=0}^6(l_k^i)$\\$+\sum_{j=0}^4||P^ij-\hat{P}^ij||_2^2$} \bigstrut\\
    \hline
    \end{tabular}}%
  \label{tab5}%
\end{table}%

ResNet-34 and ResNet-50 with the modification are widely used to extract features in an image sequence regression network. MapNet, VlocNet and AtLocPlus utilize the joint absolute and relative pose loss to improve the regression. LSG applies a motion-based constrain to the loss function while VlocNet++ adds the semantic constraint into the loss function. DGRNet combines both CTC and MSE in the loss computation. Table \ref{tab5} generally lists the publication information, neural network encoder and loss function of image sequences auxiliary-based APR methods.

\subsubsection{Absolute pose regression through videos}

Without using single images or image pairs to regress camera pose, video clips could be used to add a temporal smoothness constraint to pose regression. 

\paragraph{Problem modelling} ~{}

Videos and other sensor data can be easily accessed by mobile devices. Videos can be synchronised using temporal information to other input data such as visual odometry, Inertial Measurement Unit (IMU) sensors such as accelerometer and gyroscope and GNSS data, by aligning timestamps. Sharing a similar pipeline to single image-based and image sequence-based ARP methods, video-based APR methods also regress the translation and orientation through the CNN feature extractor and localizer regressor, which will also contain other auxiliary information as with videos. 

\paragraph{Methods} ~{}

VidLoc \cite{ref8.6} proposed a CNN-RNN based model to regress camera pose which could smooth the pose estimation from image or video input. The network is formed by using GoogLeNet Inception \cite{ref8.7} without using fully connected layers to extract image features, and a bidirectional LSTM module to model temporal information with memory cells and several gates. Adopting a LSTM network with a bidirectional model could use two hidden states to process this, forwards and backwards. This could also be concatenated with a single hidden state to get the camera pose. The network loss of VidLoc is computed by a weighted sum of translation and orientation error from the output of LSTM as follows. 
\begin{equation}
    l=\sum_{t=1}^T\alpha_1||x_t-\hat{x}_t||+\alpha_2||q_t-\hat{q}_t||
\end{equation}

where $\gamma_t=[x_t,q_t]$ and $\hat{\gamma}_t=[\hat{x}_t,\hat{q}_t]$ separately represent the ground truth and prediction value for the camera pose translation and orientation values.

MapNet+ \cite{ref8.1} and MapNet+PGO \cite{ref8.1} share the same network architecture with MapNet that extracts features through ResNet34 and uses a global average pooling layer. Not only using the absolute pose loss, VidLoc, the visual odometry loss is also computed to improve estimation quality in MapNet. The method also integrates IMU and GNSS data to help improve pose regression. This fuses the labeled data and unlabeled data from VO or sensors for self-supervised learning and demonstrates a better performance under challenging conditions, e.g., appearance changes.
\begin{equation}
    l=l_{labelled\ data}+l_{unlabelled\ data}
\end{equation}

Where the unlabeled data loss could be computed through combining relative camera pose $v_ij$ and the visual odometry $\hat{v}_{ij}$, or through other sensors, such as IMU and GNSS. 

MapNet+PGO \cite{ref8.1} could further improve the performance whilst minimizing the computation cost by using pose graph optimization (PGO) to fuse the pose from MapNet+ and the visual odometry. 
\begin{equation}
    l_{PGO}({p_i^0}_{i=1}^T)=\sum_{i=1}^T\bar{h}(p_i^o,p_i)+\sum_{i,i=1,i\ne j}^T\bar{h}(v_{ij}^o,\hat{v}_{ij})
\end{equation}

\paragraph{Critical thinking} ~{}

\begin{table}[htbp]
  \centering
  \caption{Loss function and publication information comparison of APR methods thorough video}
  \resizebox{\textwidth}{!}{
    \begin{tabular}{|l|c|c|c|}
    \hline
    Type & VidLoc \cite{ref8.6} & MapNet+ \cite{ref8.1} & MapNet+PGO \cite{ref8.1} \bigstrut\\
    \hline
    Publication & 2017-CVPR & \multicolumn{2}{c|}{2018-CVPR} \bigstrut\\
    \hline
    Cited & 163 & \multicolumn{2}{c|}{155} \bigstrut\\
    \hline
    Input & Videos & \multicolumn{2}{c|}{unlabeled videos+vo+imu+GNSS} \bigstrut\\
    \hline
    Fusion ability & / & $\checkmark$ & $\checkmark$ \bigstrut\\
    \hline
    Loss function & \tabincell{l}{$l=\sum_{t=1}^T\alpha_1||x_t-\hat{x}_t||$\\$+\alpha_2||q_t-\hat{q}_t||$} & \tabincell{l}{$l=l_{labelled\ data}$\\$+l_{unlabelled\ data}$} & \tabincell{l}{$l_{PGO}({p_i^0}_{i=1}^T)=$\\$\sum_{i=1}^T\bar{h}(p_i^o,p_i)$\\$+\sum_{i,i=1,i\ne j}^T$\\$\bar{h}(v_{ij}^o,\hat{v}_{ij})$} \bigstrut\\
    \hline
    \multirow{2}[2]{*}{Feature extraction} & \multirow{2}[2]{*}{GoogLeNet Inception} & \multicolumn{2}{l|}{\multirow{2}[2]{*}{ResNet34 + global average pooling}} \bigstrut[t]\\
      &   & \multicolumn{2}{l|}{} \bigstrut[b]\\
    \hline
    Regressor & Bidirectional RNN+1 FC & \multicolumn{2}{c|}{1FC} \bigstrut\\
    \hline
    \end{tabular}}%
  \label{tab6}%
\end{table}%

VidLoc, MapNet and MapNet+PGO use videos as input, while some of these fuse unlabeled data to help improve supervised learning. VidLoc adds a bidirectional RNN to regress camera 6DoF pose whilst outputting the probabilistic of pose estimation. MapNet+ and MapNet+PGO mainly utilize visual odometry into the loss function to optimize the regression performance. Table \ref{tab6} gives a general comparison of video-based APR methods, including publication, loss function, neural network main architectures, etc.

\subsubsection{Summary}

In this part, we discussed the work of 2D-to-2D absolute pose (consisting of localization and orientation) regression using deep neural networks entirely and using no image queries. Recent work shows that the APR methods can suffer from less accuracy, and from overfitting, compared to structure-based methods. These could be used in scene-specific environments, with the emergence of relative pose regression-based localization methods. The training process could be generally used in multiple scenes \cite{ref2.2}.

\subsection{Relative camera pose regression }
\label{subsec5.2}
A direct absolute camera pose regression model learns the mapping from images of objects’ pixels to camera poses, which is decided by the coordinate system that the specific scenes are in. Thus, cross-scene learning brings a coordinates transfer that is bounded and that delivers learnable physical geometric knowledge. In contrast to scene-specific absolute pose regression, relative camera pose regression methods compute a reference image’s relative pose and are trained on general multi-unseen scenes to increase the scalability in an end-to-end manner.

\subsubsection{Relative camera pose regression through explicit retrieval}

Relative camera pose regression could be calculated through the prior image retrieval process which computes the most similar image relative to the query image in a database and then predicts the relative pose between them and finally gets the absolute pose of the query image.

\paragraph{Problem modelling} ~{}

Given an image $I_c^a$ captured by camera $c$, its relative nearest similar image $I_c^b$ could be estimated through the image retrieval method in a database. After getting the ground truth pose $p_b$ of $I_c^b$ and relative pose $p_{a\rightarrow b}$ between $I_c^a$ and $I_c^b$, the absolute pose $p_a$ of $I_c^a$ could be defined by a mathematical transformation.
\paragraph{Methods} ~{}

NNnet \cite{ref9.1} first proposed an image retrieval based relative pose regression method. The input of the method is a query image and an image database including ground truth poses. A set of image pairs is utilized to regress the relative pose through a Siamese network with two modified ResNet34 branches with a fixed loss function. The nearest neighbor image to the query image could be computed through a feature extractor formed by the network branch, then the relative pose and neighbor’s ground truth pose could be fused to get the absolute pose of the query image.

RelocNet \cite{ref9.2} furtherly modifies NNnet \cite{ref9.1} with continuous metric learning to learn global image features with a camera frustum to improve the result, while a geometric relative pose loss is also applied. Relative pose loss learns the differential pose between two pose matrices using a representation of matrix for rotation and translation. The training loss in which, frustum loss learns the image pair overlaps is defined as follows. 
\begin{equation}
    l=\alpha l_{SE}(3)+\beta l_{frustum}
\end{equation}

To tackle the bottleneck in previous retrieval-based relative regression methods whose performance is limited because they use the same features for retrieval and regression modules, CamNet \cite{ref9.3} proposed a novel pipeline split into three steps. Coarse-retrieval, fine-retrieval, relative pose regression is used, which is based on a Siamese architecture with three branches for each of the three steps. This coarse-to-fine framework improves regression accuracy and scalability. The loss function of CamNet is based on RelocNet, which is shown as follows.
\begin{equation}
    l=l_{frustum}+l_{angle}+l_{triplet}+l_{PFR}+l_{PRP}
\end{equation}

Zhou et al. \cite{ref9.4} analyze the previous image retrieval based relative pose regression method and propose a novel framework with essential matrices and modified RANSAC for computing the absolute pose. A Siamese modified ResNet34 network with a fixed matching layer (EssNet) and a Neighborhood Consensus matching layer (NC-EssNet) is learned to produce a matching score map for a further regression, essential matrix. The loss function optimizes the Euclidean distance between the essential matrix with two 9D vectors (where a 3 × 3 matrix becomes one 9D vector)
\begin{equation}
    l_{ess}(E^*,E)=||e-e^*||_2
\end{equation}

\subsubsection{Relative camera pose regression through implicit CNN}

To avoid large collection for database and long test time consuming, some methods try to regress relative camera pose through an implicit neural network.

Relative NN \cite{ref9.5} proposed an end-to-end method to regress the relative pose between two cameras with two images as input. A Siamese Hybrid-CNN with a pre-trained AlexNet network consisting of two branches is used for regression with the fixed Euclidean loss, which has a good performance on the Technical University of Denmark Robot Image Dataset (DTU dataset) \cite{ref9.10}.

AnchorNet \cite{ref9.6} addresses the localization problem by defining anchor points as the visible landmark to learn the query image’s relative anchors and its offset. The multi-task model includes classifying the query image to which specific anchor points. and finding the offsets compared to the classified anchor point, which forms the loss function. $\hat{C}$, $X$, and $Y$ represent the classification output and the ground truth offsets.
\begin{equation}
    l=\mathop{\sum}\limits_i[(X_i-\hat{X}_i)^2+(Y_i-\hat{Y}_i)^2]\hat{C}^i
\end{equation}

\subsubsection{Summary}

\begin{table}[htbp]
  \centering
  \caption{Loss function and publication information comparison of RPR methods}
  \resizebox{\textwidth}{!}{
    \begin{tabular}{|p{1.5em}|p{8em}|p{7.5em}|p{12em}|}
    \hline
    \multicolumn{1}{|p{1.5em}|}{Type} & Method & Year-Pub-Cited & Loss function \bigstrut\\
    \hline
    \multicolumn{1}{|l|}{\multirow{4}[8]{*}{\tabincell{l}{Through\\retrieval}}} & NNnet \cite{ref9.1} & 2017-ICCVW-68 & $l = ||\hat{x}-x||_2 + \beta ||\hat{q}-\frac{q}{||q||}||_2$ \bigstrut\\
\cline{2-4}      & RelocNet \cite{ref9.2} & 2018-ECCV-70 & $l=\alpha l_{SE}(3)+\beta l_{frustum}$ \bigstrut\\
\cline{2-4}      & Camnet \cite{ref9.3} & 2019-ICCV-22 & $l=l_{frustum}+l_{angle}+l_{triplet}+l_{PFR}+l_{PRP}$ \bigstrut\\
\cline{2-4}      & \tabincell{l}{To learn or not\\to learn \cite{ref9.4}} & 2020-ICRA-16 & $l_{ess}(E^*,E)=||e-e^*||_2$ \bigstrut\\
    \hline
    \multicolumn{1}{|l|}{\multirow{2}[4]{*}{\tabincell{l}{Through\\CNN}}} & Relative NN \cite{ref9.5} & 2017-ACIVS-108 & $l = ||\hat{x}-x||_2 + \beta ||\hat{q}-\frac{q}{||q||}||_2$ \bigstrut\\
\cline{2-4}      & AnchorNet \cite{ref9.6} & 2018-BMVC-23 & $l=\mathop{\sum}\limits_i[(X_i-\hat{X}_i)^2+(Y_i-\hat{Y}_i)^2]\hat{C}^i$ \bigstrut\\
    \hline
    \end{tabular}}%
  \label{tab7}%
\end{table}%

To regress relative pose, retrieval-based methods utilize a multi-stage strategy to finally get the absolute pose with the retrieval step as fundamental to the process. CNN-based methods offer another way to regress relative pose implicitly within the network. Table \ref{tab7} summarizes the publication information and loss function of relative camera pose regression-based methods.

\section{Camera pose estimation comparisons}
\label{sec6}
We reviewed structure feature-based and regression-based pose estimation methods in section \ref{sec4} and section \ref{sec5}, in this section, we systematically compare the performance of the datasets that appear in these methods, the quantitative and qualitative results, and the real-world applicability of these methods.

\subsection{Comparison of datasets}
\label{subsec6.1}
\begin{table}[H]
  \centering
  \caption{An overview of some popular camera localization datasets}
  \resizebox{\textwidth}{!}{
    \begin{tabular}{|l|l|l|l|l|l|l|l|l|l|l|l|l|}
    \hline
    \tabincell{l}{Attri\\butes} & Affi. & Year & \tabincell{l}{Cit\\es} & \tabincell{l}{Plat\\form} & \tabincell{l}{Publi\\cation} & \tabincell{l}{Envi\\ron\\ment} & Scale & Imagery & \tabincell{l}{Sc\\en\\es} & \tabincell{l}{Train\\ima\\ges} & \tabincell{l}{Test\\ima\\ges} & \tabincell{l}{Ar\\ea} \bigstrut\\
    \hline
    \tabincell{l}{7Sce\\nes\\\cite{ref7.2}} & \tabincell{l}{Micr\\osoft} & 2015 & 1531 & Hand & ICCV & \tabincell{l}{Ind\\oor} & Room & \tabincell{l}{RBG-D\\sensor\\(Kinect)} & 7 & 26000 & 17000 & / \bigstrut\\
    \hline
    \tabincell{l}{Camb\\ridge\\\cite{ref7.2}} & \tabincell{l}{Camb\\ridge} & 2015 & 1531  & Hand & ICCV & \tabincell{l}{Hist\\oric\\city} & Street & \tabincell{l}{Mobile\\phone\\camera} & 6 & 8380 & 4841 & / \bigstrut\\
    \hline
    \tabincell{l}{Oxfo\\rd\\Robot\\Car\\\cite{ref10.2}} & \tabincell{l}{Oxfo\\rd} & 2016 & 620 & \tabincell{l}{Vehi\\cle} & IJRR & \tabincell{l}{Urb\\an} & Street & \tabincell{l}{Stereo\\\&mon\\ocular\\camera} & 11 & 20862 & 11934 & / \bigstrut\\
    \hline
    \tabincell{l}{TUM\\LSI\\\cite{ref7.4}} & TUM & 2017 & 296 & \tabincell{l}{NavV\\is\\M3} & ICCV & \tabincell{l}{Ind\\oor} & Room & \tabincell{l}{Mono\\cular\\camera} & 1 & 875 & 220 & \tabincell{l}{55\\75\\$m^2$} \bigstrut\\
    \hline
    \tabincell{l}{Dubr\\ovnik\\6K\\\cite{ref10.4}} & \tabincell{l}{Cor\\nell} & 2010 & 455 & Hand & ECCV & \tabincell{l}{Hist\\oric\\city} & \tabincell{l}{Small\\town} & \tabincell{l}{Inter\\net\\images} & 1 & 6044 & 800 & / \bigstrut\\
    \hline
    \tabincell{l}{Apo\\llo\\Scape\\\cite{ref10.5}} & Baidu & 2018 & 236 & \tabincell{l}{Vehi\\cle}  & CVPR & \tabincell{l}{Out\\door} & Street & \tabincell{l}{VMX-\\CS6\\camera\\system}  & 28 & 7481 & 7518 & / \bigstrut\\
    \hline
    \tabincell{l}{Aac\\hen\\\cite{ref3.5}} & \tabincell{l}{Aach\\en} & 2018  & 312  & \tabincell{l}{Vehi\\cle}  & CVPR  & \tabincell{l}{Hist\\oric\\city} & \tabincell{l}{Small\\town}  & \tabincell{l}{Mobile\\phone\\camera}  & 2  & 3047 & 369 & / \bigstrut\\
    \hline
    \tabincell{l}{CMU\\\cite{ref3.5}} & CMU & 2018  & 312  &  \tabincell{l}{Vehi\\cle} & CVPR  & \tabincell{l}{Urb\\an} &  Street & \tabincell{l}{2 cam\\eras}  & 10  & 7159 & 75335 & / \bigstrut\\
    \hline
    \tabincell{l}{InLoc\\\cite{ref6.31}} & \tabincell{l}{Tokyo\\Techn\\ology\\et al.} & 2018 & 136 & Hand & CVPR & \tabincell{l}{Ind\\oor} & Room & \tabincell{l}{Panor\\amic\\images} & 5 & 9972 & 356 & \tabincell{l}{18\\5.8\\$m^2$} \bigstrut\\
    \hline
    \end{tabular}}%
  \label{tab8}%
\end{table}%

\begin{table}[H]
  \centering
  \caption{Specific scenes information for 7Scenes and Cambridge datasets}
  \resizebox{\textwidth}{!}{
    \begin{tabular}{|l|l|l|l|l|l|}
    \hline
    Dataset & Scene & \tabincell{l}{Spatial extent(m)} & \tabincell{l}{Area\\or volume} & \tabincell{l}{Train\\frames} & \tabincell{l}{Test\\frames} \bigstrut\\
    \hline
    \multirow{8}[16]{*}{\tabincell{l}{7Scenes\\\cite{ref7.2}}} & All & 4x3m &  / & 26000 & 17000 \bigstrut\\
\cline{2-6}      & chess & 3x2x1$m$ & 6$m^2$ & 4000 & 2000 \bigstrut\\
\cline{2-6}      & fire & 2.5x1x1$m$ & 2.5$m^2$ & 2000 & 2000 \bigstrut\\
\cline{2-6}      & head & 2x0.5x1$m$ & 1$m^2$ & 1000 & 1000 \bigstrut\\
\cline{2-6}      & office & 2.5x2x1.5$m$ & 7.5$m^2$ & 6000 & 4000 \bigstrut\\
\cline{2-6}      & pumpkin & 2.5x2x1$m$ & 5$m^2$ & 4000 & 2000 \bigstrut\\
\cline{2-6}      & kitchen & 4x3x1.5$m$ & 18$m^2$ & 7000 & 5000 \bigstrut\\
\cline{2-6}      & stairs & 2.5x2x1.5$m$ & 7.5$m^2$ & 2000 & 1000 \bigstrut\\
    \hline
    \multirow{7}[14]{*}{\tabincell{l}{Cambridge\\\cite{ref7.2}}} & All & 100x500$m$ & /  & 8380 & 4841 \bigstrut\\
\cline{2-6}      & great court & /  & 8000$m^3$ &  1532 & 760 \bigstrut\\
\cline{2-6}      & k.college & 140x40$m$ & 5600$m^3$ & 1220 & 343 \bigstrut\\
\cline{2-6}      & street & 500x100$m$ & 50000$m^3$ & 3015 & 2923 \bigstrut\\
\cline{2-6}      & old hospital & 40x40$m$ & 2000$m^3$ & 895 & 182 \bigstrut\\
\cline{2-6}      & shop facade & 35x25$m$ & 875$m^3$ & 231 & 103 \bigstrut\\
\cline{2-6}      & st M.Church & 80x60$m$ & 4800$m^3$ & 1487 & 530 \bigstrut\\
    \hline
    \end{tabular}}%
  \label{tab9}%
\end{table}%

Large scale, multi-distribution, datasets that cover different collection platforms, environments, and imagery on challenging scenes, e.g., illimitation viewpoint, or appearance changes, are critical for evaluating advanced camera localization algorithms. Table \ref{tab8} summarizes the common datasets used for camera localization tasks, including 7Scenes, Cambridge, TUM LSI, etc. Table \ref{tab9} mainly introduces the two most important datasets of 7Scenes and Cambridge, which is popular for use in indoor and outdoor environment camera pose tests respectively.

\subsection{Comparison of published results on common benchmarks}
\label{subsec6.2}
\begin{table}[htbp]
  \centering
  \caption{A summary of published results of structure-based methods on the 7Scenes dataset}
  \resizebox{\textwidth}{!}{
    \begin{tabular}{|p{6em}|p{3em}|p{3em}|p{3em}|p{3em}|p{3.5em}|p{3em}|p{3em}|}
    \hline
    Method & Chess & Fire & Head & Office & Pumpkin & Kitchen & Stairs \bigstrut\\
    \hline
    \tabincell{l}{ScoRe\\Forest \cite{ref7.2}} & \tabincell{l}{0.03m,\\0.66°} & \tabincell{l}{0.05m,\\1.50°} & \tabincell{l}{0.06m,\\5.50°} & \tabincell{l}{0.04m,\\0.78°} & \tabincell{l}{0.04m,\\0.68°} & \tabincell{l}{0.04m,\\0.76°} & \tabincell{l}{0.32m,\\1.32°} \bigstrut\\
    \hline
    \cite{ref6.33} & \tabincell{l}{0.02m,\\0.5°} & \tabincell{l}{0.02m,\\0.9°} & \tabincell{l}{0.01m,\\0.8°} & \tabincell{l}{0.03m,\\0.7°} & \tabincell{l}{0.04m,\\1.1°} & \tabincell{l}{0.04m,\\1.1°} & \tabincell{l}{0.09m,\\2.6°} \bigstrut\\
    \hline
    \multicolumn{1}{|p{3em}|}{\cite{ref6.34}} & \multicolumn{1}{p{3em}|}{\tabincell{l}{0.02m,\\0.6°}} & \multicolumn{1}{p{3em}|}{\tabincell{l}{0.03m,\\1.0°}} & \multicolumn{1}{p{3em}|}{\tabincell{l}{0.02m,\\1.1°}} & \multicolumn{1}{p{3em}|}{\tabincell{l}{0.03m,\\0.8°}} & \multicolumn{1}{p{3em}|}{\tabincell{l}{0.04m,\\1.1°}} & \multicolumn{1}{p{3em}|}{\tabincell{l}{0.04m,\\1.2°}} & \tabincell{l}{0.25m,\\4.5°} \bigstrut\\
    \hline
    \cite{ref6.35} & \tabincell{l}{0.19m,\\1.11°} & \tabincell{l}{0.19m,\\1.24°} & \tabincell{l}{\tabincell{l}{0.11m,\\1.82°}} & \tabincell{l}{0.26m,\\1.18°} & \tabincell{l}{0.42m,\\1.41°} & \tabincell{l}{0.30m,\\1.70°} & \tabincell{l}{0.41m,\\1.42°} \bigstrut\\
    \hline
    \tabincell{l}{rgb+3d\\model \cite{ref6.35}} & \tabincell{l}{0.18m,\\1.10°} & \tabincell{l}{0.19m,\\1.24°} & \tabincell{l}{0.22m,\\1.82°} & \tabincell{l}{0.25m,\\1.15°} & \tabincell{l}{0.39m,\\1.34°} & \tabincell{l}{0.38m,\\1.68°} & \tabincell{l}{0.29m,\\1.16°} \bigstrut\\
    \hline
    rgb-d \cite{ref6.35} & \tabincell{l}{0.10m,\\1.03°} & \tabincell{l}{0.11m,\\1.05°} & \tabincell{l}{0.10m,\\1.88°} & \tabincell{l}{0.12m,\\1.03°} & \tabincell{l}{0.20m,\\1.17°} & \tabincell{l}{0.21m,\\1.41°} & \tabincell{l}{0.26m,\\1.15°} \bigstrut\\
    \hline
    \cite{ref6.36} & \tabincell{l}{0.02m,\\0.8°} & \tabincell{l}{0.02m,\\1.0°} & \tabincell{l}{0.04m,\\2.7°} & \tabincell{l}{0.03m,\\0.8°} & \tabincell{l}{0.04m,\\1.1°} & \tabincell{l}{0.04m,\\1.1°} & \tabincell{l}{0.18m,\\3.9°} \bigstrut\\
    \hline
    SANet \cite{ref6.39} & \multicolumn{1}{p{3em}|}{\tabincell{l}{0.03m,\\0.88°}} & \multicolumn{1}{p{3em}|}{\tabincell{l}{0.03m,\\1.08°}} & \multicolumn{1}{p{3em}|}{\tabincell{l}{0.02m,\\1.48°}} & \multicolumn{1}{p{3em}|}{\tabincell{l}{0.03m,\\1.00°}} & \multicolumn{1}{p{3em}|}{\tabincell{l}{0.05m,\\1.32°}} & \multicolumn{1}{p{3em}|}{\tabincell{l}{0.04m,\\1.40°}} & \multicolumn{1}{p{3em}|}{\tabincell{l}{0.16m,\\4.59°}} \bigstrut\\
    \hline
    \tabincell{l}{NetVlad+\\DensePE \cite{ref6.31}} & \tabincell{l}{0.03m,\\1.05°} & \tabincell{l}{0.03m,\\1.06°} & \tabincell{l}{0.02m,\\1.06°} & \tabincell{l}{0.03m,\\1.05°} & \tabincell{l}{0.05m,\\1.55°} & \tabincell{l}{0.04m,\\1.31°} & \tabincell{l}{0.09m,\\2.47°} \bigstrut\\
    \hline
    \tabincell{l}{NetVlad+\\SparsePE \cite{ref6.31}} & \tabincell{l}{4m,\\1.83°} & \tabincell{l}{1m,\\1.55°} & \tabincell{l}{2m,\\1.65°} & \tabincell{l}{5m,\\1.49°} & \tabincell{l}{7m,\\1.87°} & \tabincell{l}{5m,\\1.61°} & \tabincell{l}{12m,\\3.41°} \bigstrut\\
    \hline
    \end{tabular}}%
  \label{tab10}%
\end{table}%

\begin{center}
    \begin{longtable}{|p{5.2em}|p{2.8em}|p{2.8em}|p{2.8em}|p{2.8em}|p{2.8em}|p{2.8em}|p{2.8em}|}
    \caption{A summary of published results of regression-based methods on the 7Scenes dataset} 
    \label{tab11} \\
    \hline
    Methods & Chess & Fire & Head & Office & \tabincell{l}{Pump\\kin} & \tabincell{l}{Kit\\chen} & Stairs \bigstrut\\
    \hline
PoseNet \cite{ref7.2} & 
\tabincell{l}{0.32$m$,\\8.12$^{\circ}$} & 
\tabincell{l}{0.47$m$,\\14.4$^{\circ}$} & 
\tabincell{l}{0.29$m$,\\12.0$^{\circ}$} & 
\tabincell{l}{0.48$m$,\\7.68$^{\circ}$} & 
\tabincell{l}{0.47$m$,\\8.42$^{\circ}$} & 
\tabincell{l}{0.59$m$,\\ 8.64$^{\circ}$} & 
\tabincell{l}{0.47$m$,\\13.8$^{\circ}$} \bigstrut\\
    \hline
\tabincell{l}{Dense\\PoseNet \cite{ref7.2}} & 
\tabincell{l}{0.32$m$,\\6.60$^{\circ}$} & 
\tabincell{l}{0.47$m$,\\14.0$^{\circ}$} & 
\tabincell{l}{0.30$m$,\\12.2$^{\circ}$} & 
\tabincell{l}{0.48$m$,\\7.24$^{\circ}$} & 
\tabincell{l}{0.49$m$,\\8.12$^{\circ}$} & 
\tabincell{l}{0.58$m$,\\8.34$^{\circ}$} & 
\tabincell{l}{0.48$m$,\\13.1$^{\circ}$} \bigstrut\\
    \hline
\tabincell{l}{Bayesian\\PoseNet \cite{ref7.3}} & 
\tabincell{l}{0.37$m$,\\7.24$^{\circ}$} & 
\tabincell{l}{0.43$m$,\\13.7$^{\circ}$} & 
\tabincell{l}{0.31$m$,\\12.0$^{\circ}$} & 
\tabincell{l}{0.48$m$,\\8.04$^{\circ}$} & 
\tabincell{l}{0.61$m$,\\7.08$^{\circ}$} & 
\tabincell{l}{0.58$m$,\\7.54$^{\circ}$} & 
\tabincell{l}{0.48$m$,\\13.1$^{\circ}$} \bigstrut\\
    \hline
\tabincell{l}{LSTM\\PoseNet \cite{ref7.5}} & 
\tabincell{l}{0.24 $m$,\\ 5.77$^{\circ}$} & 
\tabincell{l}{0.34 $m$,\\ 11.9 $^{\circ}$} & 
\tabincell{l}{0.21 $m$,\\ 13.7$^{\circ}$} & 
\tabincell{l}{0.30 $m$,\\ 8.08$^{\circ}$} & 
\tabincell{l}{0.33 $m$,\\ 7.00$^{\circ}$} & 
\tabincell{l}{0.37 $m$,\\ 8.83$^{\circ}$} & 
\tabincell{l}{0.40 $m$,\\ 13.7 $^{\circ}$} \bigstrut\\
    \hline
\tabincell{l}{Hourglass\\PoseNet \cite{ref7.5}}  & 
\tabincell{l}{0.15$m$,\\6.17$^{\circ}$} & 
\tabincell{l}{0.27$m$,\\ 10.84$^{\circ}$} & 
\tabincell{l}{0.19$m$,\\ 11.63$^{\circ}$} & 
\tabincell{l}{0.21$m$,\\8.48$^{\circ}$} & 
\tabincell{l}{0.25$m$,\\7.01$^{\circ}$} & 
\tabincell{l}{0.27$m$,\\ 10.15$^{\circ}$} & 
\tabincell{l}{0.29$m$,\\ 12.46$^{\circ}$} \bigstrut\\
    \hline
\tabincell{l}{BranchNet\\\cite{ref7.6}} & 
\tabincell{l}{0.18$m$,\\5.17$^{\circ}$} & 
\tabincell{l}{0.34$m$,\\8.99$^{\circ}$} & 
\tabincell{l}{0.20$m$,\\ 14.15$^{\circ}$} & 
\tabincell{l}{0.30$m$,\\7.05$^{\circ}$} & 
\tabincell{l}{0.27$m$,\\5.10$^{\circ}$} & 
\tabincell{l}{0.33$m$,\\7.40$^{\circ}$} & 
\tabincell{l}{0.38$m$,\\ 10.26$^{\circ}$} \bigstrut\\
    \hline
\tabincell{l}{Geo.PoseNet\\\cite{ref7.7}} & 
\tabincell{l}{0.14$m$,\\4.50$^{\circ}$} & 
\tabincell{l}{0.27$m$,\\11.8$^{\circ}$} & 
\tabincell{l}{0.18$m$,\\12.1$^{\circ}$} & 
\tabincell{l}{0.20$m$,\\5.77$^{\circ}$} & 
\tabincell{l}{0.25$m$,\\4.82$^{\circ}$} & 
\tabincell{l}{0.24$m$,\\5.52$^{\circ}$} & 
\tabincell{l}{0.37$m$,\\10.6$^{\circ}$} \bigstrut\\
    \hline
AtLoc \cite{ref7.8} & 
\tabincell{l}{0.10$m$,\\4.07$^{\circ}$} & 
\tabincell{l}{0.25$m$,\\11.4$^{\circ}$} & 
\tabincell{l}{0.16$m$,\\11.8$^{\circ}$} & 
\tabincell{l}{0.17$m$,\\5.34$^{\circ}$} & 
\tabincell{l}{0.21$m$,\\4.37$^{\circ}$} & 
\tabincell{l}{0.23$m$,\\5.42$^{\circ}$} & 
\tabincell{l}{0.26$m$,\\10.5$^{\circ}$} \bigstrut\\
    \hline
AdPR \cite{ref7.9} & 
\tabincell{l}{0.12$m$,\\4.8$^{\circ}$} & 
\tabincell{l}{0.27$m$,\\11.6$^{\circ}$} & 
\tabincell{l}{0.16$m$,\\12.4$^{\circ}$} & 
\tabincell{l}{0.19$m$,\\6.8$^{\circ}$} & 
\tabincell{l}{0.21$m$,\\5.2$^{\circ}$} & 
\tabincell{l}{0.25$m$,\\6.0$^{\circ}$} & 
\tabincell{l}{0.28$m$,\\8.4$^{\circ}$} \bigstrut\\
    \hline
\tabincell{l}{APANet\\\cite{ref7.11}} & 
\tabincell{l}{N/A,\\N/A} & 
\tabincell{l}{0.21$m$,\\9.72$^{\circ}$} & 
\tabincell{l}{0.15$m$,\\9.35$^{\circ}$} & 
\tabincell{l}{0.15$m$,\\6.69$^{\circ}$} & 
\tabincell{l}{0.19$m$,\\5.87$^{\circ}$} & 
\tabincell{l}{0.16$m$,\\5.13$^{\circ}$} & 
\tabincell{l}{0.16$m$,\\11.77$^{\circ}$} \bigstrut\\
    \hline
\tabincell{l}{SPPNet\\\cite{ref7.12}} & 
\tabincell{l}{0.12$m$,\\4.42$^{\circ}$} & 
\tabincell{l}{0.22$m$,\\8.84$^{\circ}$} & 
\tabincell{l}{0.11$m$,\\8.33$^{\circ}$} & 
\tabincell{l}{0.16$m$,\\4.99$^{\circ}$} & 
\tabincell{l}{0.21$m$,\\4.89$^{\circ}$} & 
\tabincell{l}{0.21$m$,\\4.76$^{\circ}$} & 
\tabincell{l}{0.22$m$,\\7.17$^{\circ}$} \bigstrut\\
    \hline
\tabincell{l}{Geo.PoseNet\\(reprojec\\tion)\cite{ref7.7}} & 
\tabincell{l}{0.13$m$,\\4.48$^{\circ}$} & 
\tabincell{l}{0.27$m$,\\11.3$^{\circ}$} & 
\tabincell{l}{0.17$m$,\\13.0$^{\circ}$} & 
\tabincell{l}{0.19$m$,\\5.55$^{\circ}$} & 
\tabincell{l}{0.26$m$,\\4.75$^{\circ}$} & 
\tabincell{l}{0.23$m$,\\5.35$^{\circ}$} & 
\tabincell{l}{0.35$m$,\\12.4$^{\circ}$} \bigstrut\\
    \hline
\tabincell{l}{GPoseNet\\\cite{ref7.13}} & 
\tabincell{l}{0.20$m$,\\7.11$^{\circ}$} & 
\tabincell{l}{0.38$m$,\\12.3$^{\circ}$} & 
\tabincell{l}{0.21$m$,\\13.8$^{\circ}$} & 
\tabincell{l}{0.28$m$,\\8.83$^{\circ}$} & 
\tabincell{l}{0.37$m$,\\6.94$^{\circ}$} & 
\tabincell{l}{0.35$m$,\\8.15$^{\circ}$} & 
\tabincell{l}{0.37$m$,\\12.5$^{\circ}$} \bigstrut\\
    \hline
\tabincell{l}{MapNet\\\cite{ref8.1}} & 
\tabincell{l}{0.08$m$,\\3.25$^{\circ}$} & 
\tabincell{l}{0.27$m$,\\11.7$^{\circ}$} &  
\tabincell{l}{0.18$m$,\\13.3$^{\circ}$} & 
\tabincell{l}{0.17$m$,\\5.15$^{\circ}$} & 
\tabincell{l}{0.22$m$,\\4.02$^{\circ}$} & 
\tabincell{l}{0.23$m$,\\4.93$^{\circ}$} & 
\tabincell{l}{0.30$m$,\\12.1$^{\circ}$}  \bigstrut\\
    \hline
\tabincell{l}{LSG\\\cite{ref8.2}} & 
\tabincell{l}{0.09$m$,\\3.28$^{\circ}$} & 
\tabincell{l}{0.26$m$,\\10.92$^{\circ}$} & 
\tabincell{l}{0.17$m$,\\12.70$^{\circ}$} & 
\tabincell{l}{0.18$m$,\\5.45$^{\circ}$} & 
\tabincell{l}{0.20$m$,\\3.69$^{\circ}$} & 
\tabincell{l}{0.23$m$,\\4.92$^{\circ}$} & 
\tabincell{l}{0.23$m$,\\11.3$^{\circ}$} \bigstrut\\
    \hline
\tabincell{l}{VlocNet\\\cite{ref8.3}} & 
\tabincell{l}{0.036$m$,\\1.71$^{\circ}$} & 
\tabincell{l}{0.039$m$,\\5.34$^{\circ}$} & 
\tabincell{l}{0.046$m$,\\ 6.64$^{\circ}$} & 
\tabincell{l}{0.039$m$,\\ 1.95$^{\circ}$} & 
\tabincell{l}{0.037$m$,\\2.28$^{\circ}$} & 
\tabincell{l}{0.039$m$,\\2.20$^{\circ}$} & 
\tabincell{l}{0.097$m$,\\6.48$^{\circ}$} \bigstrut\\
    \hline
\tabincell{l}{VlocNet++\\\cite{ref8.4}} & 
\tabincell{l}{0.023$m$,\\1.44$^{\circ}$} & 
\tabincell{l}{0.018$m$,\\1.39$^{\circ}$} & 
\tabincell{l}{0.016$m$,\\0.99$^{\circ}$} & 
\tabincell{l}{0.024$m$,\\1.14$^{\circ}$} & 
\tabincell{l}{0.024$m$,\\1.45$^{\circ}$} & 
\tabincell{l}{0.025$m$,\\2.27$^{\circ}$} & 
\tabincell{l}{0.021$m$,\\1.08$^{\circ}$} \bigstrut\\
    \hline
\tabincell{l}{DGRNet\\\cite{ref8.5}} & 
\tabincell{l}{0.016$m$,\\1.72$^{\circ}$} & 
\tabincell{l}{0.011$m$,\\2.19$^{\circ}$} & 
\tabincell{l}{0.017$m$,\\3.56$^{\circ}$} & 
\tabincell{l}{0.024$m$,\\1.95$^{\circ}$} & 
\tabincell{l}{0.022$m$,\\2.27$^{\circ}$} & 
\tabincell{l}{0.018$m$,\\1.86$^{\circ}$} & 
\tabincell{l}{0.017$m$,\\4.79$^{\circ}$} \bigstrut\\
    \hline
\tabincell{l}{AtLocPlus\\\cite{ref7.8}} & 
\tabincell{l}{0.10$m$,\\3.18$^{\circ}$} & 
\tabincell{l}{0.26$m$,\\10.8$^{\circ}$} & 
\tabincell{l}{0.14$m$,\\11.4$^{\circ}$} & 
\tabincell{l}{0.17$m$,\\5.16$^{\circ}$} & 
\tabincell{l}{0.20$m$,\\3.94$^{\circ}$} & 
\tabincell{l}{0.16$m$,\\4.90$^{\circ}$} & 
\tabincell{l}{0.29$m$,\\10.2$^{\circ}$} \bigstrut\\
    \hline
\tabincell{l}{VidLoc\\\cite{ref8.6}} & 
\tabincell{l}{0.18$m$,\\N/A} & 
\tabincell{l}{0.26$m$,\\N/A}& 
\tabincell{l}{0.14$m$,\\N/A} & 
\tabincell{l}{0.26$m$,\\N/A} & 
\tabincell{l}{0.36$m$,\\N/A} & 
\tabincell{l}{0.31$m$,\\N/A} & 
\tabincell{l}{0.26$m$,\\N/A} \bigstrut\\
    \hline
\tabincell{l}{MapNet+\\\cite{ref8.1}} & 
\tabincell{l}{0.10$m$,\\3.17$^{\circ}$} & 
\tabincell{l}{0.20$m$,\\9.04$^{\circ}$} & 
\tabincell{l}{0.13$m$,\\11.13$^{\circ}$} & 
\tabincell{l}{0.18$m$,\\5.38$^{\circ}$} & 
\tabincell{l}{0.19$m$,\\3.92$^{\circ}$} & 
\tabincell{l}{0.20$m$,\\5.01$^{\circ}$} & 
\tabincell{l}{0.30$m$,\\13.37$^{\circ}$} \bigstrut\\
    \hline
\tabincell{l}{MapNet+\\PGO \cite{ref8.1}} & 
\tabincell{l}{0.09$m$,\\3.24$^{\circ}$} & 
\tabincell{l}{0.20$m$,\\9.29$^{\circ}$} & 
\tabincell{l}{0.12$m$,\\8.45$^{\circ}$} & 
\tabincell{l}{0.19$m$,\\5.42$^{\circ}$} & 
\tabincell{l}{0.19$m$,\\3.96$^{\circ}$} & 
\tabincell{l}{0.20$m$,\\4.94$^{\circ}$} & 
\tabincell{l}{0.27$m$,\\10.57$^{\circ}$} \bigstrut\\
    \hline
NNnet \cite{ref9.1}  & 
\tabincell{l}{0.13$m$,\\6.46$^{\circ}$} & 
\tabincell{l}{0.26$m$,\\12.72$^{\circ}$} & 
\tabincell{l}{0.14$m$,\\12.34$^{\circ}$} & 
\tabincell{l}{0.21$m$,\\7.35$^{\circ}$} & 
\tabincell{l}{0.24$m$,\\6.35$^{\circ}$} & 
\tabincell{l}{0.24$m$,\\8.03$^{\circ}$} & 
\tabincell{l}{0.27$m$,\\11.82$^{\circ}$} \bigstrut\\
    \hline
\tabincell{l}{RelocNet\\\cite{ref9.2}} & 
\tabincell{l}{0.12$m$,\\4.14$^{\circ}$} & 
\tabincell{l}{0.26$m$,\\10.4$^{\circ}$} & 
\tabincell{l}{0.14$m$,\\10.5$^{\circ}$} & 
\tabincell{l}{0.18$m$,\\5.32$^{\circ}$} & 
\tabincell{l}{0.26$m$,\\4.17$^{\circ}$} & 
\tabincell{l}{0.23$m$,\\5.08$^{\circ}$} & 
\tabincell{l}{0.28$m$,\\7.53$^{\circ}$} \bigstrut\\
    \hline
\tabincell{l}{CamNet\\\cite{ref9.3}} & 
\tabincell{l}{0.04$m$,\\1.73$^{\circ}$} & 
\tabincell{l}{0.03$m$,\\1.74$^{\circ}$} & 
\tabincell{l}{0.05$m$,\\1.98$^{\circ}$} & 
\tabincell{l}{0.04$m$,\\1.62$^{\circ}$} & 
\tabincell{l}{0.04$m$,\\1.64$^{\circ}$} & 
\tabincell{l}{0.04$m$,\\1.63$^{\circ}$} & 
\tabincell{l}{0.04$m$,\\1.51$^{\circ}$} \bigstrut\\
    \hline
\tabincell{l}{AnchorNet\\\cite{ref9.6}} & 
\tabincell{l}{0.08$m$,\\4.12$^{\circ}$} & 
\tabincell{l}{0.16$m$,\\11.1$^{\circ}$} & 
\tabincell{l}{0.09$m$,\\11.2$^{\circ}$} & 
\tabincell{l}{0.11$m$,\\5.38$^{\circ}$} & 
\tabincell{l}{0.14$m$,\\3.55$^{\circ}$} & 
\tabincell{l}{0.13$m$,\\5.29$^{\circ}$} & 
\tabincell{l}{0.21$m$,\\11.9$^{\circ}$} \bigstrut\\
    \hline 
\end{longtable}%
\end{center}

\begin{center}
    \begin{longtable}{|l|l|l|l|l|l|l|}
    \caption{A summary of published results of regression-based methods on the 7Scenes dataset} 
    \label{tab12} \\
    \hline
\tabincell{l}{PoseNet\\\cite{ref7.2}} & 
\tabincell{l}{N/A,\\N/A}& 
\tabincell{l}{1.66$m$,\\4.86$^{\circ}$}& 
\tabincell{l}{2.96$m$,\\6.00$^{\circ}$}& 
\tabincell{l}{2.62$m$,\\4.90$^{\circ}$}& 
\tabincell{l}{1.41$m$,\\7.18$^{\circ}$}& 
\tabincell{l}{2.45$m$,\\7.96$^{\circ}$}\bigstrut\\
    \hline
\tabincell{l}{Dense\\PoseNet \cite{ref7.2}} & 
\tabincell{l}{N/A,\\N/A}& 
\tabincell{l}{1.92$m$,\\5.40$^{\circ}$}& 
\tabincell{l}{N/A,\\N/A}& 
\tabincell{l}{2.31$m$,\\5.38$^{\circ}$}& 
\tabincell{l}{1.46$m$,\\8.08$^{\circ}$}& 
\tabincell{l}{2.65$m$,\\8.46$^{\circ}$}\bigstrut\\
    \hline
\tabincell{l}{Bayesian\\PoseNet \cite{ref7.3}} & 
\tabincell{l}{N/A,\\N/A}& 
\tabincell{l}{1.74$m$,\\4.06$^{\circ}$}& 
\tabincell{l}{2.14$m$,\\4.96$^{\circ}$}& 
\tabincell{l}{2.57$m$,\\5.14$^{\circ}$}& 
\tabincell{l}{1.25$m$,\\7.54$^{\circ}$}& 
\tabincell{l}{2.11$m$,\\8.38$^{\circ}$}\bigstrut\\
    \hline
\tabincell{l}{LSTM\\PoseNet \cite{ref7.5}} & 
\tabincell{l}{N/A,\\N/A}& 
\tabincell{l}{0.99 $m$,\\3.65$^{\circ}$}& 
\tabincell{l}{N/A,\\N/A}& 
\tabincell{l}{1.51 $m$,\\4.29$^{\circ}$}& 
\tabincell{l}{1.18$m$,\\7.44$^{\circ}$}& 
\tabincell{l}{1.52$m$,\\6.68$^{\circ}$}\bigstrut\\
    \hline
\tabincell{l}{Hourglass\\PoseNet \cite{ref7.5}} & 
\tabincell{l}{N/A,\\N/A}& 
\tabincell{l}{N/A,\\N/A}& 
\tabincell{l}{N/A,\\N/A}& 
\tabincell{l}{N/A,\\N/A}& 
\tabincell{l}{N/A,\\N/A}& 
\tabincell{l}{N/A,\\N/A}\bigstrut\\
    \hline
\tabincell{l}{SVS\\PoseNet \cite{ref7.1}} & 
\tabincell{l}{N/A,\\N/A}& 
\tabincell{l}{1.06$m$,\\2.81$^{\circ}$}& 
\tabincell{l}{N/A,\\N/A}& 
\tabincell{l}{1.50$m$,\\4.03$^{\circ}$}& 
\tabincell{l}{0.63$m$,\\5.73$^{\circ}$}& 
\tabincell{l}{2.11$m$,\\9.11$^{\circ}$}\bigstrut\\
    \hline
\tabincell{l}{Geo.PoseNet\\\cite{ref7.7}} & 
\tabincell{l}{7.00$m$,\\3.65$^{\circ}$}& 
\tabincell{l}{0.99$m$,\\1.06$^{\circ}$}& 
\tabincell{l}{20.7$m$,\\25.7$^{\circ}$}& 
\tabincell{l}{2.17$m$,\\2.94$^{\circ}$}& 
\tabincell{l}{1.05$m$,\\3.97$^{\circ}$}& 
\tabincell{l}{1.49$m$,\\3.43$^{\circ}$}\bigstrut\\
    \hline
\tabincell{l}{Geo.PoseNet\\ (reprojec\\tion) \cite{ref7.7}} & 
\tabincell{l}{6.83$m$,\\3.47$^{\circ}$}& 
\tabincell{l}{0.88$m$,\\1.04$^{\circ}$}& 
\tabincell{l}{20.3$m$,\\25.5$^{\circ}$}& 
\tabincell{l}{3.20$m$,\\3.29$^{\circ}$}& 
\tabincell{l}{0.88$m$,\\3.78$^{\circ}$}& 
\tabincell{l}{1.57$m$,\\3.32$^{\circ}$}\bigstrut\\
    \hline
PVL \cite{ref7.10} & 
\tabincell{l}{N/A,\\N/A}& 
\tabincell{l}{1.30$m$,\\1.67$^{\circ}$}& 
\tabincell{l}{N/A,\\N/A}& 
\tabincell{l}{N/A,\\N/A}& 
\tabincell{l}{1.22$m$,\\6.17$^{\circ}$}& 
\tabincell{l}{2.28$m$,\\4.80$^{\circ}$}\bigstrut\\
    \hline
\tabincell{l}{APANet\\\cite{ref7.11}} & 
\tabincell{l}{N/A,\\N/A}& 
\tabincell{l}{N/A,\\N/A}& 
\tabincell{l}{N/A,\\N/A}& 
\tabincell{l}{0.98$m$,\\1.94$^{\circ}$}& 
\tabincell{l}{0.62$m$,\\2.49$^{\circ}$}& 
\tabincell{l}{0.77$m$,\\2.25$^{\circ}$}\bigstrut\\
    \hline
SPPNet \cite{ref7.12} & 
\tabincell{l}{5.42$m$,\\2.84$^{\circ}$}& 
\tabincell{l}{0.74$m$,\\0.96$^{\circ}$}& 
\tabincell{l}{24.5$m$,\\23.8$^{\circ}$}& 
\tabincell{l}{2.18$m$,\\3.92$^{\circ}$}& 
\tabincell{l}{0.59$m$,\\2.53$^{\circ}$}& 
\tabincell{l}{1.83$m$,\\3.35$^{\circ}$}\bigstrut\\
    \hline
\tabincell{l}{GPoseNet\\\cite{ref7.13}} & 
\tabincell{l}{N/A,\\N/A}& 
\tabincell{l}{1.61$m$,\\2.29$^{\circ}$}& 
\tabincell{l}{N/A,\\N/A}& 
\tabincell{l}{2.62$m$,\\3.89$^{\circ}$}& 
\tabincell{l}{1.14$m$,\\5.73$^{\circ}$}& 
\tabincell{l}{2.93$m$,\\6.46$^{\circ}$}\bigstrut\\
    \hline
VlocNet \cite{ref8.3} & 
\tabincell{l}{N/A,\\N/A}& 
\tabincell{l}{0.836$m$,\\1.419$^{\circ}$}& 
\tabincell{l}{N/A,\\N/A}& 
\tabincell{l}{1.075$m$,\\2.411$^{\circ}$}& 
\tabincell{l}{0.593$m$,\\3.529$^{\circ}$}& 
\tabincell{l}{0.631$m$,\\3.906$^{\circ}$}\bigstrut\\
    \hline
\tabincell{l}{AnchorNet\\\cite{ref9.6}} & 
\tabincell{l}{5.89$m$,\\3.53$^{\circ}$}& 
\tabincell{l}{0.79$m$,\\0.95$^{\circ}$}& 
\tabincell{l}{11.8$m$,\\24.3$^{\circ}$}& 
\tabincell{l}{2.11$m$,\\3.05$^{\circ}$}& 
\tabincell{l}{0.77$m$,\\3.25$^{\circ}$}& 
\tabincell{l}{1.22$m$,\\3.02$^{\circ}$}\bigstrut\\
    \hline
DSAC++ \cite{ref6.33} & 
\tabincell{l}{0.40$m$,\\0.2$^{\circ}$}& 
\tabincell{l}{0.18$m$,\\0.3$^{\circ}$}& 
\tabincell{l}{N/A,\\N/A}& 
\tabincell{l}{0.20$m$,\\0.3$^{\circ}$}& 
\tabincell{l}{0.06$m$,\\0.3$^{\circ}$}& 
\tabincell{l}{0.13$m$,\\0.4$^{\circ}$}\bigstrut\\
    \hline
\tabincell{l}{Scene coord\\inate \cite{ref6.34}} & 
\tabincell{l}{0.51$m$,\\0.3$^{\circ}$}& 
\tabincell{l}{0.18$m$,\\0.3$^{\circ}$}& 
\tabincell{l}{N/A,\\N/A}& 
\tabincell{l}{0.19$m$,\\0.4$^{\circ}$}& 
\tabincell{l}{0.07$m$,\\0.3$^{\circ}$}& 
\tabincell{l}{0.25$m$,\\0.7$^{\circ}$}\bigstrut\\
    \hline
RGB \cite{ref6.35} & 
\tabincell{l}{0.335$m$,\\0.21$^{\circ}$}& 
\tabincell{l}{0.179$m$,\\0.31$^{\circ}$}& 
\tabincell{l}{N/A,\\N/A}& 
\tabincell{l}{0.212$m$,\\0.38$^{\circ}$}& 
\tabincell{l}{0.52$m$,\\0.25$^{\circ}$}& 
\tabincell{l}{0.151$m$,\\0.50$^{\circ}$}\bigstrut\\
    \hline
\tabincell{l}{rgb+3d\\model \cite{ref6.35}} & 
\tabincell{l}{0.485$m$,\\0.25$^{\circ}$}& 
\tabincell{l}{0.147$m$,\\0.29$^{\circ}$}& 
\tabincell{l}{N/A,\\N/A}& 
\tabincell{l}{0.210$m$,\\0.41$^{\circ}$}& 
\tabincell{l}{0.46$m$,\\0.25$^{\circ}$}& 
\tabincell{l}{0.134$m$,\\0.45$^{\circ}$}\bigstrut\\
    \hline
\tabincell{l}{Multi-View\\\cite{ref6.36}} & 
\tabincell{l}{0.62$m$,\\0.4$^{\circ}$}& 
\tabincell{l}{0.20$m$,\\0.3$^{\circ}$}& 
\tabincell{l}{N/A,\\N/A}& 
\tabincell{l}{0.19$m$,\\0.4$^{\circ}$}& 
\tabincell{l}{0.07$m$,\\0.3$^{\circ}$}& 
\tabincell{l}{0.20$m$,\\0.6$^{\circ}$}\bigstrut\\
    \hline
SanNet \cite{ref6.39} & 
\tabincell{l}{3.28$m$,\\1.95$^{\circ}$}& 
\tabincell{l}{0.32$m$,\\0.54$^{\circ}$}& 
\tabincell{l}{8.74$m$,\\12.64$^{\circ}$}& 
\tabincell{l}{0.32$m$,\\0.54$^{\circ}$}& 
\tabincell{l}{0.10$m$,\\0.47$^{\circ}$}& 
\tabincell{l}{0.16$m$,\\0.57$^{\circ}$}\bigstrut\\
    \hline
\end{longtable}%
\end{center}

\begin{table}[H]
  \centering
  \caption{A summary of published results on the RobotCar dataset (mean)}
    \begin{tabular}{|l|l|l|l|l|l|}
    \hline
    Methods & LOOP1 & LOOP2 & FULL1 & FULL2 & average \bigstrut\\
    \hline
PoseNet \cite{ref7.2} & 
\tabincell{l}{28.81$m$,\\19.62$^{\circ}$}& 
\tabincell{l}{25.29$m$,\\17.45$^{\circ}$}& 
\tabincell{l}{125.6$m$,\\27.1$^{\circ}$}& 
\tabincell{l}{131.06$m$,\\26.05$^{\circ}$}& 
\tabincell{l}{77.85$m$,\\22.56$^{\circ}$}\bigstrut\\
    \hline
AtLoc \cite{ref7.8} & 
\tabincell{l}{8.61$m$,\\4.58$^{\circ}$}& 
\tabincell{l}{8.86$m$,\\4.67$^{\circ}$}& 
\tabincell{l}{29.6$m$,\\12.4$^{\circ}$}& 
\tabincell{l}{48.2$m$,\\11.1$^{\circ}$}& 
\tabincell{l}{23.8$m$,\\8.19$^{\circ}$}\bigstrut\\
    \hline
MapNet \cite{ref8.1} & 
\tabincell{l}{8.76$m$,\\3.46$^{\circ}$}& 
\tabincell{l}{9.84$m$,\\3.96$^{\circ}$}& 
\tabincell{l}{41.4$m$,\\12.5$^{\circ}$}& 
\tabincell{l}{59.3$m$,\\14.8$^{\circ}$}& 
\tabincell{l}{29.8$m$,\\8.68$^{\circ}$}\bigstrut\\
    \hline
LSG \cite{ref8.2} & 
\tabincell{l}{9.07$m$,\\3.31$^{\circ}$}& 
\tabincell{l}{9.19$m$,\\3.53$^{\circ}$}& 
\tabincell{l}{31.65$m$,\\4.51$^{\circ}$}& 
\tabincell{l}{53.45$m$,\\8.60$^{\circ}$}& 
\tabincell{l}{25.84$m$,\\4.99$^{\circ}$}\bigstrut\\
    \hline
    \end{tabular}%
  \label{tab13}%
\end{table}%

\begin{table}[H]
  \centering
  \caption{A summary of published structure-based methods on the Aachen dataset}
  \resizebox{\textwidth}{!}{
    \begin{tabular}{|l|l|l|l|l|l|l|l|l|l|}
    \hline
    Methods & \multicolumn{3}{c|}{Day} & \multicolumn{3}{c|}{Night} & \multicolumn{3}{c|}{ All} \bigstrut\\
    \hline
    \tabincell{l}{Threshold\\Accuracy\%} & \tabincell{l}{(0.25\\m,2$^{\circ}$)} & \tabincell{l}{(0.50\\m,5$^{\circ}$)} & \tabincell{l}{(5.0\\m,10$^{\circ}$)} & \tabincell{l}{(0.5\\m,2$^{\circ}$)} & \tabincell{l}{(1.0\\m,5$^{\circ}$)} & \tabincell{l}{(5.0\\m,10$^{\circ}$)} & \multicolumn{1}{l|}{\tabincell{l}{(0.5\\m,2$^{\circ}$)}} & \multicolumn{1}{l|}{\tabincell{l}{(1.0\\m,5$^{\circ}$)}} & \multicolumn{1}{l|}{\tabincell{l}{(5.0\\m,10$^{\circ}$)}} \bigstrut\\
    \hline
    NetVLAD \cite{ref5.12} & \multicolumn{1}{l|}{0} & \multicolumn{1}{l|}{0.2} & \multicolumn{1}{l|}{18.9} & \multicolumn{1}{l|}{0} & \multicolumn{1}{l|}{2} & \multicolumn{1}{l|}{12.2} & \multicolumn{1}{l|}{/} & \multicolumn{1}{l|}{/} & \multicolumn{1}{l|}{/} \bigstrut\\
    \hline
    HF-Net \cite{ref5.1} & \multicolumn{1}{l|}{75.7} & \multicolumn{1}{l|}{84.3} & \multicolumn{1}{l|}{90.9} & \multicolumn{1}{l|}{40.8} & \multicolumn{1}{l|}{55.1} & \multicolumn{1}{l|}{72.4} & \multicolumn{1}{l|}{/} & \multicolumn{1}{l|}{/} & \multicolumn{1}{l|}{/} \bigstrut\\
    \hline
    UR2KID \cite{ref6.41} & \multicolumn{1}{l|}{79.9} & \multicolumn{1}{l|}{88.6} & \multicolumn{1}{l|}{93.6} & \multicolumn{1}{l|}{45.9} & \multicolumn{1}{l|}{64.3} & \multicolumn{1}{l|}{83.7} & \multicolumn{1}{l|}{/} & \multicolumn{1}{l|}{/} & \multicolumn{1}{l|}{/} \bigstrut\\
    \hline
    \tabincell{l}{Dense Seman\\tic loc \cite{ref6.45}} & \multicolumn{1}{l|}{89.3} & \multicolumn{1}{l|}{95.4} & \multicolumn{1}{l|}{97.6} & \multicolumn{1}{l|}{44.9} & \multicolumn{1}{l|}{67.3} & \multicolumn{1}{l|}{87.8} & \multicolumn{1}{l|}{/} & \multicolumn{1}{l|}{/} & \multicolumn{1}{l|}{/} \bigstrut\\
    \hline
    S2D \cite{ref6.27} & \multicolumn{1}{l|}{84.3} & \multicolumn{1}{l|}{90.9} & \multicolumn{1}{l|}{95.9} & \multicolumn{1}{l|}{46.9} & \multicolumn{1}{l|}{69.4} & \multicolumn{1}{l|}{86.7} & \multicolumn{1}{l|}{/} & \multicolumn{1}{l|}{/} & \multicolumn{1}{l|}{/} \bigstrut\\
    \hline
    CSL \cite{ref4.15} & \multicolumn{1}{l|}{45.3} & \multicolumn{1}{l|}{73.5} & \multicolumn{1}{l|}{90.1} & \multicolumn{1}{l|}{0.6} & \multicolumn{1}{l|}{2.6} & \multicolumn{1}{l|}{7.2} & \multicolumn{1}{l|}{/} & \multicolumn{1}{l|}{/} & \multicolumn{1}{l|}{/} \bigstrut\\
    \hline
    \tabincell{l}{Active\\search \cite{ref4.12}} & \multicolumn{1}{l|}{35.6} & \multicolumn{1}{l|}{67.9} & \multicolumn{1}{l|}{90.4} & \multicolumn{1}{l|}{0.9} & \multicolumn{1}{l|}{2.1} & \multicolumn{1}{l|}{4.3} & \multicolumn{1}{l|}{/} & \multicolumn{1}{l|}{/} & \multicolumn{1}{l|}{/} \bigstrut\\
    \hline
    \cite{ref6.42} & \multicolumn{1}{l|}{41.6} & \multicolumn{1}{l|}{73.3} & \multicolumn{1}{l|}{90.1} & \multicolumn{1}{l|}{0.3} & \multicolumn{1}{l|}{1.9} & \multicolumn{1}{l|}{8.2} & \multicolumn{1}{l|}{/} & \multicolumn{1}{l|}{/} & \multicolumn{1}{l|}{/} \bigstrut\\
    \hline
    \cite{ref3.5} & \multicolumn{1}{l|}{45.5} & \multicolumn{1}{l|}{77} & \multicolumn{1}{l|}{94.7} & \multicolumn{1}{l|}{2.7} & \multicolumn{1}{l|}{6.9} & \multicolumn{1}{l|}{12.1} & \multicolumn{1}{l|}{/} & \multicolumn{1}{l|}{/} & \multicolumn{1}{l|}{/} \bigstrut\\
    \hline
    DELF \cite{ref5.13} & / & / & / & / & / & / & 38.8 & 62.2 & 85.7 \bigstrut\\
    \hline
    Superpoint \cite{ref6.9} & / & / & / & / & / & / & 42.8 & 57.1 & 75.5 \bigstrut\\
    \hline
    D2-net \cite{ref6.23} & / & / & / & / & / & / & 44.9 & 64.3 & 88.8 \bigstrut\\
    \hline
    R2D2 \cite{ref6.21} & / & / & / & / & / & / & 42.8 & 57.1 & 75.5 \bigstrut\\
    \hline
    ALSFeat\cite{ref6.22} & / & / & / & / & / & / & 46.9 & 65.3 & 88.8 \bigstrut\\
    \hline
    \tabincell{l}{Describe-to\\-detect \cite{ref6.24}} & / & / & / & / & / & / & 42.9 & 64.3 & 85.7 \bigstrut\\
    \hline
    \tabincell{l}{Dense-to\\-Dense \cite{ref6.28}} & / & / & / & / & / & / & 44.9 & 68.4 & 88.8 \bigstrut\\
    \hline
    \end{tabular}}%
  \label{tab14}%
\end{table}%

\begin{table}[H]
  \centering
  \caption{A summary of part published structure based methods on RobotCar}
  \resizebox{\textwidth}{!}{
    \begin{tabular}{|c|cc|cc|cc|cc|cc|cc|}
    \hline
    Methods & \multicolumn{12}{c|}{RobotCar} \bigstrut\\
    \hline
    Scene & \multicolumn{3}{c|}{dusk} & \multicolumn{3}{c|}{sun} & \multicolumn{3}{c|}{night} & \multicolumn{3}{c|}{night-rain} \bigstrut\\
    \hline
    NetVLAD \cite{ref5.12} & \multicolumn{1}{c|}{7.4} & 29.7 & \multicolumn{1}{c|}{92.9} & 5.7 & \multicolumn{1}{c|}{16.5} & 86.7 & \multicolumn{1}{c|}{0.2} & 1.8 & \multicolumn{1}{c|}{15.5} & 0.5 & \multicolumn{1}{c|}{2.7} & 16.4 \bigstrut\\
    \hline
    HF-Net \cite{ref5.1} & \multicolumn{1}{c|}{53.9} & 81.5 & \multicolumn{1}{c|}{94.2} & 48.5 & \multicolumn{1}{c|}{69.1} & 85.7 & \multicolumn{1}{c|}{2.7} & 6.6 & \multicolumn{1}{c|}{15.8} & 4.7 & \multicolumn{1}{c|}{16.8} & 21.8 \bigstrut\\
    \hline
    Scene & \multicolumn{6}{c|}{day} & \multicolumn{6}{c|}{night} \bigstrut\\
    \hline
    S2D \cite{ref6.27} & \multicolumn{2}{c|}{45.7} & \multicolumn{2}{c|}{78} & \multicolumn{2}{c|}{95.1} & \multicolumn{2}{c|}{22.3} & \multicolumn{2}{c|}{61.8} & \multicolumn{2}{c|}{94.5} \bigstrut\\
    \hline
    CSL \cite{ref4.15} & \multicolumn{2}{c|}{52.3} & \multicolumn{2}{c|}{80} & \multicolumn{2}{c|}{94.3} & \multicolumn{2}{c|}{24.5} & \multicolumn{2}{c|}{33.7} & \multicolumn{2}{c|}{49} \bigstrut\\
    \hline
    active search \cite{ref4.12} & \multicolumn{2}{c|}{57.3} & \multicolumn{2}{c|}{83.7} & \multicolumn{2}{c|}{96.6} & \multicolumn{2}{c|}{19.4} & \multicolumn{2}{c|}{30.6} & \multicolumn{2}{c|}{43.9} \bigstrut\\
    \hline
    \end{tabular}}%
  \label{tab15}%
\end{table}%

\begin{table}[H]
  \centering
  \caption{A summary of part published structure-based methods on CMU. The accuracy is measured at threshold (0.25m, 2$^{\circ}$), (0.5m, 5$^{\circ}$)and (5.0m, 10$^{\circ}$)}
  \resizebox{\textwidth}{!}{
    \begin{tabular}{|l|r|r|r|r|r|r|}
    \hline
    Methods & \multicolumn{6}{c|}{CMU} \bigstrut\\
    \hline
    Scene & \multicolumn{3}{c|}{urban} & \multicolumn{3}{c|}{suburban} \bigstrut\\
    \hline
    Threshold & \multicolumn{1}{l|}{0.25m,2$^{\circ}$} & \multicolumn{1}{l|}{0.50m,5$^{\circ}$} & \multicolumn{1}{l|}{5.0m,10$^{\circ}$} & \multicolumn{1}{l|}{0.25m,2$^{\circ}$} & \multicolumn{1}{l|}{0.50m,5$^{\circ}$} & \multicolumn{1}{l|}{5.0m,10$^{\circ}$} \bigstrut\\
    \hline
    NetVLAD \cite{ref5.12} & 17.4 & 40.3 & 93.2 & 7.7 & 21 & 80.5 \bigstrut\\
    \hline
    HF-Net \cite{ref5.1} & 90.4 & 93.1 & 96.1 & 71.8 & 78.2 & 87.1 \bigstrut\\
    \hline
    \end{tabular}}%
  \label{tab16}%
\end{table}%

\begin{table}[H]
  \centering
  \caption{A summary of published results of APR and RPR methods on other datasets}
  \resizebox{\textwidth}{!}{
    \begin{tabular}{|c|c|c|c|c|c|}
    \hline
    Methods & \multirow{2}[4]{*}{TUM-LSI} & \multirow{2}[4]{*}{Dubrovnki 6K} & ApolloScape &   &  \bigstrut\\
\cline{1-1}\cline{4-6}    Scene &   &   & road11 & road12 & generalized \bigstrut\\
    \hline
    PoseNet \cite{ref7.2} & / & / & 13.85m, 3.49$^{\circ}$ & 11.24m, 3.55$^{\circ}$ & / \bigstrut\\
    \hline
    LSTM PoseNet \cite{ref7.4} & 1.31m, 2.79$^{\circ}$ & / & / & / & / \bigstrut\\
    \hline
    Geo.PoseNet \cite{ref7.7} & / & 9.88m, 4.73$^{\circ}$ & / & / & / \bigstrut\\
    \hline
    MapNet \cite{ref8.1} & / & / & 8.30m, 2.77$^{\circ}$ & 6.83m, 2.72$^{\circ}$ & / \bigstrut\\
    \hline
    NNnet \cite{ref9.1} & / & / & 6.90m, 3.28$^{\circ}$ & 6.34m, 3.33$^{\circ}$ & 16.60m, 3.49$^{\circ}$ \bigstrut\\
    \hline
    CamNet \cite{ref9.3} & / & / & 5.24m, 2.57$^{\circ}$ & 5.19m, 2.70$^{\circ}$ & 8.63m, 2.97$^{\circ}$ \bigstrut\\
    \hline
    \end{tabular}}%
  \label{tab17}%
\end{table}%

Table \ref{tab10}-\ref{tab17} illustrates the published results for some common benchmarks, e.g., 7Scenes \cite{ref7.2}, Cambridge \cite{ref7.2}, Oxford RobotCar \cite{ref10.2}, TUM LSI \cite{ref7.4}, Dubrovnik 6K \cite{ref10.4}, ApolloScape \cite{ref10.5}, Aachen \cite{ref3.5}, CMU \cite{ref3.5} and InLoc \cite{ref6.31} datasets. Through these results, we can see that structure-based localization methods generally surpass camera pose regression methods. From the absolute pose regression and relative camera pose regression methods results, we see that improving the neural network architecture, optimizing the loss function through adding more information such as providing sequential images or video, adding geometric-aware temporal constraints, semantic constraints, can improve the localization performance. 

\begin{table}[H]
  \centering
  \caption{Structure-based methods qualitative comparison }
  \resizebox{\textwidth}{!}{
    \begin{tabular}{|l|l|l|l|l|l|l|}
    \hline
    Paper & Input & Scene & Structure & \tabincell{l}{Robus\\tness} & \tabincell{l}{Accu\\racy} & \tabincell{l}{Effic\\iency} \bigstrut\\
    \hline
    Output & \multicolumn{6}{c|}{$H = (t,\theta)$} \bigstrut\\
    \hline
    \cite{ref6.45} & RGB-D image & 3D model & \tabincell{l}{Random Forest \\+RANSAC (PnP)} & $\checkmark$ &   &  \bigstrut\\
    \hline
    DSAC \cite{ref6.46} & RGB image patches & 3D model & CNN + DSAC &   &   &  \bigstrut\\
    \hline
    DSAC++ \cite{ref6.33} & RGB image patches & \tabincell{l}{3D model\\or not} & CNN + DSAC++ &   & $\checkmark$ &  \bigstrut\\
    \hline
    ESAC \cite{ref6.47} & RGB image patches & 3D model & CNN + ESAC & $\checkmark$ &   &  \bigstrut\\
    \hline
    \cite{ref6.49} & RGB image & 3D model & CNN + DSAC &   &   & $\checkmark$ \bigstrut\\
    \hline
    \tabincell{l}{NG-RANSAC\\\cite{ref6.48}} & RGB image & 3D model & \tabincell{l}{CNN +\\NG-RANSAC} & $\checkmark$ &   &  \bigstrut\\
    \hline
    HSC-NET \cite{ref6.50} & RGB image & 3D model & CNN + DSAC & $\checkmark$ & $\checkmark$ &  \bigstrut\\
    \hline
    HF-Net \cite{ref5.1} & RGB image & 3D model & \tabincell{l}{CNN + NN\\+ RANSAC} & $\checkmark$ & $\checkmark$ & $\checkmark$ \bigstrut\\
    \hline
    InLoc \cite{ref6.31} & RGB image & 3D model &   &   & $\checkmark$ &  \bigstrut\\
    \hline
    NetVLAD \cite{ref5.12} & RGB image & 3D model & \tabincell{l}{CNN + NN +\\RANSAC (PnP)} &   &   & $\checkmark$ \bigstrut\\
    \hline
    \end{tabular}}%
  \label{tab18}%
\end{table}%

\begin{table}[H]
  \centering
  \caption{Publication information, training model and ground truth label of structure-based methods}
  \resizebox{\textwidth}{!}{
    \begin{tabular}{|l|l|l|l|l|l|l|}
    \hline
    \multirow{2}[4]{*}{Task} & \multirow{2}[4]{*}{Method} & \multirow{2}[4]{*}{Year-Pub.-Cited} & \multirow{2}[4]{*}{Training dataset} & \multirow{2}[4]{*}{\tabincell{l}{Stru\\cture\\-based}} & \multicolumn{2}{c|}{\tabincell{l}{Ground truth\\label}} \bigstrut\\
\cline{6-7}      &   &   &   &   & \tabincell{l}{Loc.\\Corres} & \tabincell{l}{Image\\pair} \bigstrut\\
    \hline
    \multirow{2}[4]{*}{\tabincell{l}{Keypoint\\detection}} & \tabincell{l}{QuadNet\\\cite{ref6.7}} & 2017-CVPR-76 & DTU robot image & \tabincell{l}{3D\\model} & $\checkmark$ &  \bigstrut\\
\cline{2-7}      & \tabincell{l}{Key.Net\\\cite{ref6.51}} & 2019-ICCV-41 & \tabincell{l}{ImageNet ILSVRC\\2012} & Self &   & $\checkmark$ \bigstrut\\
    \hline
    \multirow{2}[4]{*}{\tabincell{l}{Descriptor\\learning}} & \tabincell{l}{HardNet\\\cite{ref6.15}} & 2017-NIPS-222 & \tabincell{l}{UBC/\\Brown dataset} & MVS & $\checkmark$ &  \bigstrut\\
\cline{2-7}      & \tabincell{l}{SOSNet\\\cite{ref6.16}} & 2019-CVPR-57 & \tabincell{l}{UBC/\\Brown dataset} & MVS & $\checkmark$ &  \bigstrut\\
    \hline
    \multirow{5}[10]{*}{\tabincell{l}{Matching\\pipeline}} & \tabincell{l}{LIFT\\\cite{ref6.18}} & 2016-ECCV-601 & Piccadilly Circus & SFM & $\checkmark$ & $\checkmark$ \bigstrut\\
\cline{2-7}      & \tabincell{l}{SuperPoint\\\cite{ref6.9}} & 2018-CVPR-341 & MS-COCO & Self & $\checkmark$ & $\checkmark$ \bigstrut\\
\cline{2-7}      & \tabincell{l}{LF-Net\\\cite{ref6.19}} & 2018-NIPS-140 & \tabincell{l}{ScanNet,\\25photo-tourism} & SFM & $\checkmark$ & $\checkmark$ \bigstrut\\
\cline{2-7}      & D2-Net \cite{ref6.23} & 2019-CVPR-73 & MegaDepth & SFM & $\checkmark$ & $\checkmark$ \bigstrut\\
\cline{2-7}      & \tabincell{l}{R2D2\\\cite{ref6.21}} & 2019-NIPS-67 & Aachen & \tabincell{l}{SFM/\\Flow/\\Style} & $\checkmark$ & $\checkmark$ \bigstrut\\
    \hline
    \multirow{2}[4]{*}{Retrieval} & \tabincell{l}{NetVLAD\\\cite{ref5.12}} & 2016-CVPR-1249 & \tabincell{l}{Google street\\wheel} & T.M. &   & $\checkmark$ \bigstrut\\
\cline{2-7}      & GEM \cite{ref5.14} & 2018-TPAMI-318 & SFM-120k & SFM &   & $\checkmark$ \bigstrut\\
    \hline
    \multirow{3}[6]{*}{Multi-task} & DELF \cite{ref5.13} & 2017-ICCV-304 & Landmark dataset & Class &   &  \bigstrut\\
\cline{2-7}      & \tabincell{l}{HF-Net\\\cite{ref5.1}} & 2019-CVPR-114 & \tabincell{l}{Google landmark,\\BDD} & Teacher & $\checkmark$ & $\checkmark$ \bigstrut\\
\cline{2-7}      & \tabincell{l}{ContextDesc\\\cite{ref6.52}} & 2019-CVPR-45 & \tabincell{l}{Photo-tourism,\\aerial dataset} & SFM & $\checkmark$ & $\checkmark$ \bigstrut\\
    \hline
    \end{tabular}}%
  \label{tab19}%
\end{table}%

\subsection{Comparison of Real-World Applicability}
\label{subsec6.3}
Generally, structure-based methods achieve a higher accuracy compared to end-to-end methods including APR. Some RPR methods keep pursuing the accuracy of structure-based methods with a less complex pipeline. A retrieval-based method has the worst effect because it uses the retrieved image pose as the calculated value.

\begin{center}
    \begin{longtable}{|p{5.5em}|p{11em}|p{4em}|p{2em}|p{2em}|p{2em}|}
    \caption{APR and RPR methods qualitative comparison} 
    \label{tab20} \\
    \hline
    Methods & Robustness & \tabincell{l}{Required\\resources} & \tabincell{l}{Time\\(ms)} & \tabincell{l}{Size\\(Mb)} & \tabincell{l}{Impl\\ement\\ation} \bigstrut\\
    \hline
    \tabincell{l}{PoseNet\\\cite{ref7.2}} & \tabincell{l}{Lighting, motion blur,\\different camera intrinsics} & \tabincell{l}{Nvidia\\Titan\\black} & 5-95 & \multicolumn{1}{l|}{50} & $\checkmark$ \bigstrut\\
    \hline
    \tabincell{l}{Bayesian\\PoseNet \cite{ref7.3}} & \tabincell{l}{Large viewpoint or\\appearance changes} & \tabincell{l}{Nvidia\\Titan X} & \multicolumn{1}{l|}{6} & \multicolumn{1}{l|}{50} & $\checkmark$ \bigstrut\\
    \hline
    \tabincell{l}{LSTM\\PoseNet \cite{ref7.5}} & \tabincell{l}{Motion blur and\\illumination changes} & \tabincell{l}{Nvidia\\Titan X} & \multicolumn{1}{l|}{9.2} & / & / \bigstrut\\
    \hline
    \tabincell{l}{Hourglass\\PoseNet \cite{ref7.5}} & \tabincell{l}{Continuous pose\\optimization} & \tabincell{l}{Nvidia\\Titan X} & / & / & $\checkmark$ \bigstrut\\
    \hline
    \tabincell{l}{SVS\\PoseNet \cite{ref7.1}} & \tabincell{l}{Large translational\\deviations of the camera\\along with the depth\\of the scene} & \tabincell{l}{Nvidia\\Titan X} & \multicolumn{1}{l|}{12.5} & \multicolumn{1}{l|}{10} & / \bigstrut\\
    \hline
    \tabincell{l}{BranchNet\\\cite{ref7.6}} & \tabincell{l}{Ambiguities, motion-blur,\\flat surfaces, lighting} & \tabincell{l}{Nvidia\\Titan X} & \multicolumn{1}{l|}{6} & \multicolumn{1}{l|}{46} & / \bigstrut\\
    \hline
    \tabincell{l}{Geo.PoseNet\\(geo.loss)\\/Geo.PoseNet \\(reprojection\\error loss)\\\cite{ref7.7}} & \tabincell{l}{Lighting, motion blur,\\unknown camera intrinsics} & \tabincell{l}{Nvidia\\Titan X} & \multicolumn{1}{l|}{5} & / & $\checkmark$ \bigstrut\\
    \hline
    \tabincell{l}{AtLoc/\\AtLocPlus\\\cite{ref7.8}} & \tabincell{l}{Dynamic objects,\\illumination} & \tabincell{l}{Nvidia\\Titan X} & \multicolumn{1}{l|}{6.3} & / & $\checkmark$ \bigstrut\\
    \hline
    AdPR \cite{ref7.9} & \tabincell{l}{Motion blur, repeating\\structures, texture-less\\surfaces} & \tabincell{l}{Nvidia\\GeForce\\RTX 2080} & \multicolumn{1}{l|}{50} & / & / \bigstrut\\
    \hline
    PVL \cite{ref7.10} & Dynamic environments & / & / & / & $\checkmark$ \bigstrut\\
    \hline
    APANet \cite{ref7.11} & Lighting, viewpoint & / & / & / & / \bigstrut\\
    \hline
    SPPNet \cite{ref7.12} & \tabincell{l}{Unevenly distributed\\image features} & \tabincell{l}{Nvidia\\Titan X} & \multicolumn{1}{l|}{2} & \multicolumn{1}{l|}{36.8} & $\checkmark$ \bigstrut\\
    \hline
    GPoseNet \cite{ref7.13} & Choice of hyperparameters & \tabincell{l}{Nvidia\\GeForce\\GTX 1070} & / & / & / \bigstrut\\
    \hline
    MapNet \cite{ref8.1} & \tabincell{l}{Online, locally smooth\\and drift-free} & / & \multicolumn{1}{l|}{9.4} & / & / \bigstrut\\
    \hline
    LSG [8.2] & \tabincell{l}{Pose uncertainties by\\content augmentation} & \tabincell{l}{Nvidia\\1080Ti} & / & / & / \bigstrut\\
    \hline
    VlocNet \cite{ref8.3} & \tabincell{l}{Environment, dynamic\\objects, structure} & \tabincell{l}{Nvidia\\Titan X} & / & / & / \bigstrut\\
    \hline
    VlocNet++ \cite{ref8.4} & \tabincell{l}{Noise, camera angle\\deviations, object scale,\\frame-level distortions} & \tabincell{l}{Nvidia\\Titan X} & \multicolumn{1}{l|}{79} & / & / \bigstrut\\
    \hline
    DGRNet \cite{ref8.5} & \tabincell{l}{Camera parameters,\\challenging environments} & \tabincell{l}{Nvidia\\GTX 1080} & 45-65 & / & / \bigstrut\\
    \hline
    VidLoc \cite{ref8.6} & Temporal smoothness & \tabincell{l}{Titan X\\Pascal} & 18-43 & / & $\checkmark$ \bigstrut\\
    \hline
    NNnet \cite{ref9.1} & Pose filtering & \tabincell{l}{Nvidia\\Titan X} & / & / & $\checkmark$ \bigstrut\\
    \hline
    RelocNet \cite{ref9.2} & Pose retrieval descriptors & / & / & / & / \bigstrut\\
    \hline
    CamNet \cite{ref9.3} & 2D-3D matching & \tabincell{l}{Nvidia\\Titan XP} & / & / & / \bigstrut\\
    \hline
    \tabincell{l}{To learn or\\not to learn\\\cite{ref9.4}} & Outlier pairs & \tabincell{l}{Nvidia\\Titan XP} & / & / & / \bigstrut\\
    \hline
    Relative NN \cite{ref9.5} & \tabincell{l}{Repetitive structures,\\textureless objects,\\large viewpoint changes} & \tabincell{l}{Nvidia\\Titan X} & / & / & / \bigstrut\\
    \hline
    AnchorNet \cite{ref9.6} & Relative anchor point & / & / & / & / \bigstrut\\
\hline 
\end{longtable}%
\end{center}

Structure feature-based localization methods rely on the correspondence between the 2D query feature and the 3D model. Matching based methods establish the correspondences between 3D point cloud in scene model and 2D query image feature by matching feature descriptor. Camera localization algorithm applications are quite mature. But they are still fragile for those repetitive local features and there remains the issue that they have a high computational requirement. Most of the work focuses on robust feature points or accurate feature descriptors under extreme conditions in 2D images. We should consider the constraints of 3D spatial geometry for the 2D feature in further complex applications, e.g., most existing scene coordinate regression methods can only be adopted on small-scale scenes. They have not yet proven their capacity in large-scale scenes. For robustness or precision, accuracy or efficiency, we need to think carefully about the trade-offs. Table 19 shows a summary of some published structure-based methods. We summarized their input format and output result $H$, in which $t$ and $\theta$ mean translation and orientation respectively. According to their innovation, we divided those approaches into three improved directions, robustness, accuracy and efficiency. Table 20 lists a general summary of comparisons based on publication information, training model and ground-truth label of structure-based methods.

Robustness and accuracy are the most important criteria for detecting positioning performance. The stronger the adaptability and robustness to changes in the scene environment, the better the positioning performance. Table 21 shows the qualitative comparison between APR and RPR methods, including robustness, time to process each picture, computing resources required for positioning tasks, scene model size, and whether there is public code implementation. 

\section{Summary and Concluding Remarks}
\label{sec7}
\subsection{Summary}
Remarkable achievements have been made in multiple research branches of image-based camera localization. This paper reviews the main branches and describes their methods, datasets, metrics, and gives detailed statistics for quantitative and qualitative comparisons of these methods. 

\subsection{Future Potential Research Directions}
The future research directions for image-based camera localization are proposed as follows:

\subsubsection{Sensor fusion}
Rather than just being limited to the image data output by a camera sensor, other sensors that can obtain more extensive positioning information, such as LiDAR, WiFi, IMU, Bluetooth, etc., could be used. An effective complementary combination of different sensor information can help build a more accurate and powerful positioning system. To achieve this, multiple sensor data fusion needs to overcome the challenge of heterogeneous characteristics of multiple sensor data. This kind of multi-modal task can merge different features by learning the joint contribution of every single task, or through learning the cross-coding between different tasks for more efficient positioning.

\subsubsection{Multi-features}
Most of the structure-based work mentioned above is based on 3D point features extracted from 2D scene images. It is limited under some challenging conditions, such as weak texture, illumination, weather, etc. But we can also use multi-features to improve localization. For example, some SLAM researchers used line or plane features to estimate camera pose and get a very impressive result. Maybe we can use CNN to improve the process of extracting line or plane features, and then design an efficient matching pipeline for all kinds of features. 

\subsubsection{Semantic information auxiliaries}
Another direction is to use the semantics of features. Semantic plays an important role in real-world scenes \cite{ref3.5}. With the help of semantic information, we can easily filter out those features on dynamic objects that affect the localization result. And we can also verify the location by comparing the semantic information between scene and query image which means we need to build a semantic 3D scene model. Semantic information is an upper-level feature. Constructing a semantic three-dimensional model of the scene can not only assist in localization but also be useful for the wider application level.

\subsubsection{Multi-cameras}
Compared with a single camera, a multi-camera system can cover a panoramic $360^{\circ}$ field of view and can significantly improve performance in robotic applications such as camera positioning. However, multiple images output by multiple cameras requires feature calculation during feature matching. One of the future research and development directions may be to use deep learning to resolve the features of multiple cameras end-to-end and speed up the feature calculation time.

\subsubsection{Challenging conditions}
In many specific positioning tasks, how to effectively improve positioning performance in challenging scenarios is a key step in building an accurate positioning system. For example, environmental changes caused by illumination, blur, and occlusion changes make the feature extraction of images less accurate. Further, the position changes of dynamic objects between different frames of images also cause interference to matching between images. In the future, we can explore methods such as using the relationship between multiple frames of images and adding other auxiliary judgment information to solve these challenges to improve the robustness of the positioning model in a variety of scenarios.

\subsubsection{Integration in light-weight devices}
In the future, applications that use camera positioning assistance will develop towards being small and lightweight, which will be more quickly and conveniently applied to small portable devices, such as embedded devices or mobile phones. In addition, algorithm applications on small ICT resource devices could use methods such as model acceleration to optimize model calculation speed, model size, and consumption of computing resources, to better integrate with device functions to serve applications such as navigation, sports, teaching, entertainment. In addition, the use of mobile phones and other devices with cameras and other sensor fusions requires the development of more augmented reality functions to serve smart navigation, guided services, and immersive games.

\section{Acknowledgement}

This research was funded in part by a PhD scholarship funded jointly by the China Scholarship Council (CSC) and QMUL and partly funded under Didi Chuxing and the Robotics and AI for Extreme Environments program’s NCNR (National Centre for Nuclear Robotics) grant no. EP/R02572X/1.

\bibliography{mybibfile}

\begin{thebibliography}{100}
\expandafter\ifx\csname url\endcsname\relax
  \def\url#1{\texttt{#1}}\fi
\expandafter\ifx\csname urlprefix\endcsname\relax\def\urlprefix{URL }\fi
\expandafter\ifx\csname href\endcsname\relax
  \def\href#1#2{#2} \def\path#1{#1}\fi

\bibitem{ref2.1}
N.~Piasco, D.~Sidib{\'e}, C.~Demonceaux, V.~Gouet-Brunet, A survey on
  visual-based localization: On the benefit of heterogeneous data, Pattern
  Recognition 74 (2018) 90--109.

\bibitem{ref2.2}
T.~Sattler, Q.~Zhou, M.~Pollefeys, L.~Leal-Taixe, Understanding the limitations
  of cnn-based absolute camera pose regression, in: Proceedings of the IEEE/CVF
  conference on computer vision and pattern recognition, 2019, pp. 3302--3312.

\bibitem{ref2.3}
Y.~Wu, F.~Tang, H.~Li, Image-based camera localization: an overview, Visual
  Computing for Industry, Biomedicine, and Art 1~(1) (2018) 1--13.

\bibitem{ref2.4}
C.~Debeunne, D.~Vivet, A review of visual-lidar fusion based simultaneous
  localization and mapping, Sensors 20~(7) (2020) 2068.

\bibitem{ref2.5}
C.~Chen, B.~Wang, C.~X. Lu, N.~Trigoni, A.~Markham, A survey on deep learning
  for localization and mapping: Towards the age of spatial machine
  intelligence, arXiv preprint arXiv:2006.12567.

\bibitem{ref2.6}
Y.~Shavit, R.~Ferens, Introduction to camera pose estimation with deep
  learning, arXiv preprint arXiv:1907.05272.

\bibitem{ref3.1}
N.~Snavely, S.~M. Seitz, R.~Szeliski, Photo tourism: exploring photo
  collections in 3d, ACM siggraph 2006 papers (2006) 835--846.

\bibitem{ref3.2}
J.~L. Schonberger, J.-M. Frahm, Structure-from-motion revisited, in:
  Proceedings of the IEEE conference on computer vision and pattern
  recognition, 2016, pp. 4104--4113.

\bibitem{ref3.3}
W.~Changchang, Visualsfm: A visual structure from motion system (2011).

\bibitem{ref3.4}
J.~Sturm, N.~Engelhard, F.~Endres, W.~Burgard, D.~Cremers, A benchmark for the
  evaluation of rgb-d slam systems, in: 2012 IEEE/RSJ international conference
  on intelligent robots and systems, IEEE, 2012, pp. 573--580.

\bibitem{ref3.5}
T.~Sattler, W.~Maddern, C.~Toft, A.~Torii, L.~Hammarstrand, E.~Stenborg,
  D.~Safari, M.~Okutomi, M.~Pollefeys, J.~Sivic, et~al., Benchmarking 6dof
  outdoor visual localization in changing conditions, in: Proceedings of the
  IEEE Conference on Computer Vision and Pattern Recognition, 2018, pp.
  8601--8610.

\bibitem{ref3.6}
Z.~Zhang, T.~Sattler, D.~Scaramuzza, Reference pose generation for long-term
  visual localization via learned features and view synthesis, International
  Journal of Computer Vision 129~(4) (2021) 821--844.

\bibitem{ref4.1}
M.~A. Fischler, R.~C. Bolles, Random sample consensus: a paradigm for model
  fitting with applications to image analysis and automated cartography,
  Communications of the ACM 24~(6) (1981) 381--395.

\bibitem{ref4.2}
D.~P. Robertson, R.~Cipolla, An image-based system for urban navigation., in:
  Bmvc, Vol.~19, Citeseer, 2004, p. 165.

\bibitem{ref4.3}
W.~Zhang, J.~Kosecka, Image based localization in urban environments, in: Third
  international symposium on 3D data processing, visualization, and
  transmission (3DPVT'06), IEEE, 2006, pp. 33--40.

\bibitem{ref4.4}
Y.~Li, N.~Snavely, D.~P. Huttenlocher, Location recognition using prioritized
  feature matching, in: European conference on computer vision, Springer, 2010,
  pp. 791--804.

\bibitem{ref4.5}
T.~Sattler, B.~Leibe, L.~Kobbelt, Fast image-based localization using direct
  2d-to-3d matching, in: 2011 International Conference on Computer Vision,
  IEEE, 2011, pp. 667--674.

\bibitem{ref4.6}
T.~Sattler, B.~Leibe, L.~Kobbelt, Improving image-based localization by active
  correspondence search, in: European conference on computer vision, Springer,
  2012, pp. 752--765.

\bibitem{ref4.24}
D.~G. Lowe, Distinctive image features from scale-invariant keypoints,
  International journal of computer vision 60~(2) (2004) 91--110.

\bibitem{ref4.7}
T.~Sattler, M.~Havlena, F.~Radenovic, K.~Schindler, M.~Pollefeys, Hyperpoints
  and fine vocabularies for large-scale location recognition, in: Proceedings
  of the IEEE International Conference on Computer Vision, 2015, pp.
  2102--2110.

\bibitem{ref4.8}
D.~G. Lowe, Distinctive image features from scale-invariant keypoints,
  International journal of computer vision 60~(2) (2004) 91--110.

\bibitem{ref4.9}
Y.~Feng, L.~Fan, Y.~Wu, Fast localization in large-scale environments using
  supervised indexing of binary features, IEEE Transactions on Image Processing
  25~(1) (2015) 343--358.

\bibitem{ref4.25}
E.~Rosten, T.~Drummond, Machine learning for high-speed corner detection, in:
  European conference on computer vision, Springer, 2006, pp. 430--443.

\bibitem{ref4.10}
L.~Liu, H.~Li, Y.~Dai, Efficient global 2d-3d matching for camera localization
  in a large-scale 3d map, in: Proceedings of the IEEE International Conference
  on Computer Vision, 2017, pp. 2372--2381.

\bibitem{ref4.11}
H.~Tong, C.~Faloutsos, J.-Y. Pan, Fast random walk with restart and its
  applications, in: Sixth international conference on data mining (ICDM'06),
  IEEE, 2006, pp. 613--622.

\bibitem{ref4.13}
L.~Svarm, O.~Enqvist, M.~Oskarsson, F.~Kahl, Accurate localization and pose
  estimation for large 3d models, in: Proceedings of the IEEE Conference on
  Computer Vision and Pattern Recognition, 2014, pp. 532--539.

\bibitem{ref4.15}
L.~Sv{\"a}rm, O.~Enqvist, F.~Kahl, M.~Oskarsson, City-scale localization for
  cameras with known vertical direction, IEEE transactions on pattern analysis
  and machine intelligence 39~(7) (2016) 1455--1461.

\bibitem{ref4.14}
B.~Zeisl, T.~Sattler, M.~Pollefeys, Camera pose voting for large-scale
  image-based localization, in: Proceedings of the IEEE International
  Conference on Computer Vision, 2015, pp. 2704--2712.

\bibitem{ref4.16}
A.~E. Johnson, M.~Hebert, Using spin images for efficient object recognition in
  cluttered 3d scenes, IEEE Transactions on pattern analysis and machine
  intelligence 21~(5) (1999) 433--449.

\bibitem{ref4.17}
R.~B. Rusu, N.~Blodow, M.~Beetz, Fast point feature histograms (fpfh) for 3d
  registration, in: 2009 IEEE international conference on robotics and
  automation, IEEE, 2009, pp. 3212--3217.

\bibitem{ref4.18}
S.~Choi, Q.-Y. Zhou, V.~Koltun, Robust reconstruction of indoor scenes, in:
  Proceedings of the IEEE Conference on Computer Vision and Pattern
  Recognition, 2015, pp. 5556--5565.

\bibitem{ref4.19}
Z.~Wu, S.~Song, A.~Khosla, F.~Yu, L.~Zhang, X.~Tang, J.~Xiao, 3d shapenets: A
  deep representation for volumetric shapes, in: Proceedings of the IEEE
  conference on computer vision and pattern recognition, 2015, pp. 1912--1920.

\bibitem{ref4.20}
Y.~Fang, J.~Xie, G.~Dai, M.~Wang, F.~Zhu, T.~Xu, E.~Wong, 3d deep shape
  descriptor, in: Proceedings of the IEEE Conference on Computer Vision and
  Pattern Recognition, 2015, pp. 2319--2328.

\bibitem{ref4.21}
S.~Song, J.~Xiao, Deep sliding shapes for amodal 3d object detection in rgb-d
  images, in: Proceedings of the IEEE conference on computer vision and pattern
  recognition, 2016, pp. 808--816.

\bibitem{ref4.22}
K.~Guo, D.~Zou, X.~Chen, 3d mesh labeling via deep convolutional neural
  networks, ACM Transactions on Graphics (TOG) 35~(1) (2015) 1--12.

\bibitem{ref4.23}
A.~Zeng, S.~Song, M.~Nie{\ss}ner, M.~Fisher, J.~Xiao, T.~Funkhouser, 3dmatch:
  Learning local geometric descriptors from rgb-d reconstructions, in:
  Proceedings of the IEEE conference on computer vision and pattern
  recognition, 2017, pp. 1802--1811.

\bibitem{ref5.1}
P.-E. Sarlin, C.~Cadena, R.~Siegwart, M.~Dymczyk, From coarse to fine: Robust
  hierarchical localization at large scale, in: Proceedings of the IEEE/CVF
  Conference on Computer Vision and Pattern Recognition, 2019, pp.
  12716--12725.

\bibitem{ref5.2}
A.~Irschara, C.~Zach, J.-M. Frahm, H.~Bischof, From structure-from-motion point
  clouds to fast location recognition, in: 2009 IEEE Conference on Computer
  Vision and Pattern Recognition, IEEE, 2009, pp. 2599--2606.

\bibitem{ref5.19}
A.~Irschara, C.~Zach, H.~Bischof, Towards wiki-based dense city modeling, in:
  2007 ieee 11th international conference on computer vision, IEEE, 2007, pp.
  1--8.

\bibitem{ref5.3}
J.~Philbin, O.~Chum, M.~Isard, J.~Sivic, A.~Zisserman, Object retrieval with
  large vocabularies and fast spatial matching, in: 2007 IEEE conference on
  computer vision and pattern recognition, IEEE, 2007, pp. 1--8.

\bibitem{ref5.4}
Y.~Avrithis, Y.~Kalantidis, Approximate gaussian mixtures for large scale
  vocabularies, in: European Conference on Computer Vision, Springer, 2012, pp.
  15--28.

\bibitem{ref5.5}
X.~Shen, Z.~Lin, J.~Brandt, Y.~Wu, Spatially-constrained similarity measurefor
  large-scale object retrieval, IEEE transactions on pattern analysis and
  machine intelligence 36~(6) (2013) 1229--1241.

\bibitem{ref5.6}
O.~Chum, A.~Mikulik, M.~Perdoch, J.~Matas, Total recall ii: Query expansion
  revisited, in: CVPR 2011, IEEE, 2011, pp. 889--896.

\bibitem{ref5.7}
G.~Tolias, H.~J{\'e}gou, Visual query expansion with or without geometry:
  refining local descriptors by feature aggregation, Pattern recognition
  47~(10) (2014) 3466--3476.

\bibitem{ref5.22}
K.~Kesorn, S.~Poslad, An enhanced bag-of-visual word vector space model to
  represent visual content in athletics images, IEEE Transactions on Multimedia
  14~(1) (2011) 211--222.
\newblock \href {http://dx.doi.org/10.1109/TMM.2011.2170665}
  {\path{doi:10.1109/TMM.2011.2170665}}.

\bibitem{ref5.8}
A.~Krizhevsky, I.~Sutskever, G.~E. Hinton, Imagenet classification with deep
  convolutional neural networks, Advances in neural information processing
  systems 25 (2012) 1097--1105.

\bibitem{ref5.9}
J.~Donahue, Y.~Jia, O.~Vinyals, J.~Hoffman, N.~Zhang, E.~Tzeng, T.~Darrell,
  Decaf: A deep convolutional activation feature for generic visual
  recognition, in: International conference on machine learning, PMLR, 2014,
  pp. 647--655.

\bibitem{ref5.10}
A.~Sharif~Razavian, H.~Azizpour, J.~Sullivan, S.~Carlsson, Cnn features
  off-the-shelf: an astounding baseline for recognition, in: Proceedings of the
  IEEE conference on computer vision and pattern recognition workshops, 2014,
  pp. 806--813.

\bibitem{ref5.11}
F.~R. G. T.~O. Chum, Cnn image retrieval learns from bow: Unsupervised
  fine-tuning with hard examples, IEEE Transaction on Image Processing.

\bibitem{ref5.12}
R.~Arandjelovic, P.~Gronat, A.~Torii, T.~Pajdla, J.~Sivic, Netvlad: Cnn
  architecture for weakly supervised place recognition, in: Proceedings of the
  IEEE conference on computer vision and pattern recognition, 2016, pp.
  5297--5307.

\bibitem{ref5.20}
G.~Amato, P.~Bolettieri, F.~Falchi, C.~Gennaro, Large scale image retrieval
  using vector of locally aggregated descriptors, in: International Conference
  on Similarity Search and Applications, Springer, 2013, pp. 245--256.

\bibitem{ref5.14}
F.~Radenovi{\'c}, G.~Tolias, O.~Chum, Fine-tuning cnn image retrieval with no
  human annotation, IEEE transactions on pattern analysis and machine
  intelligence 41~(7) (2018) 1655--1668.

\bibitem{ref5.21}
T.~Dai, J.~Cai, Y.~Zhang, S.-T. Xia, L.~Zhang, Second-order attention network
  for single image super-resolution, in: Proceedings of the IEEE/CVF Conference
  on Computer Vision and Pattern Recognition, 2019, pp. 11065--11074.

\bibitem{ref5.16}
J.~Revaud, J.~Almaz{\'a}n, R.~S. Rezende, C.~R.~d. Souza, Learning with average
  precision: Training image retrieval with a listwise loss, in: Proceedings of
  the IEEE/CVF International Conference on Computer Vision, 2019, pp.
  5107--5116.

\bibitem{ref5.13}
H.~Noh, A.~Araujo, J.~Sim, T.~Weyand, B.~Han, Large-scale image retrieval with
  attentive deep local features, in: Proceedings of the IEEE international
  conference on computer vision, 2017, pp. 3456--3465.

\bibitem{ref5.17}
M.~Teichmann, A.~Araujo, M.~Zhu, J.~Sim, Detect-to-retrieve: Efficient regional
  aggregation for image search, in: Proceedings of the IEEE/CVF Conference on
  Computer Vision and Pattern Recognition, 2019, pp. 5109--5118.

\bibitem{ref5.18}
S.~S. Husain, M.~Bober, Remap: Multi-layer entropy-guided pooling of dense cnn
  features for image retrieval, IEEE Transactions on Image Processing 28~(10)
  (2019) 5201--5213.

\bibitem{ref6.1}
E.~Rosten, T.~Drummond, Machine learning for high-speed corner detection, in:
  European conference on computer vision, Springer, 2006, pp. 430--443.

\bibitem{ref6.2}
S.~Leutenegger, M.~Chli, R.~Y. Siegwart, Brisk: Binary robust invariant
  scalable keypoints, in: 2011 International conference on computer vision,
  Ieee, 2011, pp. 2548--2555.

\bibitem{ref6.3}
E.~Rublee, V.~Rabaud, K.~Konolige, G.~Bradski, Orb: An efficient alternative to
  sift or surf, in: 2011 International conference on computer vision, Ieee,
  2011, pp. 2564--2571.

\bibitem{ref6.4}
Y.~Verdie, K.~Yi, P.~Fua, V.~Lepetit, Tilde: A temporally invariant learned
  detector, in: Proceedings of the IEEE conference on computer vision and
  pattern recognition, 2015, pp. 5279--5288.

\bibitem{ref6.5}
K.~Lenc, A.~Vedaldi, Learning covariant feature detectors, in: European
  conference on computer vision, Springer, 2016, pp. 100--117.

\bibitem{ref6.6}
X.~Zhang, F.~X. Yu, S.~Karaman, S.-F. Chang, Learning discriminative and
  transformation covariant local feature detectors, in: Proceedings of the IEEE
  conference on computer vision and pattern recognition, 2017, pp. 6818--6826.

\bibitem{ref6.7}
N.~Savinov, A.~Seki, L.~Ladicky, T.~Sattler, M.~Pollefeys, Quad-networks:
  unsupervised learning to rank for interest point detection, in: Proceedings
  of the IEEE conference on computer vision and pattern recognition, 2017, pp.
  1822--1830.

\bibitem{ref6.8}
D.~DeTone, T.~Malisiewicz, A.~Rabinovich, Toward geometric deep slam, arXiv
  preprint arXiv:1707.07410.

\bibitem{ref6.9}
D.~DeTone, T.~Malisiewicz, A.~Rabinovich, Superpoint: Self-supervised interest
  point detection and description, in: Proceedings of the IEEE conference on
  computer vision and pattern recognition workshops, 2018, pp. 224--236.

\bibitem{ref6.10}
S.~Zagoruyko, N.~Komodakis, Learning to compare image patches via convolutional
  neural networks, in: Proceedings of the IEEE conference on computer vision
  and pattern recognition, 2015, pp. 4353--4361.

\bibitem{ref6.11}
X.~Han, T.~Leung, Y.~Jia, R.~Sukthankar, A.~C. Berg, Matchnet: Unifying feature
  and metric learning for patch-based matching, in: Proceedings of the IEEE
  conference on computer vision and pattern recognition, 2015, pp. 3279--3286.

\bibitem{ref6.12}
E.~Simo-Serra, E.~Trulls, L.~Ferraz, I.~Kokkinos, P.~Fua, F.~Moreno-Noguer,
  Discriminative learning of deep convolutional feature point descriptors, in:
  Proceedings of the IEEE international conference on computer vision, 2015,
  pp. 118--126.

\bibitem{ref6.13}
V.~Balntas, E.~Riba, D.~Ponsa, K.~Mikolajczyk, Learning local feature
  descriptors with triplets and shallow convolutional neural networks., in:
  Bmvc, Vol.~1, 2016, p.~3.

\bibitem{ref6.14}
Y.~Tian, B.~Fan, F.~Wu, L2-net: Deep learning of discriminative patch
  descriptor in euclidean space, in: Proceedings of the IEEE conference on
  computer vision and pattern recognition, 2017, pp. 661--669.

\bibitem{ref6.15}
A.~Mishchuk, D.~Mishkin, F.~Radenovic, J.~Matas, Working hard to know your
  neighbor's margins: Local descriptor learning loss, arXiv preprint
  arXiv:1705.10872.

\bibitem{ref6.16}
Y.~Tian, X.~Yu, B.~Fan, F.~Wu, H.~Heijnen, V.~Balntas, Sosnet: Second order
  similarity regularization for local descriptor learning, in: Proceedings of
  the IEEE/CVF Conference on Computer Vision and Pattern Recognition, 2019, pp.
  11016--11025.

\bibitem{ref6.17}
Q.~Wang, X.~Zhou, B.~Hariharan, N.~Snavely, Learning feature descriptors using
  camera pose supervision, in: European Conference on Computer Vision,
  Springer, 2020, pp. 757--774.

\bibitem{ref6.18}
K.~M. Yi, E.~Trulls, V.~Lepetit, P.~Fua, Lift: Learned invariant feature
  transform, in: European conference on computer vision, Springer, 2016, pp.
  467--483.

\bibitem{ref6.19}
Y.~Ono, E.~Trulls, P.~Fua, K.~M. Yi, Lf-net: Learning local features from
  images, arXiv preprint arXiv:1805.09662.

\bibitem{ref6.21}
J.~Revaud, P.~Weinzaepfel, C.~De~Souza, N.~Pion, G.~Csurka, Y.~Cabon,
  M.~Humenberger, R2d2: repeatable and reliable detector and descriptor, arXiv
  preprint arXiv:1906.06195.

\bibitem{ref6.22}
Z.~Luo, L.~Zhou, X.~Bai, H.~Chen, J.~Zhang, Y.~Yao, S.~Li, T.~Fang, L.~Quan,
  Aslfeat: Learning local features of accurate shape and localization, in:
  Proceedings of the IEEE/CVF conference on computer vision and pattern
  recognition, 2020, pp. 6589--6598.

\bibitem{ref6.23}
M.~Dusmanu, I.~Rocco, T.~Pajdla, M.~Pollefeys, J.~Sivic, A.~Torii, T.~Sattler,
  D2-net: A trainable cnn for joint description and detection of local
  features, in: Proceedings of the ieee/cvf conference on computer vision and
  pattern recognition, 2019, pp. 8092--8101.

\bibitem{ref6.24}
Y.~Tian, V.~Balntas, T.~Ng, A.~Barroso-Laguna, Y.~Demiris, K.~Mikolajczyk, D2d:
  Keypoint extraction with describe to detect approach, in: Proceedings of the
  Asian Conference on Computer Vision, 2020.

\bibitem{ref6.25}
A.~Benbihi, M.~Geist, C.~Pradalier, Elf: Embedded localisation of features in
  pre-trained cnn, in: Proceedings of the IEEE/CVF International Conference on
  Computer Vision, 2019, pp. 7940--7949.

\bibitem{ref6.26}
H.~Germain, G.~Bourmaud, V.~Lepetit, Sparse-to-dense hypercolumn matching for
  long-term visual localization, in: 2019 International Conference on 3D Vision
  (3DV), IEEE, 2019, pp. 513--523.

\bibitem{ref6.27}
H.~Germain, G.~Bourmaud, V.~Lepetit, S2dnet: Learning accurate correspondences
  for sparse-to-dense feature matching, arXiv preprint arXiv:2004.01673.

\bibitem{ref6.30}
I.~Rocco, M.~Cimpoi, R.~Arandjelovi{\'c}, A.~Torii, T.~Pajdla, J.~Sivic,
  Neighbourhood consensus networks, arXiv preprint arXiv:1810.10510.

\bibitem{ref6.28}
I.~Melekhov, A.~Tiulpin, T.~Sattler, M.~Pollefeys, E.~Rahtu, J.~Kannala,
  Dgc-net: Dense geometric correspondence network, in: 2019 IEEE Winter
  Conference on Applications of Computer Vision (WACV), IEEE, 2019, pp.
  1034--1042.

\bibitem{ref6.29}
O.~Wiles, S.~Ehrhardt, A.~Zisserman, D2d: Learning to find good correspondences
  for image matching and manipulation, arXiv e-prints (2020) arXiv--2007.

\bibitem{ref6.31}
H.~Taira, M.~Okutomi, T.~Sattler, M.~Cimpoi, M.~Pollefeys, J.~Sivic, T.~Pajdla,
  A.~Torii, Inloc: Indoor visual localization with dense matching and view
  synthesis, in: Proceedings of the IEEE Conference on Computer Vision and
  Pattern Recognition, 2018, pp. 7199--7209.

\bibitem{ref6.32}
E.~Rosten, T.~Drummond, Machine learning for high-speed corner detection, in:
  European conference on computer vision, Springer, 2006, pp. 430--443.

\bibitem{ref6.33}
E.~Brachmann, C.~Rother, Learning less is more-6d camera localization via 3d
  surface regression, in: Proceedings of the IEEE Conference on Computer Vision
  and Pattern Recognition, 2018, pp. 4654--4662.

\bibitem{ref6.34}
X.~Li, J.~Ylioinas, J.~Verbeek, J.~Kannala, Scene coordinate regression with
  angle-based reprojection loss for camera relocalization, in: Proceedings of
  the European Conference on Computer Vision (ECCV) Workshops, 2018, pp. 0--0.

\bibitem{ref6.35}
E.~Brachmann, C.~Rother, Visual camera re-localization from rgb and rgb-d
  images using dsac, IEEE Transactions on Pattern Analysis and Machine
  Intelligence.

\bibitem{ref6.36}
M.~Cai, H.~Zhan, C.~Saroj~Weerasekera, K.~Li, I.~Reid, Camera relocalization by
  exploiting multi-view constraints for scene coordinates regression, in:
  Proceedings of the IEEE/CVF International Conference on Computer Vision
  Workshops, 2019, pp. 0--0.

\bibitem{ref6.37}
E.~Brachmann, F.~Michel, A.~Krull, M.~Y. Yang, S.~Gumhold, et~al.,
  Uncertainty-driven 6d pose estimation of objects and scenes from a single rgb
  image, in: Proceedings of the IEEE conference on computer vision and pattern
  recognition, 2016, pp. 3364--3372.

\bibitem{ref6.38}
I.~Budvytis, M.~Teichmann, T.~Vojir, R.~Cipolla, Large scale joint semantic
  re-localisation and scene understanding via globally unique instance
  coordinate regression, arXiv preprint arXiv:1909.10239.

\bibitem{ref6.39}
L.~Yang, Z.~Bai, C.~Tang, H.~Li, Y.~Furukawa, P.~Tan, Sanet: Scene agnostic
  network for camera localization, in: Proceedings of the IEEE/CVF
  International Conference on Computer Vision, 2019, pp. 42--51.

\bibitem{ref7.2}
A.~Kendall, M.~Grimes, R.~Cipolla, Posenet: A convolutional network for
  real-time 6-dof camera relocalization, in: Proceedings of the IEEE
  international conference on computer vision, 2015, pp. 2938--2946.

\bibitem{ref7.3}
A.~Kendall, R.~Cipolla, Modelling uncertainty in deep learning for camera
  relocalization, in: 2016 IEEE international conference on Robotics and
  Automation (ICRA), IEEE, 2016, pp. 4762--4769.

\bibitem{ref7.4}
F.~Walch, C.~Hazirbas, L.~Leal-Taixe, T.~Sattler, S.~Hilsenbeck, D.~Cremers,
  Image-based localization using lstms for structured feature correlation, in:
  Proceedings of the IEEE International Conference on Computer Vision, 2017,
  pp. 627--637.

\bibitem{ref7.8}
B.~Wang, C.~Chen, C.~X. Lu, P.~Zhao, N.~Trigoni, A.~Markham, Atloc: Attention
  guided camera localization, in: Proceedings of the AAAI Conference on
  Artificial Intelligence, Vol.~34, 2020, pp. 10393--10401.

\bibitem{ref7.1}
T.~Naseer, W.~Burgard, Deep regression for monocular camera-based 6-dof global
  localization in outdoor environments, in: 2017 IEEE/RSJ International
  Conference on Intelligent Robots and Systems (IROS), IEEE, 2017, pp.
  1525--1530.

\bibitem{ref7.5}
I.~Melekhov, J.~Ylioinas, J.~Kannala, E.~Rahtu, Image-based localization using
  hourglass networks, in: Proceedings of the IEEE international conference on
  computer vision workshops, 2017, pp. 879--886.

\bibitem{ref7.6}
J.~Wu, L.~Ma, X.~Hu, Delving deeper into convolutional neural networks for
  camera relocalization, in: 2017 IEEE International Conference on Robotics and
  Automation (ICRA), IEEE, 2017, pp. 5644--5651.

\bibitem{ref7.7}
A.~Kendall, R.~Cipolla, Geometric loss functions for camera pose regression
  with deep learning, in: Proceedings of the IEEE conference on computer vision
  and pattern recognition, 2017, pp. 5974--5983.

\bibitem{ref7.9}
M.~Bui, C.~Baur, N.~Navab, S.~Ilic, S.~Albarqouni, Adversarial networks for
  camera pose regression and refinement, in: Proceedings of the IEEE/CVF
  International Conference on Computer Vision Workshops, 2019, pp. 0--0.

\bibitem{ref7.11}
B.~Chidlovskii, A.~Sadek, Adversarial transfer of pose estimation regression,
  in: European Conference on Computer Vision, Springer, 2020, pp. 646--661.

\bibitem{ref7.10}
Z.~Huang, Y.~Xu, J.~Shi, X.~Zhou, H.~Bao, G.~Zhang, Prior guided dropout for
  robust visual localization in dynamic environments, in: Proceedings of the
  IEEE/CVF International Conference on Computer Vision, 2019, pp. 2791--2800.

\bibitem{ref7.12}
P.~Purkait, C.~Zhao, C.~Zach, Synthetic view generation for absolute pose
  regression and image synthesis., in: BMVC, 2018, p.~69.

\bibitem{ref7.13}
M.~Cai, C.~Shen, I.~Reid, A hybrid probabilistic model for camera
  relocalization, in: BMVC Press, 2019.

\bibitem{ref8.1}
S.~Brahmbhatt, J.~Gu, K.~Kim, J.~Hays, J.~Kautz, Geometry-aware learning of
  maps for camera localization, in: Proceedings of the IEEE Conference on
  Computer Vision and Pattern Recognition, 2018, pp. 2616--2625.

\bibitem{ref8.2}
F.~Xue, X.~Wang, Z.~Yan, Q.~Wang, J.~Wang, H.~Zha, Local supports global: Deep
  camera relocalization with sequence enhancement, in: Proceedings of the
  IEEE/CVF International Conference on Computer Vision, 2019, pp. 2841--2850.

\bibitem{ref8.3}
A.~Valada, N.~Radwan, W.~Burgard, Deep auxiliary learning for visual
  localization and odometry, in: 2018 IEEE international conference on robotics
  and automation (ICRA), IEEE, 2018, pp. 6939--6946.

\bibitem{ref8.4}
N.~Radwan, A.~Valada, W.~Burgard, Vlocnet++: Deep multitask learning for
  semantic visual localization and odometry, IEEE Robotics and Automation
  Letters 3~(4) (2018) 4407--4414.

\bibitem{ref8.5}
Y.~Lin, Z.~Liu, J.~Huang, C.~Wang, G.~Du, J.~Bai, S.~Lian, Deep global-relative
  networks for end-to-end 6-dof visual localization and odometry, in: Pacific
  Rim International Conference on Artificial Intelligence, Springer, 2019, pp.
  454--467.

\bibitem{ref8.6}
R.~Clark, S.~Wang, A.~Markham, N.~Trigoni, H.~Wen, Vidloc: A deep
  spatio-temporal model for 6-dof video-clip relocalization, in: Proceedings of
  the IEEE Conference on Computer Vision and Pattern Recognition, 2017, pp.
  6856--6864.

\bibitem{ref8.7}
C.~Szegedy, W.~Liu, Y.~Jia, P.~Sermanet, S.~Reed, D.~Anguelov, D.~Erhan,
  V.~Vanhoucke, A.~Rabinovich, Going deeper with convolutions, in: Proceedings
  of the IEEE conference on computer vision and pattern recognition, 2015, pp.
  1--9.

\bibitem{ref9.1}
Z.~Laskar, I.~Melekhov, S.~Kalia, J.~Kannala, Camera relocalization by
  computing pairwise relative poses using convolutional neural network, in:
  Proceedings of the IEEE International Conference on Computer Vision
  Workshops, 2017, pp. 929--938.

\bibitem{ref9.2}
V.~Balntas, S.~Li, V.~Prisacariu, Relocnet: Continuous metric learning
  relocalisation using neural nets, in: Proceedings of the European Conference
  on Computer Vision (ECCV), 2018, pp. 751--767.

\bibitem{ref9.3}
M.~Ding, Z.~Wang, J.~Sun, J.~Shi, P.~Luo, Camnet: Coarse-to-fine retrieval for
  camera re-localization, in: Proceedings of the IEEE/CVF International
  Conference on Computer Vision, 2019, pp. 2871--2880.

\bibitem{ref9.4}
Q.~Zhou, T.~Sattler, M.~Pollefeys, L.~Leal-Taixe, To learn or not to learn:
  Visual localization from essential matrices, in: 2020 IEEE International
  Conference on Robotics and Automation (ICRA), IEEE, 2020, pp. 3319--3326.

\bibitem{ref9.5}
I.~Melekhov, J.~Ylioinas, J.~Kannala, E.~Rahtu, Relative camera pose estimation
  using convolutional neural networks, in: International Conference on Advanced
  Concepts for Intelligent Vision Systems, Springer, 2017, pp. 675--687.

\bibitem{ref9.10}
H.~Aan{\ae}s, R.~R. Jensen, G.~Vogiatzis, E.~Tola, A.~B. Dahl, Large-scale data
  for multiple-view stereopsis, International Journal of Computer Vision
  120~(2) (2016) 153--168.

\bibitem{ref9.6}
S.~Saha, G.~Varma, C.~Jawahar, Improved visual relocalization by discovering
  anchor points, arXiv preprint arXiv:1811.04370.

\bibitem{ref10.2}
W.~Maddern, G.~Pascoe, C.~Linegar, P.~Newman, 1 year, 1000 km: The oxford
  robotcar dataset, The International Journal of Robotics Research 36~(1)
  (2017) 3--15.

\bibitem{ref10.4}
Y.~Li, N.~Snavely, D.~Huttenlocher, P.~Fua, Worldwide pose estimation using 3d
  point clouds, in: European conference on computer vision, Springer, 2012, pp.
  15--29.

\bibitem{ref10.5}
X.~Huang, P.~Wang, X.~Cheng, D.~Zhou, Q.~Geng, R.~Yang, The apolloscape open
  dataset for autonomous driving and its application, IEEE transactions on
  pattern analysis and machine intelligence 42~(10) (2019) 2702--2719.

\bibitem{ref6.41}
T.-Y. Yang, D.-K. Nguyen, H.~Heijnen, V.~Balntas, Ur2kid: Unifying retrieval,
  keypoint detection, and keypoint description without local correspondence
  supervision, arXiv preprint arXiv:2001.07252.

\bibitem{ref6.45}
T.~Shi, H.~Cui, Z.~Song, S.~Shen, Dense semantic 3d map based long-term visual
  localization with hybrid features, arXiv preprint arXiv:2005.10766.

\bibitem{ref4.12}
T.~Sattler, B.~Leibe, L.~Kobbelt, Efficient \& effective prioritized matching
  for large-scale image-based localization, IEEE transactions on pattern
  analysis and machine intelligence 39~(9) (2016) 1744--1756.

\bibitem{ref6.42}
M.~Geppert, P.~Liu, Z.~Cui, M.~Pollefeys, T.~Sattler, Efficient 2d-3d matching
  for multi-camera visual localization, in: 2019 International Conference on
  Robotics and Automation (ICRA), IEEE, 2019, pp. 5972--5978.

\bibitem{ref6.46}
E.~Brachmann, A.~Krull, S.~Nowozin, J.~Shotton, F.~Michel, S.~Gumhold,
  C.~Rother, Dsac-differentiable ransac for camera localization, in:
  Proceedings of the IEEE Conference on Computer Vision and Pattern
  Recognition, 2017, pp. 6684--6692.

\bibitem{ref6.47}
E.~Brachmann, C.~Rother, Expert sample consensus applied to camera
  re-localization, in: Proceedings of the IEEE/CVF International Conference on
  Computer Vision, 2019, pp. 7525--7534.

\bibitem{ref6.49}
X.~Li, J.~Ylioinas, J.~Kannala, Full-frame scene coordinate regression for
  image-based localization, arXiv preprint arXiv:1802.03237.

\bibitem{ref6.48}
E.~Brachmann, C.~Rother, Neural-guided ransac: Learning where to sample model
  hypotheses, in: Proceedings of the IEEE/CVF International Conference on
  Computer Vision, 2019, pp. 4322--4331.

\bibitem{ref6.50}
X.~Li, S.~Wang, Y.~Zhao, J.~Verbeek, J.~Kannala, Hierarchical scene coordinate
  classification and regression for visual localization, in: Proceedings of the
  IEEE/CVF Conference on Computer Vision and Pattern Recognition, 2020, pp.
  11983--11992.

\bibitem{ref6.51}
A.~Barroso-Laguna, E.~Riba, D.~Ponsa, K.~Mikolajczyk, Key. net: Keypoint
  detection by handcrafted and learned cnn filters, in: Proceedings of the
  IEEE/CVF International Conference on Computer Vision, 2019, pp. 5836--5844.

\bibitem{ref6.52}
Z.~Luo, T.~Shen, L.~Zhou, J.~Zhang, Y.~Yao, S.~Li, T.~Fang, L.~Quan,
  Contextdesc: Local descriptor augmentation with cross-modality context, in:
  Proceedings of the IEEE/CVF Conference on Computer Vision and Pattern
  Recognition, 2019, pp. 2527--2536.

\end{thebibliography}

\end{document}